\documentclass[journal]{IEEEtran}
\usepackage[utf8]{inputenc} 
\usepackage[T1]{fontenc}
\usepackage{url}              \usepackage{cite}             \usepackage[cmex10]{amsmath}  \usepackage{amsfonts}
\interdisplaylinepenalty=1000 \usepackage{mleftright}       \usepackage{cgitIEEE}		  \mleftright

\usepackage[dvipsnames]{xcolor}
\usepackage{algorithm}
\usepackage{algorithmic}
\usepackage{graphicx}
\graphicspath{{figures/}} \usepackage{hyperref}
\hypersetup{ colorlinks=true, linkcolor=black, citecolor=black, urlcolor=black, pdftitle={Your Title}, pdfauthor={Your Name}, bookmarksnumbered=true,
    bookmarksopen=true,
}
\usepackage{enumitem}
\usepackage{orcidlink}
\usepackage{balance}

\begin{document}
\title{Empirical Risk Minimization with $f$-Divergence Regularization}

\author{Francisco Daunas{\orcidlink{0009-0009-2038-9985}},
I{\~n}aki Esnaola{\orcidlink{0000-0001-5597-1718}},
Samir M. Perlaza{\orcidlink{0000-0002-1887-9215}}, and 
H.~Vincent Poor{\orcidlink{0000-0002-2062-131X}}\thanks{
This work was presented in part at the International Symposium on Information Theory (ISIT) 2024 in \cite{perlazaISIT2024a} and at the Information Theory Workshop (ITW) 2025 in~\cite{perlazaITW2025a}. This work is supported in part by a University of Sheffield ACSE PGR scholarship and Sheffield Publication scholarship.
Special thanks to Maxime Nicaise for helpful feedback and discussions that enhanced clarity over the results here presented.

F. Daunas is with the School of Electrical and Electronic Engineering, University of Sheffield, Sheffield S1 3JD, U.K.; (e-mail: j.daunastorres@sheffield.ac.uk).

I. Esnaola is with the School of Electrical and Electronic Engineering, University of Sheffield, Sheffield S1 3JD, U.K.; and also with the Department of Electrical and Computer Engineering, Princeton University, Princeton, NJ 08544 USA (e-mail: esnaola@sheffield.ac.uk).

Samir M. Perlaza is with INRIA, Centre Inria d'Universit\'e C\^ote d'Azur, 06902 Sophia Antipolis, France; also with the Department of Electrical and Computer Engineering, Princeton University, Princeton, NJ 08544 USA; and also with the GAATI Mathematics Laboratory, University of French Polynesia, 98702 Faaa, French Polynesia (e-mail: samir.perlaza@inria.fr).

H. Vincent Poor is with the Department of Electrical and Computer Engineering, Princeton University, Princeton, NJ 08544 USA (e-mail: poor@princeton.edu).
}
}
\maketitle

\begin{abstract}
In this \mydoc, the solution to the empirical risk minimization problem with $f$-divergence regularization (ERM-$f$DR) is presented and conditions under which the solution also serves as the solution to the minimization of the expected empirical risk subject to an $f$-divergence constraint are established.
The proposed approach extends applicability to a broader class of $f$-divergences than previously reported and yields theoretical results that recover previously known results. 
Additionally, the difference between the expected empirical risk of the ERM-$f$DR solution and that of its reference measure is characterized, providing insights into previously studied cases of $f$-divergences.
A central contribution is the introduction of the normalization function, a mathematical object that is critical in both the dual formulation and practical computation of the ERM-$f$DR solution.
This work presents an implicit characterization of the normalization function as a nonlinear ordinary differential equation (ODE), establishes its key properties, and subsequently leverages them to construct a numerical algorithm for approximating the normalization factor under mild assumptions.
Further analysis demonstrates structural equivalences between ERM-$f$DR problems with different $f$-divergences via transformations of the empirical risk.
Finally, the proposed algorithm is used to compute the training and test risks of ERM-$f$DR solutions under different $f$-divergence regularizers.
This numerical example highlights the practical implications of choosing different functions $f$ in ERM-$f$DR problems. \end{abstract}

\section{Introduction}
\label{sec:introduction}

Empirical risk minimization (ERM)~\cite{vapnik1964perceptron, vapnik1992principles, vapnik1993local, krzyzak1996nonparametric, zou2009generalization} is an underlying principle of statistical learning \cite{valiant1984theory, mcallester2003pac, shalev2014understanding, cullina2018pac, guedj2019free}, which serves as the basis for the fundamental analysis of many modern supervised and unsupervised learning algorithms \cite{boser1992training, dietterich1998approximate, xu2017global}.
In statistical learning, several frameworks, such as Bayesian inference, PAC-Bayesian learning, and information-theoretic learning, adopt a probabilistic perspective by characterizing algorithms as probability distributions over the hypothesis class conditioned on the available training data \cite{geman1984stochastic, pmlr-v83-cheng18a, aminian2021exact, wishart1928generalised, kolouri2016sliced}.
This perspective facilitates both uncertainty quantification and generalization analysis \cite{perlaza2024HAL, hafez2020conditioning, haddouche2020pacbayes, asadi2018chaining}.
Under the statistical perspective, ERM is often posed as an optimization problem regularized by an information-theoretic metric, as a measure of dissimilarity between two distributions.
These metrics include the family of $f$-divergences, R\'enyi divergence, Wasserstein distance, and maximum mean discrepancy~\cite{esposito2021generalization, wang2019information, dalalyan2024user, li2025regularization, peng2025information, futami2023informationtheoretic, huang2021stochastic}.
Among these, the $f$-divergences, first introduced in~\cite{csiszar1967information, csiszar1996maxent, ali1966general} and later expanded in~\cite{csiszar1975divergence, palomar2008lautum, sason2016fdivergence}, have received particular attention, with one of the most extensively studied cases involving ERM with relative entropy regularization, also known as ERM-RER~\cite{raginsky2016information, xu2017information, russo2019much, perlaza2024ERMRER, perlazaISIT2022, daunas2024TITAsymmetry}.
The importance of the ERM-RER problem is a consequence of its solution being a Gibbs probability distribution, highlighting its analytical tractability and deep connections to Bayesian learning~\cite{gibbs1902bookElementary, jiang2008gibbs, martino2018recycling, kuzborskij2019distribution}, as well as to the Boltzmann distribution in statistical physics \cite{boltzmann1902book1, boltzmann1905book2, ackley1985learning}.
Remarkably, this holds regardless of whether the reference measure is a probability measure or a $\sigma$-finite measure, as established in~\cite{perlaza2024ERMRER}.
Other works address the general case of $f$-divergences, known as ERM with $f$-divergence regularization (ERM-$f$DR).
The case of discrete sets of models in the ERM-$f$DR problem is explored in~\cite{teboulle1992entropic} and~\cite{beck2003mirror}, while more general settings are covered in~\cite{alquier2021non} and~\cite{perlazaISIT2024a}.

Recent works have shown that $f$-divergence regularization can improve the robustness of learning algorithms in the context of distributionally robust optimization (DRO)~\cite{wei2021optimizing, liu2023smoothed,namkoong2016stochastic}.
Beyond robustness, the ERM-$f$DR formulation has been applied to analyze the generalization behavior of learning algorithms under specific $f$-divergence regularization~\cite{aminian2021jensen, aminian2022tighter, lopez2018generalization, zou2024generalization, Wu2024OnGeneralization}.
More recently, this line of research has been extended to general $f$-divergence regularization. 
It offers insights into how different divergence choices affect learning stability and complexity, via bounds that describe the transition from weak to strong generalization~\cite{yao2025weak}.
In deep learning, ERM-$f$DR has been formulated as an $f$-divergence-based maximum a posteriori (MAP) estimation problem~\cite{novello2024f, letizia2024mutual, novello2025robust}, enabling a probabilistic interpretation of regularization that connects with variational inference and stochastic optimization techniques. 
In these analyses, the dependence between the learning algorithms and the training data is studied through the mutual $f$-information~\cite{masiha2023fdivergence, esposito2020robust, van2024generalized} in order to provide generalization bounds.

\subsection{Motivation}

Together, the developments mentioned above underscore the relevance of the ERM-$f$DR framework in both machine learning practice and theoretical analysis.
ERM-$f$DR flexibility and close relation to information-theoretic metrics make it a strong candidate for addressing challenges in learning, such as generalization, robustness, and uncertainty quantification.
A central focus in existing literature is the ERM-$f$DR using the relative entropy, also known as the Kullback-Leibler (KL) divergence, as the regularizer. This choice is due to its analytical tractability and strong connections to Bayesian learning~\cite{raginsky2016information, xu2017information, russo2019much, perlaza2024ERMRER}.
However, concentrating solely on KL divergence leaves a gap in the literature over the choice of regularizer.
For instance, some divergences emphasize tail behavior, others improve robustness to outliers or enhance sensitivity to model misspecification \cite{zhang2024variational}.
Hence, extending ERM-$f$DR theory can enable the principled selection of divergence functions based on desired performance criteria or domain-specific constraints, ultimately offering more control over the trade-off between fit and generalization~\cite{picard2025good}.

A critical bottleneck in both the theoretical analysis and practical implementation of the ERM-$f$DR solution is that it is generally known only up to a normalization factor~\cite{teboulle1992entropic, beck2003mirror, alquier2021non}. 
This poses serious challenges, as for many $f$-divergences, explicit expressions for such a normalization factor are unknown.
In practice, this limitation affects the feasibility of sampling methods from algorithms such as Markov Chain Monte Carlo (MCMC) and rejection sampling.
In MCMC, the normalization factor appears in the likelihood ratio, which is required for constructing valid transition probabilities.
In rejection sampling, the absence of a normalization constant prevents the definition of an efficient proposal distribution, making sampling impractical.
Additionally, the comparison of concentration sets arising from different $f$-divergence regularizations cannot be meaningfully established without knowing the normalization factor.
Even in cases in which a closed-form expression exists, such as with relative entropy (or KL divergence), the normalization factor corresponds to the log-partition function\cite{perlazaISIT2023a, InriaRR9591, hu2013kulback}, whose computation is especially difficult.
This is because it requires evaluating the empirical risk across the entire support of the reference measure, a task that is known to be $\#P$-hard\cite{bulatov2005complexity, bulatov2013complexity, mcquillan2013computational}.

\subsection{Contributions}
This work presents the solution to the ERM-$f$DR problem under mild conditions on $f$,  that is, strict convexity and differentiability.
Under these assumptions, the solution to the ERM-$f$DR problem also solves the minimization of the expected empirical risk subject to an $f$-divergence constraint with respect to a reference probability distribution. 
The proposed approach broadens the scope of ERM-$f$DR to encompass a wider range of $f$-divergences beyond those previously treated in the literature, including divergences such as Jensen-Shannon and Hellinger.
This work establishes a link between the ERM-$f$DR solution and its dual optimization counterpart via the Legendre-Fenchel transform. This relationship is significant for two key reasons:
First, it demonstrates that the duality gap between the ERM-$f$DR problem and its dual is zero.
This strong duality implies that the optimal value of the ERM-$f$DR problem evaluated at its solution can be characterized by the value of the corresponding dual problem at its solution.
Second, by exploiting properties of the Legendre-Fenchel transform and applying the implicit function theorem, a nonlinear ordinary differential equation (ODE) is derived for the normalization function.
Unlike prior work, which often treats this normalization as an arbitrary real value satisfying certain constraints, this approach yields a tractable expression for it.
It is important to note that solving such an ODE depends directly on the choice of $f$, and leaves open the problem for an explicit solution in terms of general $f$-divergences.
Then, the properties of the normalization function are analyzed and subsequently used to develop a numerical algorithm capable of approximating the normalization factor under the assumption of an integrable prior $Q$.
The properties of the ERM-$f$DR solution are studied, and general expressions for its expected empirical risk are provided, which recover existing results of specific divergences as well as new results.
Furthermore, the analysis establishes a general equivalence principle: under suitable transformations of the empirical risk, distinct ERM-$f$DR problems associated with different $f$-divergences may yield equivalent learning algorithms.
This insight reveals a structural relationship among divergence-based regularization schemes and the formulation of the risk in the learning problem.
Numerical examples support the theoretical findings and demonstrate how the choice of $f$-divergence impacts training and test performance.
The observed performance of different regularizers indicates that Jensen-Shannon and Hellinger divergences outperform those obtained using relative entropy and reverse relative entropy.
Notably, the generalization gap remains small for a wide range of $\lambda$ values, suggesting that appropriate tuning of the regularization parameter can control overfitting while maintaining test accuracy.
Overall, this work advances the theoretical foundations of ERM with divergence-based regularization and offers concrete tools for analyzing and implementing robust learning algorithms across a broad spectrum of divergence measures.

\section{Preliminaries}
\subsection{Notation}
Let $\Omega$ be an arbitrary subset of $\reals^{d}$, with $d \in \ints$, and let $\BorSigma{\Omega}$ denote the Borel $\sigma$-field on $\Omega$. The set of probability measures that can be defined upon the measurable space $\left(\Omega, \BorSigma{\Omega} \right)$ is denoted by~$\bigtriangleup(\Omega)$.
Given a probability measure $Q \in \bigtriangleup(\Omega)$, the set exclusively containing the probability measures in $\bigtriangleup(\Omega)$ that are absolutely continuous with respect to $Q$ is denoted by $\bigtriangleup_{Q}(\Omega)$. That is,
\begin{IEEEeqnarray}{rCl}
\label{DefSetTriangUp}
\bigtriangleup_{Q}(\Omega) & \triangleq & \{P\in \bigtriangleup(\Omega): P \ll Q \},
\end{IEEEeqnarray}
where the notation $P \ll Q$ stands for the measure $P$ being absolutely continuous with respect to the measure $Q$.
The Radon-Nikodym derivative of the measure $P$ with respect to $Q$ is denoted by $\frac{\diff P}{\diff Q}:\Omega\rightarrow [0,\infty)$.
Using this notation, an $f$-divergence is defined as follows
\begin{definition}[$f$-divergence~\cite{csiszar1967information}]
\label{Def_fDivergence}
Let $f:[0,\infty)\rightarrow \reals$ be a convex function with $f(1)= \dot{f}(1)=0$ and $f(0) \triangleq \lim_{u\rightarrow 0^+}f(u)$.
Let $P$ and $Q$ be two probability measures on the same measurable space, with $P$ absolutely continuous with $Q$.
The $f$-divergence of $P$ with respect to $Q$, denoted by $\KLf{P}{Q}$, is
\begin{equation}
\label{EqD_f}
\KLf{P}{Q} \triangleq \int f(\frac{\diff P}{\diff Q}(\thetav)) \diff Q(\thetav).
\end{equation}
\end{definition}
In the case in which the function $f$ in~\eqref{EqD_f} is continuous and differentiable, the derivative of the function~$f$ is denoted by
\begin{equation}
\label{EqDefDiffF}
\dot{f}: (0, \infty) \to \reals. 
\end{equation}
If the inverse of the function~$\dot{f}$ exists, it is denoted by 
\begin{equation}
\label{EqDefInvDiffF}
\dot{f}^{-1}: \reals \to (0, \infty). 
\end{equation}
Note that the function $f$ in \eqref{Def_fDivergence} can, without loss of generality, be chosen to satisfy
\begin{equation}
\label{EqCanonicalF}
\dot{f}(1) = 0.
\end{equation}
This condition can always be imposed since the $f$-divergence is invariant under affine transformations of $f$. Although such a normalization does not affect the value of the divergence, it is essential when comparing solutions associated with different $f$-divergences, as these are characterized through their derivatives; see Appendix~\ref{sec:AppendixA}, Lemma~\ref{lemm_canformFDiv}.
Furthermore, the function $\frac{\diff P}{\diff Q}$ in~\eqref{EqD_f} is the Radon-Nikodym derivative of $P$ with respect to $Q$ \cite{radon1913RND}, whose properties are well described in \cite{halmos1949application} and \cite{InriaRR9591}.
The Kullback-Leibler divergence (or relative entropy) of $P$ with respect to $Q$ under the assumption in \eqref{EqCanonicalF}, denoted by $\KL{P}{Q}$, is obtained from~\eqref{EqD_f} when the function $f$ satisfies $f(u) = u \log(u)-u+1$~(see \cite{kullback1951information}). More specifically,
\begin{equation}
\label{EqDefKL_D}
\KL{P}{Q} = \int (\frac{\diff P}{\diff Q}(\thetav) \log(\frac{\diff P}{\diff Q}(\thetav))-\frac{\diff P}{\diff Q}(\thetav))\diff Q(\thetav)+1.
\end{equation}

\subsection{The Learning Problem}
Let~$\set{M}$,~$\set{X}$ and~$\set{Y}$, with~$\set{M} \subseteq \reals^{d}$ and~$d \in \ints$, be sets of \emph{models}, \emph{patterns}, and \emph{labels}, respectively.
A pair $(x,y) \in \mathcal{X} \times \mathcal{Y}$ is referred to as a \emph{labeled pattern} or \emph{data point}, and a \emph{dataset} is represented by the tuple $((x_1, y_1), (x_2, y_2), \ldots,(x_n, y_n))\in ( \set{X} \times \set{Y} )^n$.
Let the function~$h: \set{M} \times \mathcal{X} \rightarrow \mathcal{Y}$ be such that the label assigned to a pattern $x \in \set{X}$ according to the model $\thetav \in \set{M}$ is $h(\thetav,x)$.
Then, given a dataset
\begin{equation}
\label{EqTheDataSet}
\vect{z} = \big((x_1, y_1), (x_2, y_2 ), \ldots, (x_n, y_n )\big)  \in ( \set{X} \times \set{Y} )^n,
\end{equation}
the objective is to obtain a model $\thetav \in \set{M}$, such that, for all $i \inCountK{n}$, the label assigned to the pattern $x_i$, which is $h(\thetav,x_i)$, is ``close'' to the label $y_i$.
This notion of ``closeness'' is formalized by the function
\begin{equation}
\label{EqEll}
    \ell: \set{Y} \times \set{Y} \rightarrow [0, \infty),
\end{equation}
such that the loss or risk induced by choosing the model $\thetav \in \set{M}$  with respect to the labeled pattern $(x_i, y_i)$, with $i\inCountK{n}$, is $\ell(h(\thetav,x_i),y_i)$.
The risk function $\ell$ is assumed to be nonnegative and to satisfy $\ell( y, y ) = 0$, for all $y\in\set{Y}$.

The \emph{empirical risk} induced by a model $\vect{\theta}$ with respect to the dataset $\vect{z}$ in~\eqref{EqTheDataSet} is determined by the function $\mathsf{L}_{\vect{z}}\!:\! \set{M} \rightarrow [0, \infty)$, which satisfies
\begin{IEEEeqnarray}{rcl}
\label{EqLxy}
\mathsf{L}_{\vect{z}} (\vect{\theta} )  &\ = \ &
\frac{1}{n}\sum_{i=1}^{n}  \ell ( h(\vect{\theta}, x_i), y_i ).
\end{IEEEeqnarray}
The expectation of the empirical risk $\mathsf{L}_{\vect{z}} (\vect{\theta} )$ in~\eqref{EqLxy} when~$\vect{\theta}$ is sampled from a probability measure $P \in \bigtriangleup(\set{M})$ is determined by the functional $\mathsf{R}_{\dset{z}}: \bigtriangleup(\set{M}) \rightarrow  [0, \infty)$, such that
\begin{equation}
\label{EqRxy}
\foo{R}_{\dset{z}}( P ) = \int \foo{L}_{\dset{z}} ( \thetav )  \diff P(\thetav).
\end{equation}

\section[The ERM-fDR Primal And Dual Problems]{The ERM-$f$DR Primal And Dual Problems}
\label{sec:ERMfDR}
\subsection[The ERM-fDR Primal Problem]{The ERM-$f$DR Primal Problem}
\label{sec:ERMfDRPrimal}
The ERM-$f$DR problem is parametrized by a probability measure $Q \in \bigtriangleup(\set{M})$, a positive real $\lambda$, and a function $f:[0,\infty)\to\reals$ that satisfies the conditions in Definition~\ref{Def_fDivergence}.
The measure $Q$ is referred to as the \emph{reference measure} and $\lambda$ as the \emph{regularization factor}.
Given the dataset~$\dset{z} \in (\set{X} \times \set{Y})^n$ in~\eqref{EqTheDataSet}, the \mbox{ERM-$f$DR} problem, with parameters~$Q$,~$\lambda$ and $f$, consists of the following optimization problem:
\begin{IEEEeqnarray}{rcl}
\label{EqOp_f_ERMRERNormal}
\min_{P \in \bigtriangleup_{Q}(\set{M})} & \quad \foo{R}_{\dset{z}} ( P ) + \lambda \KLf{P}{Q},
\end{IEEEeqnarray}
where the functional $\foo{R}_{\dset{z}}$ is defined in~\eqref{EqRxy} and the $f$-divergence $\foo{D}_{f}$ is defined in~\eqref{EqD_f}. The optimization problem in~\eqref{EqOp_f_ERMRERNormal} is closely related to the following optimization problem:
\vspace{-2mm}
\begin{subequations}
\label{EqOp_f_ERM_RND2}
\begin{IEEEeqnarray}{cCl}
	\min_{P \in \bigtriangleup_{Q}(\set{M})}
	& \quad & \foo{R}_{\dset{z}} ( P ),\\
	\text{s.t.} 
 	& &  \Divf{P}{Q} \leq \eta,\label{EqOp_f_ERM_RND2_c_s1}
\end{IEEEeqnarray}
\end{subequations}
with $\eta \in [0,\infty)$.
The optimization problems in~\eqref{EqOp_f_ERMRERNormal} and~\eqref{EqOp_f_ERM_RND2} do not share the same solutions when for all $\thetav \in \supp Q$, $\foo{L}_{\dset{z}}(\thetav) = c$, for some $c > 0$ and $\foo{L}_{\dset{z}}$ in~\eqref{EqLxy}.
More specifically, the set of solutions to the problem in~\eqref{EqOp_f_ERM_RND2} is  $\{P \in \bigtriangleup_{Q}(\set{M}): \Divf{P}{Q} \leq \eta\}$, while the set of solutions to~\eqref{EqOp_f_ERMRERNormal} is the singleton $\{Q\}$.
This distinction is mathematically significant but can be ignored in practice, as it arises only when the function $\foo{L}_{\dset{z}}$ is a constant almost surely with respect to $Q$.
In order to avoid the above case, the notion of separable empirical risk functions~\cite[Definition 5]{perlaza2024ERMRER} is adopted.

\begin{definition}[Separable Empirical Risk Function \cite{perlaza2024ERMRER}]
\label{Def_SeparableLxy}
The empirical risk function $\foo{L}_{\dset{z}}$ in~\eqref{EqLxy} is said to be separable with respect to a $\sigma$-finite measure $P\in \bigtriangleup(\set{M})$, if there exists a positive real $c>0$ and two subsets $\set{M}_{1} $ and $\set{M}_{2}$ of $\set{M}$ that are nonnegligible with respect to $P$, such that for all $(\thetav_1, \thetav_2) \in \set{M}_{1} \times \set{M}_{2}$,
\vspace{-2mm}
\begin{IEEEeqnarray}{rCCCCCl}
\foo{L}_{\dset{z}}(\thetav_1)& < & c & < &  \foo{L}_{\dset{z}}(\thetav_2)	& < & \infty.
\end{IEEEeqnarray}
\end{definition}

Essentially, a nonseparable empirical risk function with respect to the measure $Q$ in~\eqref{EqOp_f_ERMRERNormal} satisfies
\begin{IEEEeqnarray}{rCl}
Q(\{\thetav \in \set{M}:\foo{L}_{\dset{z}}(\thetav)=a\})& = & 1,
\end{IEEEeqnarray}
for some $a > 0$.
The solutions to the ERM-$f$DR problems in~\eqref{EqOp_f_ERMRERNormal} and~\eqref{EqOp_f_ERM_RND2} are presented under the following assumptions:
\begin{itemize}
\item[\namedlabel{assume:a}{$(a)$}] The function $f$ is strictly convex and differentiable;
\item[\namedlabel{assume:b}{$(b)$}] There exists a $\beta$ such that
\begin{subequations}
\label{EqfKrescConstrainAll}
\begin{equation}
\label{EqDefSetB}
\beta \in \left\lbrace t\in \reals: \forall \vect{\theta} \in \supp Q , 0 < \dot{f}^{-1} \left( -\frac{t + \foo{L}_{\vect{z}}(\thetav)}{\lambda} \right) \right\rbrace,
\end{equation}
and
\begin{IEEEeqnarray}{rCl}
\label{EqEqualToABigOne}
\int \dot{f}^{-1}(-\frac{\beta + \foo{L}_{\dset{z}}(\thetav)}{\lambda}) \diff Q(\thetav) & = & 1,
\IEEEeqnarraynumspace
\end{IEEEeqnarray}
where the function $\foo{L}_{\dset{z}}$ is defined in~\eqref{EqLxy}; and 
\item[\namedlabel{assume:c}{$(c)$}] The function $\foo{L}_{\dset{z}}$ is separable with respect to the probability measure $Q$. 
\end{subequations}
\end{itemize}
Under Assumptions \ref{assume:a}, \ref{assume:b} and~\ref{assume:c}, the following theorem shows that the problems in in~\eqref{EqOp_f_ERMRERNormal} and~\eqref{EqOp_f_ERM_RND2} share the same unique solution subject to a condition on the parameters $\lambda$ in~\eqref{EqOp_f_ERMRERNormal} and $\eta$ in~\eqref{EqOp_f_ERM_RND2}.
\begin{theorem}
\label{Theo_f_ERMRadNik}
Under Assumptions~\ref{assume:a} and~\ref{assume:b}, the solution to the optimization problem in~\eqref{EqOp_f_ERMRERNormal}, denoted $\Pgibbs{P}{Q} \in \bigtriangleup_{Q}(\set{M})$, is unique, and for all $\thetav \in \supp Q$, 
\begin{equation}
\label{EqGenpdffDv}
\frac{\diff \Pgibbs{P}{Q}}{\diff Q} ( \thetav ) = \dot{f}^{-1}(-\frac{\beta + \foo{L}_{\dset{z}}(\thetav)}{\lambda}),
\end{equation}
where the functions $\dot{f}^{-1}$ and $\foo{L}_{\dset{z}}$ are respectively defined in~\eqref{EqDefInvDiffF} and~\eqref{EqLxy}.
Moreover, under Assumptions~\ref{assume:a}, \ref{assume:b}, and~\ref{assume:c}, if $\lambda$ in~\eqref{EqOp_f_ERMRERNormal} and $\eta$ in~\eqref{EqOp_f_ERM_RND2} satisfy
\begin{equation}
\label{EqGenDivf_eta}
\Divf{\Pgibbs{P}{Q}}{Q} = \eta,
\end{equation}
then, the probability measure $\Pgibbs{P}{Q}$ in~\eqref{EqGenpdffDv} is also the unique solution to the optimization problem in~\eqref{EqOp_f_ERM_RND2}.
\end{theorem}
\begin{IEEEproof}
The proof is presented in Appendix~\ref{app_theo_f_ERMRadNik2}.
\end{IEEEproof}
Assumption~\ref{assume:b} in Theorem~\ref{Theo_f_ERMRadNik}, leads to observing that for all $\vect{\theta} \in \supp Q$, 
\begin{IEEEeqnarray}{rCl}
\label{EqJune24at8h26in2024}
\frac{\diff \Pgibbs{P}{Q}}{\diff Q} ( \thetav ) & > & 0. 
\end{IEEEeqnarray}
This observation leads to the following lemma.
\begin{lemma}
\label{lemm_mutuallyAbsCont}
Under Assumptions~\ref{assume:a} and~\ref{assume:b} in Theorem~\ref{Theo_f_ERMRadNik}, the probability measures~$Q$ and~$\Pgibbs{P}{Q}$ in~\eqref{EqGenpdffDv} are mutually absolutely continuous.
\end{lemma}
\begin{IEEEproof}
The proof is presented in Appendix~\ref{AppProoflemm_mutuallyAbsCont}.
\end{IEEEproof}
Lemma~\ref{lemm_mutuallyAbsCont} reveals that the support of the reference measure $Q$ establishes an inductive bias that cannot be overcome, regardless of the choice of $f$-divergence. That is, the support of the probability measure $\Pgibbs{P}{Q}$ in~\eqref{EqGenpdffDv} is identical to the support of the reference measure $Q$.
In summary, the use of any $f$-divergence regularization, under Assumptions~\ref{assume:a} and~\ref{assume:b} in Theorem~\ref{Theo_f_ERMRadNik}, inadvertently forces the solution to the optimization problem in~\eqref{EqOp_f_ERMRERNormal} to coincide with the support of the reference measure $Q$, independently of the training data. 
This observation has already been pointed out for particular cases. For instance, when the function $f$ is such that $f(u) = u\log(u)$ in \cite[Lemma~3]{perlaza2024ERMRER}; and $f(u) = - \log(u)$ in \cite[Lemma~3]{perlazaISIT2023a}.
In general, this observation is particularly important for choosing the reference measure~$Q$.  
\subsection{Examples }
\label{sec:CaseResult}

This section presents the solutions to the optimization problem in~\eqref{EqOp_f_ERMRERNormal} for specific choices of the function~$f$.

\subsubsection{Relative Entropy}
\label{SubSubKL}
Let the function $f:[0,\infty) \rightarrow \reals$ be such that $f(u) = u\log(u)-u+1$.
In this case, the resulting $f$-divergence $\KLf{P}{Q}$ is the relative entropy of $P$ with respect to $Q$. That is,
\begin{IEEEeqnarray}{rCl}
\label{Eq_f_KL_DivEq}
\KLf{P}{Q} & = & \KL{P}{Q},
\end{IEEEeqnarray}
where $\foo{D}$ is defined in~\eqref{EqDefKL_D}.
From~Theorem~\ref{Theo_f_ERMRadNik}, it holds that for all $\vect{\theta} \in \supp Q$,
\begin{IEEEeqnarray}{rCl}
\label{Eq_f_KL_dPdQ}
\frac{\diff \Pgibbs{P}{Q}}{\diff Q}(\thetav) & = & \exp(- \frac{\beta + \foo{L}_{\dset{z}}(\thetav) }{\lambda}),
\end{IEEEeqnarray}
where $\beta$ can be explicitly obtained using~\eqref{EqEqualToABigOne}, which yields
\begin{IEEEeqnarray}{rCl}
1 
  & = &  \int  \exp(- \frac{\beta + \foo{L}_{\dset{z}}(\thetav) }{\lambda}) \diff Q(\thetav)\\
  & = &  \exp(- \frac{\beta}{\lambda})\int \exp(- \frac{ \foo{L}_{\dset{z}}(\thetav) }{\lambda}) \diff Q(\thetav),
\end{IEEEeqnarray}
and thus,
\begin{IEEEeqnarray}{rCl}
\label{Eq_f_KL_dPdQ_beta}
\beta & = & \lambda \log(\int \exp(-\frac{\foo{L}_{\dset{z}}(\thetav)}{\lambda}) \diff Q(\thetav)).
\end{IEEEeqnarray}
Substituting~\eqref{Eq_f_KL_dPdQ_beta} into~\eqref{Eq_f_KL_dPdQ} yields
\begin{IEEEeqnarray}{rCl}
\frac{\diff \Pgibbs{P}{Q}}{\diff Q}(\thetav) & = & \frac{\exp(- \frac{1}{\lambda}\foo{L}_{\dset{z}}(\thetav))}{\int \exp(- \frac{1}{\lambda}\foo{L}_{\dset{z}}(\nuv)) \diff Q (
\vect{\nuv})}.\label{Eq_f_KL_dPdQ_s2}
\end{IEEEeqnarray}
This result has been independently reported by several authors in \cite{perlazaISIT2022,perlazaISIT2023b,perlaza2024ERMRER, raginsky2016information,zou2009generalization}, and proved via a large variety of methods.

\subsubsection{Reverse Relative Entropy}
\label{SubReverse}
Let the function $f:[0,\infty) \rightarrow \reals$ be such that $f(u) = -\log(u)+u-1$.
In this case, the resulting $f$-divergence $\KLf{P}{Q}$ is the relative entropy of $Q$ with respect to $P$. That is,
\begin{IEEEeqnarray}{rCl}
\label{Eq_f_KLr_DivEq}
\KLf{P}{Q} & = & \KL{Q}{P},
\end{IEEEeqnarray}
where $\foo{D}$ is defined in~\eqref{EqDefKL_D}.
This result, in contrast to the previous example, justifies referring to $\KLf{P}{Q}$ as the \emph{reverse relative entropy}.
From~Theorem~\ref{Theo_f_ERMRadNik}, it holds that for all $\vect{\theta} \in \supp Q$,
\begin{IEEEeqnarray}{rCl}
\label{Eq_f_rKL_dPdQ}
\frac{\diff \Pgibbs{P}{Q}}{\diff Q}(\thetav) & = & \frac{\lambda}{\lambda + \beta + \foo{L}_{\dset{z}}(\thetav)}.
\end{IEEEeqnarray}
Note that if the assumption $\dot{f}(1) = 0$ is relaxed and the function $f:[0,\infty) \rightarrow \reals$ is chosen as $f(u) = -\log(u)$, then the solutions to the optimization problem in~\eqref{EqOp_f_ERMRERNormal} satisfy, for all $\vect{\theta} \in \supp Q$,
\begin{IEEEeqnarray}{rCl}
\label{Eq_f_rKL_dPdQOld}
\frac{\diff \Pgibbs{P}{Q}}{\diff Q}(\thetav) & = & \frac{\lambda}{ \tilde{\beta} + \foo{L}_{\dset{z}}(\thetav)},
\end{IEEEeqnarray}
where, $\tilde{\beta} = \beta + \lambda$.
The relaxation of the affine shift is advantageous because, although no closed-form expression for $\beta$ in \eqref{Eq_f_rKL_dPdQ} is available under $\dot{f}(1)=0$, the regularization parameter $\lambda$ admits a closed-form expression in terms of the modified constant $\tilde{\beta}$. Specifically,
\begin{IEEEeqnarray}{rCl}
\label{Eq_f_rKL_dPdQ_beta}
\lambda & = & \frac{1}{\int \frac{1}{\tilde{\beta} + \foo{L}_{\dset{z}}(\thetav)}\diff Q(\thetav)},
\end{IEEEeqnarray}
which follows from~\eqref{EqEqualToABigOne}.
Note that~\eqref{Eq_f_rKL_dPdQ_beta} establishes a one-to-one relation between $\lambda$ and $\beta$. This observation leads to the following  
\begin{IEEEeqnarray}{rCl}
\frac{\diff \Pgibbs{P}{Q}}{\diff Q}(\thetav) & = & \frac{(\foo{L}_{\dset{z}}(\thetav)+\tilde{\beta})^{-1}}{\displaystyle\int (\foo{L}_{\dset{z}}(\nuv)+\tilde{\beta})^{-1} \diff Q (\nuv)}, \label{Eq_f_KL_rdPdQ_s2}
\end{IEEEeqnarray}
which follows from substituting~\eqref{Eq_f_rKL_dPdQ_beta} into~\eqref{Eq_f_rKL_dPdQOld}.
This result has been previously reported in~\cite{perlazaISIT2023a}.
Additionally, from \eqref{Eq_f_KL_rdPdQ_s2}, the expression in \eqref{Eq_f_rKL_dPdQ_beta} suggests that, for other choices of $f$-divergences, an appropriate selection of the affine shift applied to $f$ may likewise allow $\beta$ to be solved explicitly, while preserving the solution of the underlying optimization problem.

\subsubsection{Jeffreys Divergence}
\label{SecJeffreyDivExample}
Let the function $f:[0,\infty) \rightarrow \reals$ be such that $f(u) = u\log(u) - \log(u)$.
In this case, the resulting $f$-divergence $\KLf{P}{Q}$ is the Jeffreys divergence between $P$ and $Q$, also known as symmetrized relative entropy or symmetrized Kullback-Leibler divergence, which follows from observing that
\begin{IEEEeqnarray}{rCl}
\label{Eq_f_Jeff_DivEq}
\KLf{P}{Q} & = & \KL{P}{Q} + \KL{Q}{P},
\end{IEEEeqnarray}
where $\foo{D}$ is defined in~\eqref{EqDefKL_D}.
From~Theorem~\ref{Theo_f_ERMRadNik}, it holds that for all $\vect{\theta} \in \supp Q$,
\begin{IEEEeqnarray}{rCl}
\IEEEmulticolD{
\frac{\diff \Pgibbs{P}{Q}}{\diff Q}\!(\thetav)
}
& = &\! \exp(\!W_0\!(\exp(\!\frac{\beta\!+\! \lambda\!+\!\foo{L}_{\dset{z}}\!(\thetav)}{\lambda})) \!-\! \frac{\beta\!+\!\lambda\! +\! \foo{L}_{\dset{z}}\!(\thetav)}{\lambda}\!)\!,
\label{Eq_f_Jeff_dPdQ}
\end{IEEEeqnarray}
where the function $W_0:[0,\infty)\rightarrow [0,\infty)$ is the Lambert function, which for a function $g: \reals \to \reals$ such that $g(u) = u\exp(u)$ satisfies $W_0(g(x)) = x$. The coupling induced by the Lambert function in~\eqref{Eq_f_Jeff_dPdQ} between the parameters $\beta$ and $\lambda$ prevents the characterization of $\beta$ in closed-form.  
Hence, $\beta$ must be obtained via numerical methods using
\begin{equation}
\!1 = \!\!\int\!\!\exp(\!W_0\!(\exp(\!\frac{\beta\!+\!\lambda\!+\!\foo{L}_{\dset{z}}\!(\thetav)}{\lambda}))\!-\!\frac{\beta\!+\!\lambda\!+\!\foo{L}_{\dset{z}}\!(\thetav)}{\lambda}\!)\! \mathrm{d}Q(\vect{\theta}),
\end{equation}
which follows from~\eqref{EqEqualToABigOne}.

\subsubsection{Jensen-Shannon Divergence}
\label{SubJensenShannonDivExample}

Let the function $f:[0,\infty) \rightarrow \reals$ be such that $f(u) = u \log(\frac{2u}{u+1}) + \log(\frac{2}{u+1})$.
In this case, the resulting $f$-divergence $\Divf{P}{Q}$ is the Jensen-Shannon's divergence between $P$ and $Q$, and similarly to the Jeffreys divergence, it is also symmetric, which follows from observing that
\begin{IEEEeqnarray}{rCl}
\label{Eq_f_JS_DivEq}
\Divf{P}{Q} & = & \KL{P\Big}{\frac{P+Q}{2}} +\KL{Q\Big}{\frac{P+Q}{2}},
\end{IEEEeqnarray}
where $\foo{D}$ is defined in~\eqref{EqDefKL_D}.
From~Theorem~\ref{Theo_f_ERMRadNik}, it holds that for all $\vect{\theta} \in \supp Q$,
\begin{IEEEeqnarray}{rCl}
\label{Eq_f_JS_dPdQ}
\frac{\diff \Pgibbs{P}{Q}}{\diff Q}(\thetav)
& = & \frac{1}{2\exp(\frac{\beta + \foo{L}_{\dset{z}}(\thetav)}{\lambda}) - 1},
\end{IEEEeqnarray}
where $\beta =  - \lambda \log(2c)$ and $c > 0$ satisfies the following fixed point equation:
\begin{IEEEeqnarray}{rCl}
c & =  & \frac{1}{\displaystyle\int  \frac{1}{\exp\left(\frac{1}{\lambda} \mathsf{L}_{\vect{z}}\left( \vect{\theta}\right)  \right) - c }\mathrm{d}Q \left( \vect{\theta} \right)},
\end{IEEEeqnarray}
which follows from~\eqref{EqEqualToABigOne}.
\subsubsection{Hellinger Divergence}\label{SubSubHellingerDivExample}

Let the function $f:[0,\infty) \rightarrow \reals$ be such that $ f(x) = (1-\sqrt{x})^2$.
In this case, the resulting $f$-divergence $\KLf{P}{Q}$ is the Hellinger's divergence of $P$ with respect to $Q$, which satisfies
\begin{IEEEeqnarray}{rCl}
\label{Eq_f_Hell_DivEq}
\Divf{P}{Q} & = & \int (1-\sqrt{\frac{\diff P}{\diff Q}(\thetav)})^2\diff Q(\thetav).
\end{IEEEeqnarray}

From~Theorem~\ref{Theo_f_ERMRadNik}, it holds that for all $\vect{\theta} \in \supp Q$,
\begin{IEEEeqnarray}{rCl}
\label{Eq_f_Hell_dPdQ}
\frac{\diff \Pgibbs{P}{Q}}{\diff Q}(\thetav) & = & (\frac{\lambda}{\beta + \lambda +\foo{L}_{\dset{z}}(\thetav)})^2,
\end{IEEEeqnarray}
where $\beta$ is chosen to satisfy 
\begin{IEEEeqnarray}{rCl}
\int (\frac{\lambda}{\beta + \lambda +\foo{L}_{\dset{z}}(\thetav)})^2 \mathrm{d}Q(\vect{\theta}) & = & 1,
\end{IEEEeqnarray}
which follows from~\eqref{EqEqualToABigOne}.
\subsubsection{$\chi^2$ Divergence}
\label{SubSubChiDivExample}
Let the function $f:[0,\infty) \rightarrow \reals$ be such that $f(x) = x^2-1$.
In this case, the resulting $f$-divergence $\KLf{P}{Q}$ is the Pearson-divergence, also known as the $\chi^{2}$-divergence between~$P$ and~$Q$.
From~Theorem~\ref{Theo_f_ERMRadNik}, it holds that for all $\vect{\theta} \in \supp Q$,
\begin{IEEEeqnarray}{rCl}
\label{Eq_f_X2_dPdQ}
\frac{\diff \Pgibbs{P}{Q}}{\diff Q}(\thetav) & = & -\frac{ \beta + \foo{L}_{\dset{z}}(\thetav)}{2\lambda},
\end{IEEEeqnarray}
where $\beta$ satisfies~\eqref{EqEqualToABigOne}, which implies
\begin{IEEEeqnarray}{rCl}
\label{Eq_f_X2_dPdQ_pre_beta}
1 
& = & - \int( \frac{\beta}{2\lambda} + \frac{\foo{L}_{\dset{z}}(\thetav)}{2\lambda}) \diff Q(\thetav),
\end{IEEEeqnarray}
and thus,
\begin{IEEEeqnarray}{rCl}
\label{Eq_f_X2_dPdQ_beta}
\beta & = & -(\lambda+\foo{R}_{\dset{z}}(Q)).
\end{IEEEeqnarray}
Substituting~\eqref{Eq_f_X2_dPdQ_beta} into~\eqref{Eq_f_X2_dPdQ} yields
\begin{IEEEeqnarray}{rCl}
\frac{\diff \Pgibbs{P}{Q}}{\diff Q}(\thetav) & = & \frac{\lambda +\foo{R}_{\dset{z}}(Q) -\foo{L}_{\dset{z}}(\thetav)}{2\lambda} . \label{Eq_f_X2_rdPdQ_s2}
\end{IEEEeqnarray}
From~\eqref{Eq_f_X2_rdPdQ_s2}, it is easy to see that there exist cases in which Assumption~\ref{assume:b} is not met. 
For instance, if there exists a model $\thetav \in \supp Q$ for which $\lambda +\foo{R}_{\dset{z}}(Q) < \foo{L}_{\dset{z}}(\thetav)$, then, for such a model $\vect{\theta}$, it holds that $\dot{f}^{-1}(-\frac{\beta + \foo{L}_{\dset{z}}(\thetav)}{\lambda}) = \frac{\lambda +\foo{R}_{\dset{z}}(Q) -\foo{L}_{\dset{z}}(\thetav)}{2\lambda} < 0$, which implies condition~\eqref{EqDefSetB} does not hold.

When Assumptions~\ref{assume:a} and~\ref{assume:b} in Theorem~\ref{Theo_f_ERMRadNik} are not met, 
the solution to the optimization problems in~\eqref{EqOp_f_ERMRERNormal} and~\eqref{EqOp_f_ERM_RND2} cannot be obtained from Theorem~\ref{Theo_f_ERMRadNik}. Moreover, in such a case, nothing can be asserted on whether solutions to these optimization problems exist.  
In summary, the existence of the solution to the optimization problems in~\eqref{EqOp_f_ERMRERNormal} and~\eqref{EqOp_f_ERM_RND2} when Assumptions~\ref{assume:a} and~\ref{assume:b} in Theorem~\ref{Theo_f_ERMRadNik} are not met is still an open problem. 
Some interesting $f$-divergences do not meet both Assumptions~\ref{assume:a} and~\ref{assume:b} in Theorem~\ref{Theo_f_ERMRadNik}, which is the case of the total variation divergence.  
\subsubsection{Total Variation Divergence}
\label{SubSubTV}
Let the function $f:[0,\infty) \rightarrow \reals$ be such that $f(x) = \abs{x-1}$.
The resulting $f$-divergence $\Divf{P}{Q}$ is the total variation between $P$ and $Q$. Nonetheless, the function $f$ is not strictly convex and nondifferentiable at~$1$. Hence, the solution to the optimization problems in~\eqref{EqOp_f_ERMRERNormal} and~\eqref{EqOp_f_ERM_RND2} cannot be obtained from Theorem~\ref{Theo_f_ERMRadNik}.

\subsection[The ERM-fDR Dual Problem]{The ERM-$f$DR Dual Problem}
\label{sec:dual}
The duality principle \cite[Chapter~5]{boyd2004convex} enables the analysis of the optimization problem in~\eqref{EqOp_f_ERMRERNormal} by studying an alternative form, known as the dual problem.
In this section, this dual problem is derived using the Legendre-Fenchel transform~\cite{boyd2004convex}, which is defined below.
\begin{definition}[Legendre-Fenchel transform {\cite{boyd2004convex}}]
\label{DefLT_cnvxcnj}
Consider a function $f:\set{I}_1\rightarrow \set{I}_2$,with $\set{I}_i \subseteq \reals$, with $i \in \{1,2\}$. The Legendre-Fenchel transform of the function $f$, denoted by $f^{*}:\set{J} \rightarrow \reals$, is 
\begin{align}
\label{EqDefLT_cnvxcnj}
	f^{*}(v) & = \smash{\sup_{u\in \set{I}_1}}( vu- f(u))\\ \intertext{with}
\label{EqDefJinLFT}
	\set{J} & = \{v \in \reals :f^{*}(v)<\infty\}.
\end{align}
\end{definition}
Using this notation, consider the following problem: 
\begin{IEEEeqnarray}{rcl}
\label{EqOp_f_ERMRERDual}
  \min_{\beta \in \reals} & \quad \lambda \int f^{*}(-\frac{\beta+\foo{L}_{\dset{z}}(\thetav)}{\lambda})\diff Q(\thetav) + \beta,
\end{IEEEeqnarray}
where the real $\lambda$, the measure $Q$ and the function $f$ are those in~\eqref{EqOp_f_ERMRERNormal}; and the functions $\foo{L}_{\dset{z}}$ and $f^{*}$ are defined in~\eqref{EqLxy} and~\eqref{EqDefLT_cnvxcnj}, respectively.
An important observation for the analysis of the problem in~\eqref{EqOp_f_ERMRERDual} is that the equality in~\eqref{EqGenpdffDv} can be written in terms of the \emph{normalization function}, introduced in~\cite{perlazaISIT2024a} and defined hereunder.
\begin{definition}[Normalization Function]
Let the set $\set{A}_{Q, \dset{z}}\subseteq (0,\infty)$, be the set that contains all regularization factors $\lambda$ for which Assumption~\ref{assume:b} in Theorem~\ref{Theo_f_ERMRadNik} holds.
The normalization function of the problems in~\eqref{EqOp_f_ERMRERNormal} and~\eqref{EqOp_f_ERM_RND2}, denoted by
\begin{subequations}
\label{EqDefNormFunction}
\begin{equation}
\label{EqDefMapNormFunction}
N_{Q, \dset{z}}: \set{A}_{Q, \dset{z}} \rightarrow \set{B}_{Q, \dset{z}},
\end{equation}
with $\set{B}_{Q, \dset{z}}\subseteq \reals$, is such that for all $\lambda \in \set{A}_{Q, \dset{z}}$,
\begin{IEEEeqnarray}{rCl}
\label{EqReasonNisNormFoo}
\int \dot{f}^{-1}\left(-\frac{N_{Q, \dset{z}}(\lambda) + \foo{L}_{\dset{z}}(\thetav)}{\lambda}\right) \diff Q(\thetav) = 1,
\IEEEeqnarraynumspace
\end{IEEEeqnarray}
\end{subequations}
where the functions $\dot{f}^{-1}$ and $\foo{L}_{\dset{z}}$ are defined in~\eqref{EqDefInvDiffF} and~\eqref{EqLxy}.
\end{definition}
More specifically, the set $\set{A}_{Q, \dset{z}}$ in~\eqref{EqDefMapNormFunction} contains the regularization factors $\lambda$ for which the problem in~\eqref{EqOp_f_ERMRERNormal} has a solution.
The equality in~\eqref{EqReasonNisNormFoo} justifies referring to the function $N_{Q, \dset{z}}$ as the \emph{normalization function}, as it ensures that the measure $\Pgibbs{P}{Q}$ in~\eqref{EqGenpdffDv} is a probability measure, such that for all $\thetav \in \supp Q$,
\begin{equation}
\label{EqDivfKrescaling}
	\frac{\diff \Pgibbs{P}{Q}}{\diff Q} ( \thetav ) = \dot{f}^{-1}(-\frac{N_{Q, \dset{z}}(\lambda) + \foo{L}_{\dset{z}}(\thetav)}{\lambda}).
\end{equation}
The following theorem introduces the solution to the problem in~\eqref{EqOp_f_ERMRERDual}. 
\begin{theorem}
\label{Theo_dual_is_N}
Under Assumptions \ref{assume:a} and \ref{assume:b} in Theorem~\ref{Theo_f_ERMRadNik}, the solution to the optimization problem in~\eqref{EqOp_f_ERMRERDual} is $N_{Q, \dset{z}}(\lambda)$, where the function $N_{Q, \dset{z}}$ is defined in~\eqref{EqDefNormFunction}.
\end{theorem}
\begin{IEEEproof}
Let $G:\reals \to \reals$ be a function such that 
\begin{IEEEeqnarray}{rcl}
\label{EqDefDualCost}
  	G(\beta) &\ = \ & \lambda \int f^{*}(-\frac{\beta+\foo{L}_{\dset{z}}(\thetav)}{\lambda})\diff Q(\thetav) + \beta,
\end{IEEEeqnarray}
which is the objective function of the optimization problem in~\eqref{EqOp_f_ERMRERDual}. 
Note that from Assumption~\ref{assume:a} in Theorem~\ref{Theo_f_ERMRadNik} and Definition~\ref{DefLT_cnvxcnj}, the function $G$ in~\eqref{EqDefDualCost} is a strictly convex function, which satisfies
\begin{IEEEeqnarray}{rcl}
  	\!\!\!\frac{\diff}{\diff \beta}G(\beta) 
        &\ = \ &\! \frac{\diff}{\diff \beta}(\lambda \int f^{*}(-\frac{\beta+\foo{L}_{\dset{z}}(\thetav)}{\lambda})\diff Q(\thetav) + \beta)\ \ \\
&\ = \ & \lambda \int \frac{\diff}{\diff \beta}f^{*}(-\frac{\beta+\foo{L}_{\dset{z}}(\thetav)}{\lambda})\diff Q(\thetav) + 1\label{EqDiffG_beta_s2}\\
  	&\ = \ & - \int \dot{f^{*}}(-\frac{\beta+\foo{L}_{\dset{z}}(\thetav)}{\lambda})\diff Q(\thetav) + 1,\label{EqDiffG_beta_s3}
\end{IEEEeqnarray}
where $\dot{f^{*}}$ is the derivative of the function $f^{*}$ in~\eqref{EqOp_f_ERMRERDual}; and~\eqref{EqDiffG_beta_s2} follows from the dominated convergence theorem \cite[Theorem 1.6.9]{ash2000probability}.
Let the solution to the optimization problem in~\eqref{EqDefDualCost} be denoted by $\widehat{\beta} \in \reals$ and note that the derivative of the function $G$ evaluated at $\widehat{\beta}$ is equal to zero, that is
\begin{IEEEeqnarray}{rcl}
  	\int \dot{f^{*}}(-\frac{\widehat{\beta}+\foo{L}_{\dset{z}}(\thetav)}{\lambda})\diff Q(\thetav) &\ = \ & 1.\label{EqDiffG_betaStar}
\end{IEEEeqnarray}
From~\cite[Corollary~23.5.1]{rockafellar1970conjugate} and Assumption~\ref{assume:a} in Theorem~\ref{Theo_f_ERMRadNik}, the following equality holds for all $v \in \set{J}$, with $\set{J}$ in~\eqref{EqDefJinLFT},
\begin{IEEEeqnarray}{rCCCl}
\label{EqDefLFTDiffFNqz}
\frac{\diff}{\diff v}f^{*}(v) & = & \dot{f^{*}}(v) & = & \dot{f}^{-1}(v),
\end{IEEEeqnarray}
where the functions $\dot{f}^{-1}$ and $\dot{f^{*}}$ are defined in~\eqref{EqDefInvDiffF} and~\eqref{EqDiffG_beta_s3}, respectively.
From~\eqref{EqDiffG_betaStar} and~\eqref{EqDefLFTDiffFNqz}, it follows that
\begin{IEEEeqnarray}{rcl}
  	\int \dot{f}^{-1}(-\frac{\widehat{\beta}+\foo{L}_{\dset{z}}(\thetav)}{\lambda})\diff Q(\thetav) &\ = \ & 1,
\end{IEEEeqnarray}
which combined with~\eqref{EqReasonNisNormFoo} and Assumption~\ref{assume:b} in Theorem~\ref{Theo_f_ERMRadNik} yields
\begin{IEEEeqnarray}{rcl}
  	 N_{Q, \dset{z}}(\lambda) &\ = \ & \widehat{\beta}.
\end{IEEEeqnarray}
Note that $\hat{\beta}$ is unique as it is the minimum of a strictly convex function.
This observation completes the proof.
\end{IEEEproof}

The following lemma establishes that the problem in~\eqref{EqOp_f_ERMRERDual} is the dual problem to the ERM-$f$DR problem in~\eqref{EqOp_f_ERMRERNormal} and characterizes the difference between their optimal values, which is often referred to as duality gap \cite[Section 8.3]{luenberger1964observing}.
\begin{lemma}
\label{lemm_ZeroDualGap}
Under Assumptions~\ref{assume:a} and~\ref{assume:b} in Theorem~\ref{Theo_f_ERMRadNik}, the optimization problem in~\eqref{EqOp_f_ERMRERDual} is the dual problem to the ERM-$f$DR problem in~\eqref{EqOp_f_ERMRERNormal}.
Moreover, the duality gap is zero. 	
\end{lemma}
\begin{IEEEproof}
	Under Assumption~\ref{assume:a} in Theorem~\ref{Theo_f_ERMRadNik} and \cite[Section 3.3.2]{boyd2004convex}, it can be verified that for all $v \in \set{J}$, with $\set{J}$ in~\eqref{EqDefJinLFT}, the function $f^{*}$ in~\eqref{EqDefLT_cnvxcnj} satisfies 
	\begin{IEEEeqnarray}{rcl}
	\label{EqLFT_cool_equality}
  	f^{*}(v) &\ = \ & v\dot{f^{*}}(v)-f\big(\dot{f^{*}}(v)\big),
  	\IEEEeqnarraynumspace
	\end{IEEEeqnarray}
where the function $\dot{f}^{*}$ is the same as in~\eqref{EqDefLFTDiffFNqz}.
From Assumption~\ref{assume:a} in Theorem~\ref{Theo_f_ERMRadNik} and~\eqref{EqDefLFTDiffFNqz}, the Radon-Nikodym derivative $\frac{\diff \Pgibbs{P}{Q}}{\diff Q}$ in~\eqref{EqGenpdffDv} satisfies for all $\thetav \in \supp Q$,
	\begin{IEEEeqnarray}{rCl}
	\label{EqGenpdffDvLFT}
  	\frac{\diff \Pgibbs{P}{Q}}{\diff Q}(\thetav) &\ = \ & \dot{f^{*}}(-\frac{N_{Q, \dset{z}}(\lambda) + \foo{L}_{\dset{z}}(\thetav)}{\lambda}),
  	\IEEEeqnarraynumspace
	\end{IEEEeqnarray}
	where the functions $\foo{L}_{\dset{z}}$ and $N_{Q, \dset{z}}$ are defined in~\eqref{EqLxy} and~\eqref{EqDefNormFunction}, respectively.
	Then, from~\eqref{EqLFT_cool_equality} and~\eqref{EqGenpdffDvLFT}, it holds that for all $\thetav \in \supp Q$,
	\begin{IEEEeqnarray}{rcl}
	\IEEEmulticolD{
  	\!f^{*}\!(\!-\frac{ N_{Q, \dset{z}}(\lambda)\!+\!\foo{L}_{\dset{z}}(\thetav)}{\lambda}\!)
	}
  	& = &-\frac{ N_{Q, \dset{z}}(\lambda)\!+\!\foo{L}_{\dset{z}}(\thetav)}{\lambda}\frac{\diff \Pgibbs{P}{Q}}{\diff Q}(\thetav)\!-\!f\Bigg(\frac{\diff \Pgibbs{P}{Q}}{\diff Q}(\thetav)\!\!\Bigg)\!.  \ \ \ \ \label{EqPfDualLTequival}
	\end{IEEEeqnarray}
	Rearranging~\eqref{EqPfDualLTequival} yields
	\begin{IEEEeqnarray}{rcl}
	\IEEEmulticolD{
  	\!\!\foo{L}_{\dset{z}}(\thetav)\!+\!\lambda f\!\Bigg(\frac{\diff \Pgibbs{P}{Q}}{\diff Q}(\thetav)\!\!\Bigg)\frac{\diff Q}{\diff \Pgibbs{P}{Q}}(\thetav)
	}  
  	& = & -\lambda f^{*}\!\Big(\!-\!\frac{ N_{Q, \dset{z}}(\lambda)\!+\!\foo{L}_{\dset{z}}(\thetav)}{\lambda}\Big)\frac{\diff Q}{\diff \Pgibbs{P}{Q}}(\thetav)\!
	\splitR 
	- N_{Q, \dset{z}}(\lambda).\quad \ \ \label{EqPfDualLTArrange}
	\end{IEEEeqnarray}
	Taking the expectation of both sides of~\eqref{EqPfDualLTArrange} with respect to the probability measure $\Pgibbs{P}{Q}$ in~\eqref{EqGenpdffDv} yields
	\begin{IEEEeqnarray}{rcl}
	\IEEEmulticolD{
  	\foo{R}_{\dset{z}}(\Pgibbs{P}{Q})+\lambda \Divf{\Pgibbs{P}{Q}}{Q}  
  	}
  	& =  &- \lambda\! \int\! f^{*}\Big(\!-\frac{ N_{Q, \dset{z}}(\lambda) \!+\! \foo{L}_{\dset{z}}(\thetav)}{\lambda}\Big)\diff Q(\thetav)\!
	\splitR 
	- N_{Q, \dset{z}}(\lambda).\qquad \label{EqPfDualLTExpect}
	\end{IEEEeqnarray}
	Using Theorem~\ref{Theo_f_ERMRadNik} and Theorem~\ref{Theo_dual_is_N} in the left-hand and right-hand sides of~\eqref{EqPfDualLTExpect}, respectively, yields
	\begin{IEEEeqnarray}{rcl}
	\IEEEmulticolD{
  	\min_{P\in\bigtriangleup_{Q}(\set{M})}\foo{R}_{\dset{z}}(P)+\lambda \Divf{P}{Q}  
  	}
  	& = & \max_{\beta\in \reals} - \lambda\! \int f^{*}\Big(-\frac{ \beta + \foo{L}_{\dset{z}}(\thetav)}{\lambda}\Big)\diff Q(\thetav)-\beta.  \ \  \label{EqPfDualLTOpt}
	\end{IEEEeqnarray}
The proof that the optimization problem in~\eqref{EqOp_f_ERMRERDual} is the dual to the ERM-$f$DR problem in~\eqref{EqOp_f_ERMRERNormal} follows from~\eqref{EqPfDualLTOpt} and \cite[Theorem 1, Section 8.4]{luenberger1997bookOptimization}. The zero duality gap is established by the equality in~\eqref{EqPfDualLTOpt}, which completes the proof.
	\end{IEEEproof}
	
	Lemma~\ref{lemm_ZeroDualGap} builds on \cite[Theorem~29.1]{rockafellar1970conjugate} and is applied throughout the \mydoc. The zero duality gap ensures strong duality, meaning the optimal value of the dual variables can be used to recover the primal optimal measure. However, solving the dual problem in~\eqref{EqOp_f_ERMRERDual} remains equally challenging, as the solution must satisfy the constraints of Assumption~\ref{assume:b} in Theorem~\ref{Theo_f_ERMRadNik}.

\section{Analysis of the Normalization Function}
\label{sec:analysisRegFact}
\subsection{Implicit Characterization via Functional Properties}
The purpose of this section is to characterize the function $N_{Q, \dset{z}}$ and the sets $\set{A}_{Q, \dset{z}}$ and $\set{B}_{Q, \dset{z}}$ in~\eqref{EqDefNormFunction}.
Given a real~$\delta\in [0, \infty)$, consider the Rashomon set $\set{L}_{\dset{z}}(\delta)$, which is defined as follows
\begin{equation}
\label{EqType2LsetLamb2zero}
	\smash{\set{L}_{\dset{z}}(\delta) \triangleq \{\thetav \in \set{M}: \foo{L}_{\dset{z}}(\thetav) \leq \delta \}},
\end{equation}
where the function $\foo{L}_{\dset{z}}$ is defined in~\eqref{EqLxy}.
Consider also the real numbers $\delta^\star_{Q, \dset{z}}$ and $\lambda^\star_{Q, \dset{z}}$ defined as follows
\begin{IEEEeqnarray}{rCl}
\label{EqDefDeltaStar}
\delta^\star_{Q, \dset{z}} &\triangleq & \inf \{\delta \in [0, \infty): Q(\set{L}_{\dset{z}}(\delta))>0\},
\end{IEEEeqnarray}
and
\begin{IEEEeqnarray}{rCl}
	 \label{EqDefLambdaStar}
		\lambda^{\star}_{Q, \dset{z}} &\triangleq & \inf\set{A}_{Q, \dset{z}}.
\end{IEEEeqnarray}
Let also $\set{L}^{\star}_{Q, \dset{z}}$ be the level set of the empirical risk function $\foo{L}_{\dset{z}}$ in~\eqref{EqLxy} for the value $\delta^\star_{Q, \dset{z}}$. That is,
\begin{equation}
\label{EqDefSetLStarQz}
	\set{L}^{\star}_{Q, \dset{z}} \triangleq \{\thetav \in \set{M}: \foo{L}_{\dset{z}}(\thetav) = \delta^\star_{Q, \dset{z}}\}.
\end{equation}
Using this notation, the following theorem introduces one of the main properties of the function $N_{Q, \dset{z}}$.

\begin{theorem}
\label{Theo_InfDevKfDR}
The normalization function~$N_{Q, \dset{z}}$ in~\eqref{EqDefNormFunction}, under the assumption that the function $f$ is twice differentiable, is continuous within the interior of $\set{A}_{Q, \dset{z}}$ in~\eqref{EqDefMapNormFunction} and for all $\lambda \in \set{A}_{Q, \dset{z}}$, the value $N_{Q, \dset{z}}$ is unique.
\end{theorem}
\begin{IEEEproof}
	The proof is presented in Appendix~\ref{AppProofLemmaInfDevKDivf}.
\end{IEEEproof}
Theorem~\ref{Theo_InfDevKfDR} allows obtaining an implicit expression for the \emph{normalization function} in the form of a nonlinear ordinary differential equation, which is formalized by the following theorem.
\begin{theorem}
\label{Theo_ODE_NQz}
	 The function $N_{Q, \dset{z}}$ in~\eqref{EqDefNormFunction}, under the assumption that the function $f$ in~\eqref{EqDefNormFunction} is twice differentiable, satisfies  for all $\lambda \in \set{A}_{Q, \dset{z}}$,
\begin{IEEEeqnarray}{rcl}
\label{EqNQzExplisit}
	N_{Q, \dset{z}}(\lambda) 
	&\ = \ & \lambda \frac{\diff }{\diff \lambda}N_{Q, \dset{z}}(\lambda) - \foo{R}_{\dset{z}}\Big(P\Big),
\end{IEEEeqnarray}
where the functional $\foo{R}_{\dset{z}}$ is defined in~\eqref{EqRxy} and the probability measure $P \in \bigtriangleup_{Q}(\set{M})$ satisfies for all $\thetav \in \supp Q$,
\begin{IEEEeqnarray}{rcl}
	\frac{\diff P}{\diff Q}(\thetav) &\ = \ & \frac{\Bigg(\displaystyle\ddot{f}\Bigg(\frac{\diff \Pgibbs[\dset{z}]{P}{Q}}{\diff Q}(\thetav)\Bigg)\Bigg)^{-1}}{\displaystyle\int \Bigg(\ddot{f}\Bigg(\frac{\diff \Pgibbs[\dset{z}]{P}{Q}}{\diff Q}(\nuv)\Bigg)\Bigg)^{-1}\!\! \diff Q (\nuv)}; 
\end{IEEEeqnarray}
the function $\ddot{f}$ is the second derivative of the function $f$, and the measure $\Pgibbs{P}{Q}$ is that in~\eqref{EqGenpdffDv}.
\end{theorem}
\begin{IEEEproof}
The proof is presented in Appendix~\ref{app_proof_theo_ODE_NQz}.
\end{IEEEproof}
The results of Theorem~\ref{Theo_InfDevKfDR} and Theorem~\ref{Theo_ODE_NQz} do not yield an explicit expression that can be directly computed for the normalization function~\eqref{EqDefNormFunction}. Instead, these theorems establish key properties of the normalization function, which can be used to characterize further aspects of the function and to motivate approximation methods for the normalization parameter required in order to obtain the solution to the optimization problem~\eqref{EqOp_f_ERMRERNormal}.

\begin{lemma}
\label{lemm_NQz_decreas}
Under the assumption that the function $f$ is twice differentiable, and the function $\dot{f}^{-1}$ is log convex (see \cite[Definition~3.5.1]{boyd2004convex}). Then, the normalization function $N_{Q, \dset{z}}$ in~\eqref{EqDefNormFunction} is monotonically decreasing.
\end{lemma}
\begin{IEEEproof}
The proof is presented in Appendix~\ref{app_proof_lemm_monotologcvx}.
\end{IEEEproof}
The continuity and monotonicity exhibited by the function $N_{Q, \dset{z}}$ allow the following characterization of the set $\set{A}_{Q, \dset{z}}$.
\begin{lemma}
\label{lemm_fDR_kset}
The set $\set{A}_{Q, \dset{z}}$ in~\eqref{EqDefMapNormFunction} is either empty or a convex subset of $(0,\infty)$, that satisfies 
\begin{equation}
	\label{Eq_DivfConstrainOpen}
 			(\lambda^{\star}_{Q, \dset{z}}, \infty) \subseteq \set{A}_{Q, \dset{z}} \subseteq [\lambda^{\star}_{Q, \dset{z}}, \infty),
\end{equation}
where $\lambda^{\star}_{Q, \dset{z}}$ is defined in~\eqref{EqDefLambdaStar}.
\end{lemma}
\begin{IEEEproof}
	The proof is presented in Appendix~\ref{app_proof_lemm_fDR_kset}.
\end{IEEEproof}

Lemma~\ref{lemm_fDR_kset} highlights two facts. First, the set $\set{A}_{Q, \dset{z}}$ is a convex subset of positive reals. 
Second, if there exists a solution to the optimization problems in~\eqref{EqOp_f_ERMRERNormal} and~\eqref{EqOp_f_ERM_RND2} for some $\lambda>0$, then there exists a solution to such problems when $\lambda$ is replaced by~$\bar{\lambda} \in(\lambda,\infty)$.

The following lemma presents a case in which the set $\set{A}_{Q, \dset{z}}$ in~\eqref{EqDefMapNormFunction} can be fully characterized.
\begin{lemma}
\label{lemm_fDR_No_minRegF_nneg}
If the function $\dot{f}^{-1}$ in~\eqref{EqGenpdffDv} is strictly positive, then the set $\set{A}_{Q, \dset{z}}$ is identical to $(0,\infty)$.
\end{lemma}
\begin{IEEEproof}
The proof is presented in Appendix~\ref{app_proof_lemm_fDR_No_minRegF_nneg}.
\end{IEEEproof}

The condition of Lemma~\ref{lemm_fDR_No_minRegF_nneg} is satisfied by the \emph{Kullback-Leibler Divergence} (Section \ref{SubSubKL}), \emph{Jeffreys Divergence} (Section \ref{SecJeffreyDivExample}), and \emph{Hellinger Divergence} (Section \ref{SubSubHellingerDivExample}). For those cases, it holds that $\set{A}_{Q, \dset{z}} = (0,\infty)$.
Alternatively, the \emph{Reverse Relative Entropy Divergence} (Section \ref{SubReverse}), \emph{Jensen-Shannon Divergence} (Section \ref{SubJensenShannonDivExample}), and \emph{$\chi^2$ Divergence} (Section \ref{SubSubChiDivExample})  do not satisfy the  condition of Lemma~\ref{lemm_fDR_No_minRegF_nneg}.

\subsection{Efficient Approximation via Root-finding}
\label{sec:sub:algorithm}

A major bottleneck in statistical learning is formed by the computational challenges associated with: $(i)$ Evaluating expectations with respect to the prior $Q$ (see, for example, \eqref{EqEqualToABigOne}), and $(ii)$ Determining the normalization factor $\beta$ in~\eqref{EqGenpdffDv} for implementing the resulting algorithms.
This subsection discusses the significance of the continuity and monotonicity properties of the normalization function  $N_{Q,\vect{z}}$ in~\eqref{EqDefNormFunction} and their practical implications towards the challenges in $(ii)$,  under the assumption that expectations with respect to the prior $Q$ are computationally tractable.
For certain choices of $f$, an explicit expression for the normalization function cannot be obtained solely in terms of the regularization factor $\lambda$ as shown in \cite{daunas2024TITAsymmetry}.
However, the bijection property of $N_{Q,\vect{z}}$ resulting from Theorem~\ref{Theo_InfDevKfDR} implies that finding the inverse of the normalization function is enough for calculations.
For example, if $f(x) = -\log x$, the inverse normalization function $N_{Q, \dset{z}}^{-1}$  satisfies
\begin{IEEEeqnarray}{rCl}
    N_{Q, \dset{z}}^{-1}(\beta) & = & \smash{(\displaystyle\int \frac{1}{\foo{L}_{\dset{z}}(\thetav) + \beta} \diff Q(\thetav))^{-1}},
\end{IEEEeqnarray}
as shown in \cite[Theorem~1]{daunas2024TITAsymmetry}, which suffices to solve the optimization problem in~\eqref{EqOp_f_ERMRERNormal}.
Nevertheless, for certain $f$-divergences, neither the normalization function nor its inverse can be explicitly obtained. An example of such a case is the Hellinger divergence (see Example~\ref{SubSubHellingerDivExample}).
An alternative approach is a numerical approximation of the function $N_{Q, \dset{z}}$ leveraging the properties from Theorem~\ref{Theo_InfDevKfDR} and Lemma~\ref{lemm_fDR_kset}.
The following algorithm captures this intuition.

\begin{algorithm}[H]
\caption{Root-finding for $N_{Q, \dset{z}}(\lambda)$}
\begin{algorithmic}[1]
\renewcommand{\algorithmicrequire}{\textbf{Input:}}
\renewcommand{\algorithmicensure}{\textbf{Output:}}
\REQUIRE $\foo{L}_{\dset{z}}: \set{M} \to [0,\infty)$, $Q$, $f:[0,\infty) \to \mathbb{R}$, $\lambda > 0$, $\delta^{\star}_{Q, \dset{z}}$, tolerance $\epsilon > 0$, max iterations $N_{\text{max}}$
\ENSURE $\beta$ satisfying $\left|\int \dot{f}^{-1}\left(-\frac{\foo{L}_{\dset{z}}(\thetav)+\beta}{\lambda}\right) \diff Q(\thetav) - 1\right| \leq \epsilon$
\\ \textit{Initialization}:
\STATE $b_{\text{low}} \gets \delta^{\star}_{Q, \dset{z}}-\lambda\dot{f}(0)$, $b_{\text{high}} \gets \lambda$
\STATE $n \gets 0$, $b \gets \frac{1}{2}(b_{\text{low}} + b_{\text{high}})$
\STATE $I \gets \int  \dot{f}^{-1}\left(-\frac{\foo{L}_{\dset{z}}(\thetav)+b}{\lambda}\right) \diff Q(\thetav)$
\\ \textit{Root-Finding Process}:
\WHILE{$|I - 1| > \epsilon$ \AND $n < N_{\text{max}}$}
    \IF{$I > 1$}
        \STATE $b_{\text{high}} \gets b$
    \ELSE
        \STATE $b_{\text{low}} \gets b$
    \ENDIF
    \STATE $b \gets \frac{1}{2}(b_{\text{low}} + b_{\text{high}})$
    \STATE $I \gets \int \dot{f}^{-1}\left(-\frac{\foo{L}_{\dset{z}}(\thetav)+b}{\lambda}\right) \diff Q(\thetav)$, $n \gets n + 1$
\ENDWHILE
\RETURN $\beta = b$
\end{algorithmic}
\label{Algo_RootFind}
\end{algorithm}
The algorithm leverages the strict monotonicity of $N_{Q, \dset{z}}$, established in Theorem~\ref{Theo_InfDevKfDR}, along with the uniqueness of the ERM‑$f$DR solution shown in Theorem~\ref{Theo_f_ERMRadNik}.
Together, these properties imply that the integral in~\eqref{EqEqualToABigOne} is strictly monotonic in $\beta$, with a unique root that determines whether $\beta$ should increase or decrease.
In Section~\ref{sec:subSimExample}, this Algorithm~\ref{Algo_RootFind} is used to compute, for each $\lambda$, the corresponding $\beta$ for different $f$-divergence regularizers.
\section[Properties of the ERM-fDR solution]{Properties of the ERM-$f$DR solution}
\label{sec:propertiesRN}
This section studies the properties of the solution to the ERM-$f$DR problem in~\eqref{EqOp_f_ERMRERNormal} and~\eqref{EqOp_f_ERM_RND2}.
First, note that from Theorem~\ref{Theo_f_ERMRadNik} models resulting in lower empirical risks correspond to greater values of the Radon-Nikodym derivative $\frac{\diff \Pgibbs{P}{Q}}{\diff Q}$ in~\eqref{EqGenpdffDv}.
The following lemma formalizes this observation.
\begin{lemma}
\label{lemm_ERM_RER_Divf_ineq}
For all~$(\thetav_1, \thetav_2) \in \left( \supp Q \right)^2$, such that~$\foo{L}_{\dset{z}}(\thetav_1)\leq \foo{L}_{\dset{z}}(\thetav_2)$, with~$\foo{L}_{\dset{z}}$ in~\eqref{EqLxy}, the \RadonNikodym derivative in~\eqref{EqGenpdffDv} satisfies
\begin{equation}\label{Eq_lem_DivfineqRNd}
	\frac{\diff \Pgibbs{P}{Q}}{\diff Q}(\thetav_2) \leq \frac{\diff \Pgibbs{P}{Q}}{\diff Q}(\thetav_1),
\end{equation}
with equality if and only if~$\foo{L}_{\dset{z}}(\thetav_1) = \foo{L}_{\dset{z}}(\thetav_2)$.
\end{lemma}
\begin{IEEEproof}
The proof is presented in Appendix~\ref{app_lemm_ERM_RER_Divf_ineq}.
\end{IEEEproof}
Lemma~\ref{lemm_ERM_RER_Divf_ineq} aligns with the intuition of assigning greater probability mass to models that achieve lower risk on the dataset $\dset{z}$.
Second, the Radon-Nikodym derivative~$\frac{\diff \Pgibbs{P}{Q}}{\diff Q}$ in~\eqref{EqGenpdffDv} is always finite and strictly positive.
This observation is formalized by the following lemma.

\begin{lemma}
\label{lemm_ERM_RER_Divf_RNdBounds} 
The Radon-Nikodym derivative~$\frac{ \diff \Pgibbs{P}{Q}}{\diff Q}$ in~\eqref{EqGenpdffDv} satisfies for all~$\thetav \in  \supp Q$
\begin{equation}
	0 < \frac{ \diff \Pgibbs{P}{Q}}{\diff Q}(\thetav) \leq \dot{f}^{-1}(-\frac{\delta^\star_{Q, \dset{z}} + \beta}{\lambda}) < \infty,
\end{equation}
where the equality holds if and only if~$\thetav \in \set{L}^{\star}_{Q, \dset{z}}\cap \supp Q$.
\end{lemma}
\begin{IEEEproof}
The proof is presented in Appendix~\ref{app_lemm_ERM_RER_Divf_RNdBounds}.
\end{IEEEproof}

The next lemma shows that the expected empirical risk induced by the solution to the optimization problems in~\eqref{EqOp_f_ERMRERNormal} and~\eqref{EqOp_f_ERM_RND2}, $\Pgibbs{P}{Q}$ in~\eqref{EqGenpdffDv}, monotonically decreases with respect to the regularization factor $\lambda$.
This observation is formalized by the following lemma.

\begin{lemma}
\label{lemm_ERM_fDR_Rz_Lambda}
For all $(\lambda_1,\lambda_2) \in (0,\infty)^{2}$, such that $\lambda_1 < \lambda_2$, the probability measures $\Pgibbs[\dset{z}][\lambda_1]{P}{Q}$ and $\Pgibbs[\dset{z}][\lambda_2]{P}{Q}$ in~\eqref{EqGenpdffDv} satisfy
\begin{IEEEeqnarray}{rCcCl}
	\delta^{\star}_{Q, \dset{z}} \leq \foo{R}_{\dset{z}}(\Pgibbs[\dset{z}][\lambda_1]{P}{Q}) 
	& \leq & \foo{R}_{\dset{z}}(\Pgibbs[\dset{z}][\lambda_2]{P}{Q})
	& \leq & \foo{R}_{\dset{z}}(Q),
\end{IEEEeqnarray}
where $\delta^{\star}_{Q, \dset{z}}$ is defined in~\eqref{EqDefDeltaStar}, the functional $\foo{R}_{\dset{z}}$ is defined in~\eqref{EqRxy} and the equalities simultaneously hold if and only if~$\foo{L}_{\dset{z}}$ in~\eqref{EqLxy} is nonseparable with respect to the probability measure $Q$.
\end{lemma}
\begin{IEEEproof}
The proof follows from the preliminary results in Lemma~\ref{lemm_UpperboundRQ} and Lemma~\ref{lemm_RP1LessRP2}, presented in Appendix~\ref{sec:AppendixA}.
\end{IEEEproof}

From \cite[Corollary~23.5.1]{rockafellar1970conjugate}, given a convex function $f$, the derivative of the Legendre-Fenchel transform of $f^{*}$ in Definition~\ref{DefLT_cnvxcnj}, denoted by $\dot{f^{*}}: \set{J} \to \set{I}_1$, with the sets $\set{I}_1$ and $\set{J}$ in Definition~\ref{DefLT_cnvxcnj}, satisfies for all $v \in \set{J}$,
\begin{IEEEeqnarray}{rCCCl}
\label{EqDefLFTDiffF}
\frac{\diff}{\diff v}f^{*}(v) & = & \dot{f^{*}}(v) & = & \dot{f}^{-1}(v),
\end{IEEEeqnarray}
with the function $\dot{f}^{-1}$ defined in~\eqref{EqDefInvDiffF}. 
This observation leads to the following corollary.
\begin{corollary}
	The Radon-Nikodym derivative~$\frac{ \diff \Pgibbs{P}{Q}}{\diff Q}$ in~\eqref{EqGenpdffDv} satisfies for all~$\thetav \in  \supp Q$,
\begin{IEEEeqnarray}{rCl}
		\frac{ \diff \Pgibbs{P}{Q}}{\diff Q}(\thetav) & = & \dot{f^{*}}(-\frac{\foo{L}_{\dset{z}}(\thetav)+N_{Q, \dset{z}}(\lambda)}{\lambda}),
	\end{IEEEeqnarray}
	where the functions $\foo{L}_{\dset{z}}$, $N_{Q, \dset{z}}$ and $\dot{f^{*}}$ are defined in~\eqref{EqLxy}, \eqref{EqDefNormFunction} and~\eqref{EqDefLFTDiffF}, respectively.
\end{corollary}
Furthermore, from \cite[Corollary~23.5.1]{rockafellar1970conjugate} and~\eqref{EqDefLFTDiffF}, it follows that for all $u \in \set{I}_1$,
\begin{IEEEeqnarray}{rCl}
\label{EqDefLFTequal}
f^{*}(\dot{f}(u)) & = & \dot{f}(u)u - f(u).
\end{IEEEeqnarray}
The following theorem specifies the value of the objective function in the optimization problem defined in~\eqref{EqOp_f_ERMRERNormal}, evaluated at the solution $\Pgibbs{P}{Q}$ given in~\eqref{EqGenpdffDv}.

\begin{theorem}
\label{Theo_ERM_fDR_LT}
Let the function $g:[0,\infty)\to \reals$ be $g(u)=\frac{1}{u}f(u)$, with $f$ in~\eqref{EqOp_f_ERMRERNormal}. The probability measure $\Pgibbs{P}{Q}$in~\eqref{EqGenpdffDv} satisfies
\begin{IEEEeqnarray}{rCl}
\IEEEmulticolR{
\IEEEmulticolD{
	\foo{R}_{\dset{z}}(\Pgibbs{P}{Q}) +\lambda \Divf{\Pgibbs{P}{Q}}{Q}
}}
	& = & -\lambda \int\! f^{*}(-\frac{\foo{L}_{\dset{z}}(\thetav) + N_{Q, \dset{z}}(\lambda)}{\lambda})\diff Q(\thetav) - N_{Q, \dset{z}}(\lambda),\quad \label{EqPf_ERM_fDR_LT}
\end{IEEEeqnarray}
and 
\begin{IEEEeqnarray}{rCl}
\IEEEmulticolR{
\IEEEmulticolD{
	\foo{R}_{\dset{z}}(Q)+\!\lambda\int\!\!g\Bigg(\frac{\diff \Pgibbs{P}{Q}}{\diff Q}(\thetav)\Bigg) \diff Q(\thetav)
}}
	& = & -\lambda\!\int\!\! f^{*}(\!-\frac{\foo{L}_{\dset{z}}(\thetav) \!+\! N_{Q, \dset{z}}(\lambda)}{\lambda})\frac{\diff Q}{\diff \Pgibbs{P}{Q}}(\thetav) \diff Q(\thetav)
	\splitD
	\!-N_{Q, \dset{z}}(\lambda), \quad\label{EqTheo_ERM_fDR_LT_sub2}
\end{IEEEeqnarray}
where $f^{*}$ is the Legendre-Fenchel transform of $f$ (see  Definition~\ref{DefLT_cnvxcnj}); the functions $\foo{L}_{\dset{z}}$ and $N_{Q, \dset{z}}$ are defined in~\eqref{EqLxy} and~\eqref{EqDefNormFunction}, respectively; and the functional $\foo{R}_{\dset{z}}$ is defined in~\eqref{EqRxy}.
\end{theorem}
\begin{IEEEproof}
The proof is presented in Appendix~\ref{app_theo_ERM_fDR_LT}.
\end{IEEEproof}
Note that the function $g$ in~\eqref{EqTheo_ERM_fDR_LT_sub2} is not necessarily a convex function and thus it may not induce an $f$-divergence as in~Definition~\ref{Def_fDivergence}.
Additionally, Theorem~\ref{Theo_ERM_fDR_LT} provides an upper bound for the expected empirical risk in terms of the normalization function. This is formalized by the following lemma.
\begin{lemma}
\label{lemm_CharRzJensen}
The probability measure $\Pgibbs{P}{Q}$ in~\eqref{EqGenpdffDv} satisfies
\begin{IEEEeqnarray}{rCl}
\IEEEmulticolD{
\foo{R}_{\dset{z}}(\Pgibbs{P}{Q}) + \lambda \Divf{\Pgibbs{P}{Q}}{Q}
}
& \leq & -\lambda f^{*}(-\frac{1}{\lambda}\foo{R}_{\dset{z}}(Q)-\frac{N_{Q, \dset{z}}(\lambda)}{\lambda})
\splitR 
- N_{Q, \dset{z}}(\lambda),
\end{IEEEeqnarray}
where the functions $\foo{R}_{\dset{z}}$ and $N_{Q, \dset{z}}$ are defined in~\eqref{EqRxy} and~\eqref{EqDefNormFunction}, respectively. The equality holds if and only if the Legendre-Fenchel transform of $f$ is a linear function.
\end{lemma}
\begin{IEEEproof}
The proof follows from the fact that $f^{*}$ is strictly convex and using Jensen inequality in~\eqref{EqPf_ERM_fDR_LT}.
\end{IEEEproof}

The following examples show Theorem~\ref{Theo_ERM_fDR_LT} and Lemma~\ref{lemm_CharRzJensen} for specific choices of $f$-divergence regularization, recovering existing results as well as yielding new insights.
\subsection*{Relative Entropy}
Let the function $f:[0,\infty) \rightarrow \reals$ be such that $f(u) = u\log(u)-u+1$, and note that the Legendre-Fenchel transform of the function $f$ satisfies
\begin{IEEEeqnarray}{rCl}
f^{*}(v) & = & \exp(v)-1.
\label{Eq_f_KL_LFT_s2}
\end{IEEEeqnarray}
Furthermore, under relative entropy regularization, the normalization function is directly related to the cumulant generating function $K_{Q, \dset{z}}: (0,\infty) \rightarrow \reals$,
\begin{IEEEeqnarray}{rCl}
\label{Eq_K_cumm}
K_{Q, \dset{z}}(-\frac{1}{\lambda})
& = &  \log(\int \exp(-\frac{\foo{L}_{\dset{z}}(\thetav)}{\lambda}) \diff Q(\thetav)),
\end{IEEEeqnarray}
introduced in \cite[Eq. (22)]{perlaza2024ERMRER}. The relation of the normalization and cumulant generating function is such that for all $\lambda \in \set{A}_{Q, \dset{z}}$, with $\set{A}_{Q, \dset{z}}$ in~\eqref{EqDefMapNormFunction}, the function $N_{Q, \dset{z}}$ satisfies that
\begin{IEEEeqnarray}{rCl}
\label{Eq_f_KL_LFT_dPdQ_beta}
N_{Q, \dset{z}}(\lambda) 
& = & \lambda K_{Q, \dset{z}}(-\frac{1}{\lambda}),
\label{Eq_f_KL_LFT_dPdQ_beta_s2}
\end{IEEEeqnarray}
where~\eqref{Eq_f_KL_LFT_dPdQ_beta} follows from~\eqref{Eq_f_KL_dPdQ_beta}, and~\eqref{Eq_f_KL_LFT_dPdQ_beta_s2} follows from the definition of $K_{Q, \dset{z}}$ in~\eqref{Eq_K_cumm}.
Substituting~\eqref{Eq_f_KL_LFT_s2} and~\eqref{Eq_f_KL_LFT_dPdQ_beta} into~\eqref{EqPf_ERM_fDR_LT} yields
\begin{IEEEeqnarray}{rCl}
	\IEEEmulticolR{
	\IEEEmulticolD{
	\foo{R}_{\dset{z}}(\Pgibbs{P}{Q}) +\lambda \KL{\Pgibbs{P}{Q}}{Q}
	}}
	& = & \!-\lambda\!\!\int\!(\!\exp(\!-\frac{N_{Q, \dset{z}}(\lambda)\!+\!\foo{L}_{\dset{z}}(\thetav) }{\lambda}\!)\!\!-\!1\!)\diff Q(\thetav) \!-\! N_{Q, \dset{z}}(\lambda)\quad\ \ 
	\label{Eq_f_KL_LFT_Rz}\\
	& = &\lambda -\lambda \int\frac{\exp(- \frac{1}{\lambda}\foo{L}_{\dset{z}}(\thetav))}{\int \exp(- \frac{1}{\lambda}\foo{L}_{\dset{z}}(\nuv)) \diff Q(\nuv)}\diff Q(\thetav) - N_{Q, \dset{z}}(\lambda)\\
	& = & \lambda - \lambda \int \frac{\diff \Pgibbs{P}{Q}}{\diff Q}(\thetav) \diff Q(\thetav) - \lambda K_{Q, \dset{z}}(-\frac{1}{\lambda})\\
	& = &  -\lambda K_{Q, \dset{z}}(-\frac{1}{\lambda}).
	\label{Eq_f_KL_LFT_Rz_s4}
\end{IEEEeqnarray}
This result has been previously reported by other authors in \cite[Lemma 3]{perlazaISIT2023b} and \cite[Lemma 20]{perlaza2024ERMRER} for the case of relative entropy regularization.
From~\eqref{EqTheo_ERM_fDR_LT_sub2} and~\eqref{Eq_f_KL_LFT_s2}, the expected empirical risk induced by the reference measure $Q$ in~\eqref{EqOp_f_ERMRERNormal} satisfies that
\begin{IEEEeqnarray}{rCl}
	\IEEEmulticolR{
	\IEEEmulticolD{
	\foo{R}_{\dset{z}}(Q) 
	}}
	& = & -\lambda\int\!(\!g(\frac{\diff \Pgibbs{P}{Q}}{\diff Q}(\thetav)\!)\!
	\splitD[1]
	-\! \dot{g}(\!\frac{\diff \Pgibbs{P}{Q}}{\diff Q}(\thetav)\!)\!) \diff \Pgibbs{P}{Q}(\thetav)\!-\!N_{Q, \dset{z}}(\lambda)
	\label{EqTh2_eq70KL_s1}\\
	& = & -\lambda\int\!(\frac{\diff Q}{\diff \Pgibbs{P}{Q}}(\thetav)f(\frac{\diff \Pgibbs{P}{Q}}{\diff Q}(\thetav)) 
	\splitR[1] \splitD[1]
	- \frac{\diff Q}{\diff \Pgibbs{P}{Q}}(\thetav) f^{*}\!(\!-\frac{\foo{L}_{\dset{z}}(\thetav) \!+\! N_{Q, \dset{z}}(\lambda)}{\lambda}\!)\!) \diff Q(\thetav)
	\splitD 
	- N_{Q, \dset{z}}(\lambda)
	\label{EqTh2_eq70KL_s2}\qquad\\
	& = & 
	-\lambda\!\int\!\frac{\diff Q}{\diff \Pgibbs{P}{Q}}(\thetav)\! \exp(\!-\frac{\foo{L}_{\dset{z}}(\thetav) \!+\! N_{Q, \dset{z}}(\lambda)}{\lambda}\!-\!1\!) \diff Q(\thetav)
	\splitR \splitD
	-\lambda\int \log(\frac{\diff \Pgibbs{P}{Q}}{\diff Q}(\thetav)) \diff Q(\thetav) - N_{Q, \dset{z}}(\lambda) \quad
	\label{EqTh2_eq70KL_s3}\\
	& = & \lambda\int \log(\frac{\diff Q}{\diff \Pgibbs{P}{Q}}(\thetav)) \diff Q(\thetav) +\lambda - \lambda K_{Q, \dset{z}}(-\frac{1}{\lambda})
	\splitR \splitD
	-\lambda\!\int\!\frac{\exp(-\frac{\foo{L}_{\dset{z}}(\thetav)}{\lambda} - K_{Q, \dset{z}}(-\frac{1}{\lambda}))}{\exp(-\frac{\foo{L}_{\dset{z}}(\thetav)}{\lambda} - K_{Q, \dset{z}}(-\frac{1}{\lambda}))} \diff Q(\thetav)
	\label{EqTh2_eq70KL_s4}\\
	& = & \lambda\KL{Q}{\Pgibbs{P}{Q}} -\lambda +\lambda - \lambda K_{Q, \dset{z}}(-\frac{1}{\lambda})\label{EqTh2_eq70KL_s5}\\
	& = & \lambda(\KL{Q}{\Pgibbs{P}{Q}}- K_{Q, \dset{z}}(-\frac{1}{\lambda})),\label{EqTh2_eq70KL_s6}
\end{IEEEeqnarray}
where the function $g$ in~\eqref{EqTh2_eq70KL_s1} is defined in Theorem~\ref{Theo_ERM_fDR_LT}, \eqref{EqTh2_eq70KL_s2} follows from~\eqref{EqDefLFTDiffF};~\eqref{EqTh2_eq70KL_s3} follows from~\eqref{Eq_f_KL_LFT_s2}; and~\eqref{EqTh2_eq70KL_s4} follows from~\eqref{Eq_f_KL_LFT_dPdQ_beta}.
Rearranging the terms in~\eqref{EqTh2_eq70KL_s6} yields
\begin{IEEEeqnarray}{rCl}
	\foo{R}_{\dset{z}}(Q) - \lambda\KL{Q}{\Pgibbs{P}{Q}} & = & - \lambda K_{Q, \dset{z}}(-\frac{1}{\lambda}),
\end{IEEEeqnarray}
where the function $K_{Q, \dset{z}}$ is defined in~\eqref{Eq_K_cumm}.
This result has also been reported in \cite[Lemma 20]{perlaza2024ERMRER}.
Furthermore, the difference of~\eqref{Eq_f_KL_LFT_Rz_s4} and~\eqref{EqTh2_eq70KL_s6} yields,
\begin{IEEEeqnarray}{rCl}
	\IEEEmulticolD{
	\foo{R}_{\dset{z}}(Q) - \foo{R}_{\dset{z}}(\Pgibbs{P}{Q})
	}
	& = & \lambda( \KL{\Pgibbs{P}{Q}}{Q}+\KL{Q}{\Pgibbs{P}{Q}}),
\end{IEEEeqnarray}
which appeared first in \cite[Corollary~3]{perlaza2024ERMRER}.
Then, substituting~\eqref{Eq_f_KL_LFT_s2} and~\eqref{Eq_f_KL_LFT_dPdQ_beta} into  Lemma~\ref{lemm_CharRzJensen} yields
\begin{IEEEeqnarray}{rCl}
	\IEEEmulticolR{
	\IEEEmulticolD{
	\foo{R}_{\dset{z}}(\Pgibbs{P}{Q}) + \lambda \KL{\Pgibbs{P}{Q}}{Q}
	}}
	& < & -\lambda \exp(-\frac{\foo{R}_{\dset{z}}(Q)+ N_{Q, \dset{z}}(\lambda)}{\lambda}) - N_{Q, \dset{z}}(\lambda)\label{Eq_f_KL_LFTJen_Rz}\\
	& = & \lambda(\!1\! -\! \exp(\!-\frac{\foo{R}_{\dset{z}}(Q)\!+\! K_{Q, \dset{z}}(-\frac{1}{\lambda})}{\lambda}) \!-\!  K_{Q, \dset{z}}(-\frac{1}{\lambda})\!), \qquad 
\end{IEEEeqnarray}
where the function $K_{Q, \dset{z}}$ is defined in~\eqref{Eq_K_cumm}.
This result has been previously reported by several authors in \cite[Lemma 20]{perlaza2024ERMRER} and \cite[Lemma 3]{perlazaISIT2023b}.

\subsection*{Reverse Relative Entropy}
Let the function $f:[0,\infty) \rightarrow \reals$ be such that $f(u) = -\log(u)+u-1$, and note that the Legendre-Fenchel transform of the function $f$ satisfies
\begin{IEEEeqnarray}{rCl}
f^{*}(t) & = & -\log(1-t).
\label{Eq_f_rKL_LFT_s2}
\end{IEEEeqnarray}
Substituting~\eqref{Eq_f_rKL_LFT_s2} into~\eqref{EqPf_ERM_fDR_LT} yields
\begin{IEEEeqnarray}{rCl}
	\IEEEmulticolR{
	\IEEEmulticolD{
	\foo{R}_{\dset{z}}(\Pgibbs{P}{Q})
	}}
	& = & \lambda \KL{Q}{\Pgibbs{P}{Q}} - N_{Q, \dset{z}}(\lambda)
	\splitR \splitD
	-\!\lambda\!\int(\!-\log( \frac{N_{Q, \dset{z}}(\lambda) \!+\! \foo{L}_{\dset{z}}(\thetav)}{\lambda}) \!-1\!)\diff Q(\thetav) 
	\label{Eq_f_rKL_LFT_Rz}\\
	& = & \lambda \KL{Q}{\Pgibbs{P}{Q}} + \lambda - N_{Q, \dset{z}}(\lambda)
	\splitD
	-\lambda\!\int \log( \frac{\lambda}{N_{Q, \dset{z}}(\lambda)\!+\!\foo{L}_{\dset{z}}(\thetav) }) \diff Q(\thetav) \\
	& = & \lambda \KL{Q}{\Pgibbs{P}{Q}} +\lambda - N_{Q, \dset{z}}(\lambda)
	\splitD 
	- \lambda \int \log(\frac{\diff \Pgibbs{P}{Q}}{\diff Q}(\thetav)) \diff Q(\thetav)\\
	& = & \lambda \KL{Q}{\Pgibbs{P}{Q}} \!-\! \lambda \KL{Q}{\Pgibbs{P}{Q}} \!+\!\lambda \!-\! N_{Q, \dset{z}}(\lambda)\qquad
	\label{Eq_f_rKL_LFT_Rz_s4}\\
	& = &  \lambda - N_{Q, \dset{z}}(\lambda).
	\label{Eq_f_rKL_LFT_Rz_s5}
\end{IEEEeqnarray}
This result has been previously reported in~\cite[Lemma 3]{perlazaISIT2023a}, and characterizes the expected empirical risk with only the regularization factor and normalization function.
As in the previous example, from~\eqref{EqTheo_ERM_fDR_LT_sub2} and the assumption that the Legendre-Fenchel transform of $f$ leads to~\eqref{Eq_f_rKL_LFT_s2}, the expected empirical risk induced by the reference measure $Q$ in~\eqref{EqOp_f_ERMRERNormal} satisfies that
\begin{IEEEeqnarray}{rCl}
	\IEEEmulticolD{
	\foo{R}_{\dset{z}}(Q)
	}
	& = & -\lambda\!\int g(\frac{\diff \Pgibbs{P}{Q}}{\diff Q}(\thetav)) \diff \Pgibbs{P}{Q}(\thetav) - N_{Q, \dset{z}}(\lambda)
	\splitR \splitD
	- \lambda\!\int \dot{g}(\frac{\diff \Pgibbs{P}{Q}}{\diff Q}(\thetav)) \diff \Pgibbs{P}{Q}(\thetav)
	\label{Eq_f_rKL_LFT_RzQ_s1}\\
	& = & -\lambda\!\int \frac{\diff Q}{\diff \Pgibbs{P}{Q}}(\thetav)f\!(\frac{\diff \Pgibbs{P}{Q}}{\diff Q}(\thetav)) \diff Q(\thetav) \!-\! N_{Q, \dset{z}}(\lambda)
	\splitR \splitD
	- \lambda\!\int\!\frac{\diff Q}{\diff \Pgibbs{P}{Q}}(\thetav) f^{*}\!(\!-\frac{\foo{L}_{\dset{z}}(\thetav)\!+\!N_{Q, \dset{z}}(\lambda)}{\lambda}\!) \diff Q(\thetav) 
	\label{Eq_f_rKL_LFT_RzQ_s2}\\
	& = & -\lambda\!\int \frac{\diff Q}{\diff \Pgibbs{P}{Q}}(\thetav)\log\!(\!\frac{\diff Q}{\diff \Pgibbs{P}{Q}}(\thetav)\!) \diff Q(\thetav) \!-\! N_{Q, \dset{z}}(\lambda)
	\splitR \splitD
	-\lambda\!\int\!\!\frac{\diff Q}{\diff \Pgibbs{P}{Q}}(\thetav)(\!-\log\!(\!\frac{\foo{L}_{\dset{z}}(\thetav)\!+\!\lambda\!+\!N_{Q, \dset{z}}(\lambda)}{\lambda}\!)
	\splitD[1]
	-1\!\vphantom{\frac{\foo{L}_{\dset{z}}(\thetav)}{\lambda}}) \diff Q(\thetav)
	\label{Eq_f_rKL_LFT_RzQ_s3}\\
	& = & -\lambda\!\int \frac{\diff Q}{\diff \Pgibbs{P}{Q}}(\thetav)\log\!(\!\frac{\diff Q}{\diff \Pgibbs{P}{Q}}(\thetav)\!) \diff Q(\thetav) 
	\splitD
	+\lambda\!\int \frac{\diff Q}{\diff \Pgibbs{P}{Q}}(\thetav)\diff Q(\thetav)
	\splitR \splitD
	+ \lambda\!\int \frac{\diff Q}{\diff \Pgibbs{P}{Q}}(\thetav)\log(\frac{\diff Q}{\diff \Pgibbs{P}{Q}}(\thetav)) \diff Q(\thetav) 
	\splitD 
	- N_{Q, \dset{z}}(\lambda)
	\label{Eq_f_rKL_LFT_RzQ_s4}\\
	& = & \lambda\!\int \frac{\diff Q}{\diff \Pgibbs{P}{Q}}(\thetav)\diff Q(\thetav)- N_{Q, \dset{z}}(\lambda),
	\label{Eq_f_rKL_LFT_RzQ_s5}
\end{IEEEeqnarray}
where~\eqref{Eq_f_rKL_LFT_RzQ_s2} follows from~\eqref{EqDefLFTequal}; and~\eqref{Eq_f_rKL_LFT_RzQ_s3} follows from~\eqref{Eq_f_rKL_LFT_s2}. The importance of~\eqref{Eq_f_rKL_LFT_RzQ_s5} is that it complements the previous work in \cite{perlazaISIT2023a} by providing a new expression to compute the difference between the empirical risk introduced by the measure $\Pgibbs{P}{Q}$ in~\eqref{EqGenpdffDv} and the reference measure $Q$ in terms of the Radon-Nikodym derivative as follows:
\begin{IEEEeqnarray}{rCl}
	\foo{R}_{\dset{z}}(Q)\!-\!\foo{R}_{\dset{z}}(\Pgibbs{P}{Q})\!
	& \!= & \!\lambda\!\!\int\!(\!\frac{\diff Q}{\diff \Pgibbs{P}{Q}}(\thetav) \!-\! 1\!)\diff Q(\thetav), \qquad
	\label{EqRecoverTypeII}
\end{IEEEeqnarray}
which follows from taking the difference between~\eqref{Eq_f_rKL_LFT_Rz_s5} and~\eqref{Eq_f_rKL_LFT_RzQ_s5}.
Note that~\eqref{EqRecoverTypeII} complements the work in \cite{daunas2024TITAsymmetry}, where only~\eqref{Eq_f_rKL_LFT_Rz_s5} was known, and thus highlighting the value of Theorem~\ref{Theo_ERM_fDR_LT} to the existing body of work. 

Then, substituting~\eqref{Eq_f_rKL_LFT_s2} into Lemma~\ref{lemm_CharRzJensen} yields
\begin{IEEEeqnarray}{rCl}
	\IEEEmulticolD{
	\foo{R}_{\dset{z}}(\Pgibbs{P}{Q}) + \lambda \KL{Q}{\Pgibbs{P}{Q}}
	}
	& < & \lambda \log( \frac{\lambda}{\foo{R}_{\dset{z}}(Q)+ N_{Q, \dset{z}}(\lambda)}) +\lambda  
	\splitR 
	- N_{Q, \dset{z}}(\lambda).
	\label{Eq_f_rKL_LFTJen_Rz}
\end{IEEEeqnarray}
The upper bound derived in~\eqref{Eq_f_rKL_LFTJen_Rz}, in conjunction with~\eqref{Eq_f_rKL_LFT_Rz_s5}, can be equivalently expressed as a lower bound on the difference between the expected empirical risks of the reference measure $Q$ and the probability measure $\Pgibbs{P}{Q}$. More specifically,
\begin{IEEEeqnarray}{rCl}
	\IEEEmulticolD{
	\foo{R}_{\dset{z}}(Q) - \foo{R}_{\dset{z}}(\Pgibbs{P}{Q})
	}
	&>& \lambda (\exp(\KL{Q}{\Pgibbs{P}{Q}}) - 1),
\end{IEEEeqnarray}  
where $\foo{R}_{\dset{z}}$ is defined in~\eqref{EqRxy}. This lower bound has been previously established in \cite[Lemma 18]{daunas2024TITAsymmetry} for the specific setting of reverse relative entropy regularization.

\section[Equivalence of the f-Regularization via Transformation of the Empirical Risk]{Equivalence of the $f$-Regularization via Transformation of the Empirical Risk}
\label{sec:equivalence}
This section shows that given two strictly convex and differentiable functions $f$ and $g$ that satisfy the conditions in Definition~\ref{Def_fDivergence}, there exists a function $r:[0,\infty)\rightarrow \reals$, such that the solution to the optimization problem in~\eqref{EqOp_f_ERMRERNormal} is identical to the solution of the following problem:
\begin{IEEEeqnarray}{rcl}
\label{EqOpERMLink_fDiv}
  \min_{P \in \bigtriangleup_{Q}(\set{M})} & \quad & \int r(\foo{L}_{\dset{z}}(\thetav)) \diff P(\thetav) + \lambda\Divf[g]{P}{Q},
\end{IEEEeqnarray}
with $\lambda$ and $Q$ in~\eqref{EqOp_f_ERMRERNormal}.
The main result of this section is presented in the following theorem.
\begin{theorem}
\label{Theo_ERMfDRType1To2}
Let $f$ and $g$ be two strictly convex and differentiable functions satisfying the conditions in Definition~\ref{Def_fDivergence}. If the problem in~\eqref{EqOp_f_ERMRERNormal} possesses a solution, then
\begin{IEEEeqnarray}{rCl}
\IEEEmulticolR{
\IEEEmulticolD{
\arg \min_{P \in \bigtriangleup_{Q}(\set{M})} \int \foo{L}_{\dset{z}}(\thetav) \diff P(\thetav) + \lambda \Divf[f]{P}{Q}
}}
\label{EqDefHfromF2G}
& = & \arg \min_{P \in \bigtriangleup_{Q}(\set{M})} \int r(\foo{L}_{\dset{z}}(\thetav)) \diff P(\thetav)  + \lambda\Divf[g]{P}{Q},\ 
\end{IEEEeqnarray}
where the function $r:[0,\infty) \rightarrow \reals$ is such that
\begin{IEEEeqnarray}{rCl}
\label{EqFromf2gfDR}
r(t)
& = & -\lambda\dot{g}(\dot{f}^{-1}(-\frac{N_{Q, \dset{z}}(\lambda)+t}{\lambda}))-c,
\end{IEEEeqnarray}
with $N_{Q, \dset{z}}$ being the normalization function defined in~\eqref{EqDefNormFunction} for the optimization problem in~\eqref{EqOp_f_ERMRERNormal}, and $c \in \reals$ is a constant.
\end{theorem}
\begin{IEEEproof}
The proof is presented in Appendix~\ref{app_lemm_ERM_fDR_f2g}.	
\end{IEEEproof}

Theorem~\ref{Theo_ERMfDRType1To2} establishes an equivalence between two ERM problems, each subject to a different $f$-divergence regularization.
This equivalence holds whenever the divergences are defined by strictly convex and differentiable functions.
Theorem~\ref{Theo_ERMfDRType1To2} extends the earlier result in \cite[Theorem~3]{daunas2024TITAsymmetry} in two key aspects.
First, it generalizes the result to all strictly convex and differentiable $f$-divergences.
Second, and more importantly, it establishes the equivalence in both directions, whereas the previous result only proved the direction from reverse relative entropy to relative entropy.

The following examples illustrate the equivalence between two $f$-divergence regularizations.
\begin{example}
\label{Ex4_Divf}
Consider the optimization problems in~\eqref{EqOp_f_ERMRERNormal} and~\eqref{EqOpERMLink_fDiv} with $f(u) = -\log(u)+u-1$ and $g(u) = u\log(u)-u+1$, respectively. 
Under the current assumptions, the objective of this example is to demonstrate the equivalence of the solutions to the optimization problems in~\eqref{EqOp_f_ERMRERNormal} and~\eqref{EqOpERMLink_fDiv}.
Denote by $\Pgibbs{\hat{P}}{Q}$ the solution to the optimization problem in~\eqref{EqOpERMLink_fDiv}.
From Theorem \ref{Theo_f_ERMRadNik}, it follows that for all $\thetav \in \supp Q$,
\begin{IEEEeqnarray}{rcl}
\label{Eq_g_example4}
	\frac{\diff \Pgibbs{\hat{P}}{Q}}{\diff Q}(\thetav)
	& = & \frac{\lambda}{r(\foo{L}_{\dset{z}}(\thetav))+ \lambda + \beta},
\end{IEEEeqnarray}
where the function $r$ is defined in~\eqref{EqFromf2gfDR} and under the assumptions of this example satisfies for all $\thetav \in \supp Q$,
\begin{IEEEeqnarray}{rCl}
\IEEEmulticolD{
	r(\foo{L}_{\dset{z}}(\thetav)) 
	}
	& = & \frac{\lambda}{\exp(\!-\frac{\foo{L}_{\dset{z}}(\thetav)}{\lambda} \!-\! \log(\int\!\exp(\!-\frac{\foo{L}_{\dset{z}}(\nuv)}{\lambda})\!\diff Q(\nuv))\!)}\!-\!\lambda\!-\!\beta. \qquad
	\label{Eq_V_example4}
\end{IEEEeqnarray}
Substituting~\eqref{Eq_V_example4} into~\eqref{Eq_g_example4} yields
\begin{IEEEeqnarray}{rcl}
	\frac{\diff \Pgibbs{\hat{P}}{Q}}{\diff Q}(\thetav) & = & \frac{\exp(- \frac{1}{\lambda}\foo{L}_{\dset{z}}(\thetav))}{\int \exp(- \frac{1}{\lambda}\foo{L}_{\dset{z}}(\nuv)) \diff Q (\vect{\nuv})},
\end{IEEEeqnarray}
which is the solution to the optimization problem in~\eqref{EqOp_f_ERMRERNormal} presented in Section \ref{SubSubKL}.
\end{example}

\begin{example}
\label{Ex5_Divf}
Consider the optimization problems in~\eqref{EqOp_f_ERMRERNormal} and~\eqref{EqOpERMLink_fDiv} with $f(u) = u\log(u)-u+1$ and $g(u) = -\log(u)+u-1$, respectively. 
Under the current assumptions, the objective of this example is to demonstrate the converse equivalence for the optimization problems in~\eqref{EqOp_f_ERMRERNormal} and~\eqref{EqOpERMLink_fDiv}.
Denote by $\Pgibbs{\hat{P}}{Q}$ the solution to the optimization problem in~\eqref{EqOpERMLink_fDiv}.
From Theorem \ref{Theo_f_ERMRadNik}, it follows that for all $\thetav \in \supp Q$,
\begin{IEEEeqnarray}{rcl}
\label{Eq_g_example5}
	\frac{\diff \Pgibbs{\hat{P}}{Q}}{\diff Q}(\thetav)
	& = & \frac{\exp(-\frac{r(\foo{L}_{\dset{z}}(\thetav))}{\lambda})}{\int \exp(-\frac{r(\foo{L}_{\dset{z}}(\nuv))}{\lambda}) \diff Q(\nuv)},
\end{IEEEeqnarray}
where the function $r$ is defined in~\eqref{EqFromf2gfDR} and under the assumptions of this example satisfies for all $\thetav \in \supp Q$,
\begin{IEEEeqnarray}{rCl}
r(\foo{L}_{\dset{z}}(\thetav)) & = & \lambda \log(N_{Q, \dset{z}}(\lambda)+\lambda+\foo{L}_{\dset{z}}(\thetav)).
	\label{Eq_V_example5}
\end{IEEEeqnarray}
Substituting~\eqref{Eq_V_example5} into~\eqref{Eq_g_example5} yields
\begin{IEEEeqnarray}{rcl}
	\frac{\diff \Pgibbs{\hat{P}}{Q}}{\diff Q}(\thetav) & = & \frac{\lambda}{\beta +\lambda + \foo{L}_{\dset{z}}(\thetav)},
\end{IEEEeqnarray}
which is the solution to the optimization problem in~\eqref{EqOp_f_ERMRERNormal} presented in Section \ref{SubReverse}.
\end{example}
An important distinction exists between the equivalence result established in Example~\ref{Ex4_Divf} and the log expected empirical risk in \cite{daunas2024TITAsymmetry}. 
While the latter demonstrated equivalence specifically between relative entropy regularization and reverse relative entropy regularization, Theorem~\ref{Theo_ERMfDRType1To2} extends this result to the class of $f$-divergences satisfying Assumptions~\ref{assume:a} and \ref{assume:b} in Theorem~\ref{Theo_f_ERMRadNik}. 
Moreover, the present equivalence holds bidirectionally, representing a significant generalization beyond previous work.

\begin{figure}
\centering
\resizebox{\columnwidth}{!}{
	\begin{tikzpicture}
		\def\radius{3} \def\arcr{10}  \node (eq1) at (0, 0) {{\large Relative Entropy}};
\node (eq2) at (2*\radius-0.05, 0) {{\large Reverse Relative Entropy}};
		
\node (eq3) at (\radius, \radius) {$r(\foo{L}_{\dset{z}}(\thetav))  = \lambda\log(\hat{N}_{Q, \dset{z}}(\lambda) +\lambda+\foo{L}_{\dset{z}}(\thetav))$};
		\node (eq4) at (\radius, -\radius) {$\hat{r}(\foo{L}_{\dset{z}}(\thetav)) \! =\! \frac{\lambda}{\exp(\!-\frac{\foo{L}_{\dset{z}}(\thetav)}{\lambda}\!-\!\log(\!\int \exp(\!-\frac{\foo{L}_{\dset{z}}(\nuv)}{\lambda})\diff Q(\nuv)))}\!-\!\lambda\!-\!\hat{N}_{Q, \dset{z}}(\lambda)$};
		
\draw[->,line width=0.5mm] (eq1.north) to[out=75, in=105] (2*\radius, 0.4);
		\draw[->,line width=0.5mm] (eq2.south) to[out=255, in=285] (0, -0.4);
	\end{tikzpicture}
}
\caption{Representation of the empirical risk transformation from the $f$-divergence induced by $f(u)=-\log(u)+u-1$ and the $g$-divergence induced by $g(u)=u\log(u)-u+1$.}
\end{figure}

\section{Numerical Examples}
\label{sec:subSimExample}

	In machine learning, the generalization error quantifies how well a learning algorithm performs on unseen data~\cite{perlaza2024HAL}.
	More specifically, it reflects how effectively a probability measure (an algorithm) concentrates on models that correctly label test data points.
	Recent work has shown that the sensitivity of the optimization problem in~\eqref{EqOp_f_ERMRERNormal} is closely tied to generalization performance~\cite[Theorem~16]{perlaza2024ERMRER}; see also~\cite{zou2024WorstCase, zou2024generalization, perlaza2024HAL}, and~\cite{aminian2022tighter}.
	Thus, the remaining of this section focuses on the sensitivity of algorithms arising from ERM-$f$DR problems.
	A probability measure $P \in \bigtriangleup_{Q}(\set{M})$, which may represent a learning algorithm, exhibits higher generalization error when it is more sensitive to perturbations in the training data. This increased sensitivity suggests overfitting to the training set. See~\cite{aminian2021exact} and \cite{zou2024generalization}.
	
	To illustrate the impact of different $f$-divergence regularizations in ERM-$f$DR, the problem of classifying two handwritten digits from the MNIST dataset of handwritten digits from 0 to 9~\cite{lecun1998gradient} is studied.
	The dataset consists of $60{,}000$ training images and $10{,}000$ test images. Among these, $12{,}183$ training images and $1{,}986$ test images correspond to the digits six and seven, which form the focus of our binary classification task.
	Each image is a $28 \times 28$ grayscale matrix with pixel values in $[0,1]^{28 \times 28}$. To reduce computational complexity, images are preprocessed following the procedure described in Appendix~\ref{AppNumericalSimulation}.
	The ERM-$f$DR problem defined in~\eqref{EqOp_f_ERMRERNormal} is evaluated, using the following regularization choices: relative entropy (ERM-RER Type-I), reverse relative entropy (ERM-RER Type-II), Jensen-Shannon, and Hellinger divergences, as specified in~\eqref{Eq_f_KL_dPdQ_s2}, \eqref{Eq_f_KL_rdPdQ_s2}, \eqref{Eq_f_JS_dPdQ}, and~\eqref{Eq_f_Hell_dPdQ}. The experimental setup assumes the following:
\begin{enumerate}[label=(\roman*)]
	    \item The model space is defined as $\set{M} = [-50,50]^2$;
	    \item The input space $\set{X}$ consists of HOG (Histogram of Oriented Gradients) feature vectors extracted from the images of digits six and seven, so that $\set{X} \subset \reals^{1296}$;
	    \item The label space is $\set{Y} = \{6,7\}$;
	    \item The reference measure $Q$ is a uniform probability measure on $\set{M}$;
	    \item The loss function $h$ in~\eqref{EqLxy} is defined as
	    		\begin{IEEEeqnarray}{rCl}
				h(\thetav, x) 
				& = & 
				\begin{cases}
					6 & \text{if } 0 < (x\vect{W})\thetav,\\
					7 & \text{if } 0 > (x\vect{W})\thetav, 
				\end{cases}
			\end{IEEEeqnarray}
where the matrix $\vect{W}$ is defined in~\eqref{EqDefWPACsim} in Appendix~\ref{AppNumericalSimulation};
		\item The loss function $\ell$ in~\eqref{EqEll} satisfies
			\begin{IEEEeqnarray}{rCl}
				\ell(h(\thetav, x), y) & = & \ind{h(\thetav, x)\neq y}.
			\end{IEEEeqnarray}
	\end{enumerate}
For the training of each EMR-$f$DR algorithm, $8{,}100$ samples are drawn uniformly at random from the training images to construct the \emph{training dataset} $\dset{z}_1$.
	Likewise, $1{,}300$ samples are drawn from the test images to form the \emph{test dataset} $\dset{z}_2$.
	This sampling procedure is repeated independently over $100$ iterations.
	
\begin{figure}[h]
	\centering
	\begin{minipage}{0.46\linewidth}
\centering
        \includegraphics[width=0.45\linewidth]{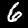}
\caption{ $28\times28$ Image of a handwritten 6 from MNIST dataset.}
	\end{minipage}
\begin{minipage}{0.08\linewidth}	
	\end{minipage}
	\begin{minipage}{0.46\linewidth}
\centering
    	\includegraphics[width=0.45\linewidth]{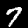}
\caption{ $28\times28$ Image of a handwritten 7 from MNIST dataset.}
	\end{minipage}
	\end{figure}
\begin{figure}[h!]
\centering
        \includegraphics[width=0.8\linewidth]{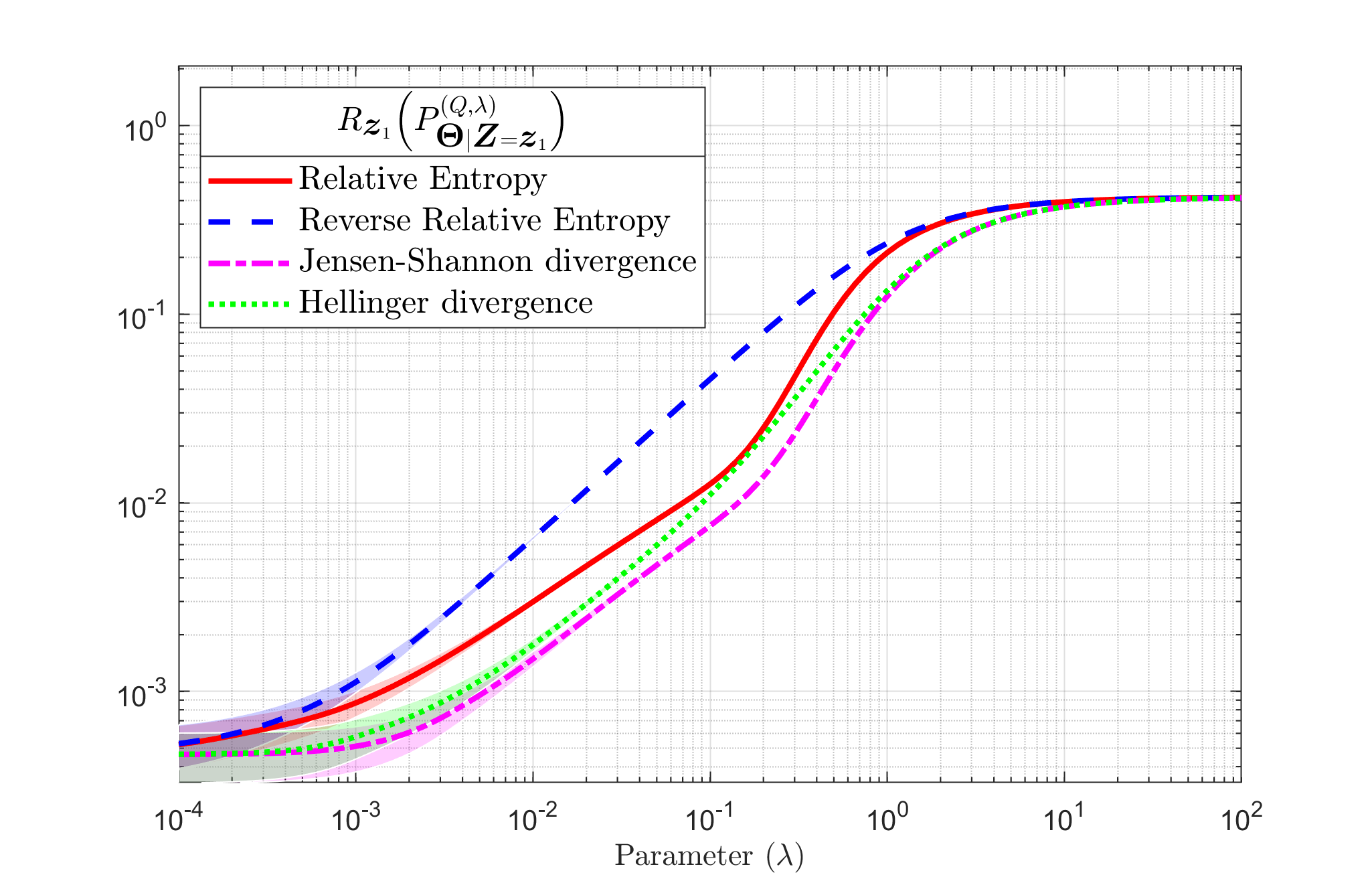}
\caption{
  		Average Training Error: average of the expected empirical risks $\foo{R}_{\dset{z}_1}(\Pgibbs[\dset{z}_1]{P}{Q})$, computed for four different $f$-divergences: relative entropy, reverse relative entropy, Jensen-Shannon divergence, and Hellinger divergence defined in~\eqref{Eq_f_KL_DivEq}, \eqref{Eq_f_KLr_DivEq}, \eqref{Eq_f_JS_DivEq}, and~\eqref{Eq_f_Hell_DivEq}, respectively.
  		The results are the average of 100 random partitions of the training datasets.
  		}
  		\label{FigTrainingPlot}
	\end{figure}
Figure~\ref{FigTrainingPlot} illustrates the training expected risk $\foo{R}_{\dset{z}_1}(\Pgibbs[\dset{z}_1]{P}{Q})$ for different choices of $f$-divergence in the ERM-$f$DR framework, computed over 100 independent iterations.
	Among the $f$-divergences compared, the solution obtained using the reverse relative entropy (ERM-RER Type-II) yields the highest training expected risk.
	This behavior is expected, as reverse KL divergence is known to have heavier tails, based on the work in \cite{daunas2024TITAsymmetry}, which results in a broader allocation of probability mass over $\supp Q$ compared to the classical relative entropy (ERM-RER Type-I) solutions.
	In contrast, the Hellinger and Jensen-Shannon divergences result in significantly lower training risks, even outperforming the ERM-RER Type-I.
	This suggests that both Hellinger and Jensen-Shannon divergences induce more concentrated posteriors around empirical risk minimizers, which leads to tighter fits on the training set.
	An interesting observation is the narrow shaded regions representing the empirical standard deviations across the 100 trials. Their small width across all methods implies that the training expected risk is largely invariant to randomness in the dataset.
	This highlights the dominant role played by the regularization parameter $\lambda$ and the prior $Q$ in the selection of a desired training expected risk for the solution of ERM-$f$DR based algorithms.
	
	\begin{figure}[h!]
		\centering
        \includegraphics[width=0.8\linewidth]{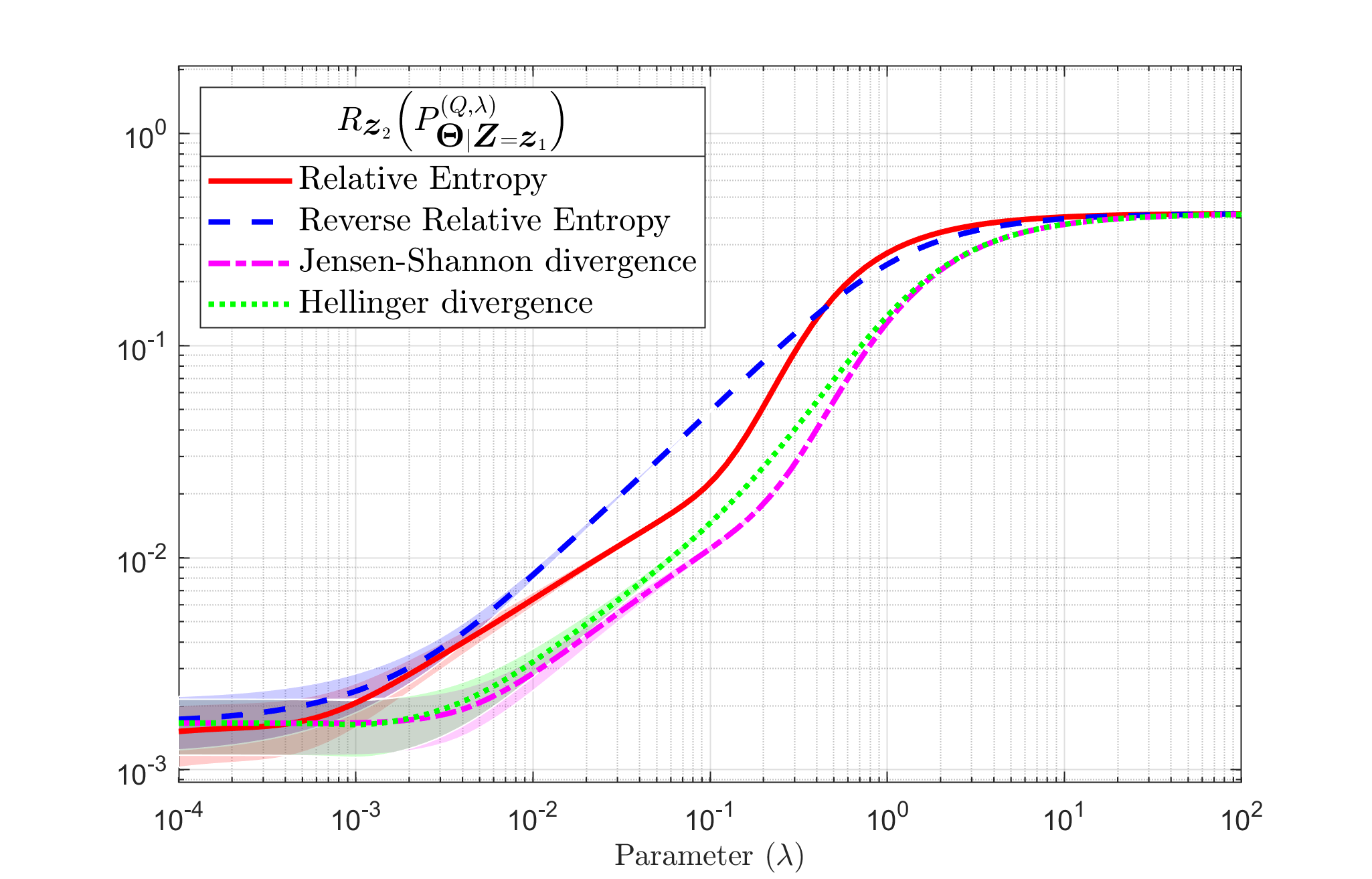}
\caption{Average Test Error: average of the expected empirical risks $\foo{R}_{\dset{z}_2}(\Pgibbs[\dset{z}_1]{P}{Q})$, computed for four different $f$-divergences: relative entropy, reverse relative entropy, Jensen-Shannon divergence, and Hellinger divergence defined in~\eqref{Eq_f_KL_DivEq}, \eqref{Eq_f_KLr_DivEq}, \eqref{Eq_f_JS_DivEq}, and~\eqref{Eq_f_Hell_DivEq}, respectively.
  		The results are the average of 100 random partitions of the training and test datasets.
  		}
  		\label{FigRiskPlot}
	\end{figure}
Figure~\ref{FigRiskPlot} displays the expected test risk $\foo{R}_{\dset{z}_2}(\Pgibbs[\dset{z}_1]{P}{Q})$ for various ERM-$f$DR solutions, computed across 100 independent trials. Each curve corresponds to a different $f$-divergence used in the regularization: reverse relative entropy (Type-II ERM-RER), classical relative entropy (Type-I ERM-RER), Jensen-Shannon, and Hellinger.
	ERM-RER Type-II (Reverse KL) continues to yield the highest test risk across nearly all values of $\lambda$, consistent with its high training expected risk.
	Smilarly, the Jensen-Shannon and Hellinger divergences yield the lowest test risks for most values of $\lambda$.
	This example indicates that the training expected risk is indeed a good descriptor of the performance of an ERM-$f$DR algorithm in the test expected empirical risk.
	This is particularly interesting as it points to the possibility of guaranteeing generalization performance based on training performance.
	However, it is important to point out that for smaller regularization factors, the empirical variance increases, which may indicate that guarantees are also dependent on the choice of regularization factor $\lambda$.
	
\begin{figure}[h!]
		\centering
    	\includegraphics[width=0.8\linewidth]{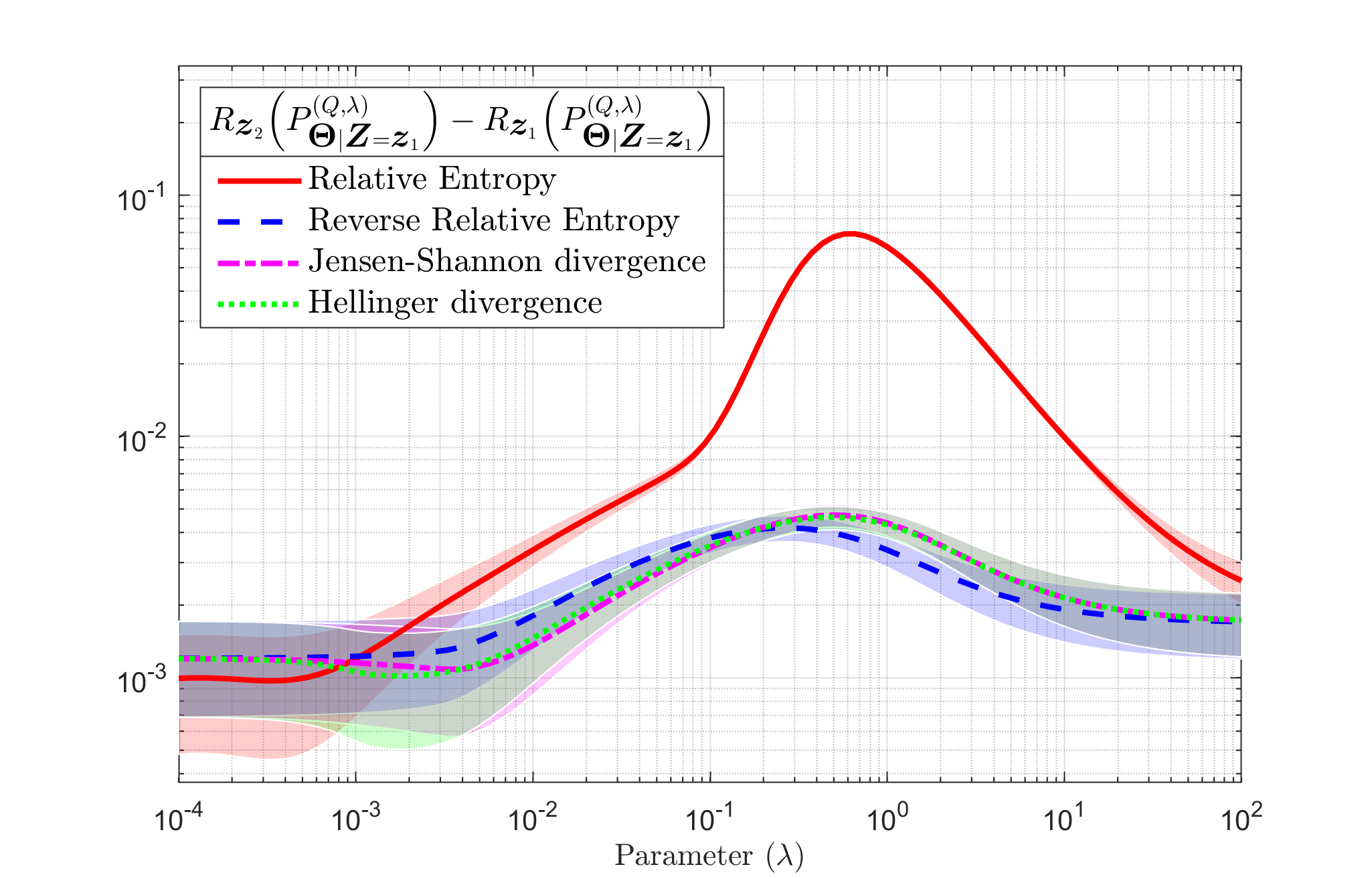}
\caption{Average of the differences $\foo{R}_{\dset{z}_2}(\Pgibbs[\dset{z}_1]{P}{Q})- \foo{R}_{\dset{z}_1}(\Pgibbs[\dset{z}_1]{P}{Q})$, computed for four different $f$-divergences: relative entropy, reverse relative entropy, Jensen-Shannon divergence, and Hellinger divergence defined in~\eqref{Eq_f_KL_DivEq}, \eqref{Eq_f_KLr_DivEq}, \eqref{Eq_f_JS_DivEq}, and~\eqref{Eq_f_Hell_DivEq}, respectively.
  		The results are the average of 100 random partitions of the training and test datasets.
  		}
    	\label{FigSensPlot}
	\end{figure}
Figure~\ref{FigSensPlot} depicts tdifference between the expected test risk and the expected training risk $\foo{R}_{\dset{z}_2}(\Pgibbs[\dset{z}_1]{P}{Q})- \foo{R}_{\dset{z}_1}(\Pgibbs[\dset{z}_1]{P}{Q})$ for various ERM-$f$DR solutions across 100 iterations.
For most of the range of the regularization parameter $\lambda$, the ERM-RER \mbox{Type-I} exhibits a larger sensitivity than ERM-RER \mbox{Type-II}, Jensen-Shannon, and Hellinger.
Notably, the Type-I ERM-RER largest gap is at $\lambda$ around one, indicating a tendency to overfit the training data more significantly compared to the other methods.
This behavior contrasts with the other divergences in this example.
Remarkably, for small values of the regularization factor $\lambda$, the generalization gap of relative entropy is significantly smaller on average than that of reverse relative entropy, Jensen-Shannon, and Hellinger.
It is worth noting that while the test and training expected risks are monotonically decreasing, the generalization gap can take the same value for different regularization factors $\lambda$.
Hence, Figure~\ref{FigSensPlot} indicates that in this numerical analysis, there exist cases in which an appropriate choice of the regularization factor can keep the generalization error small while maintaining good test performance.
Additionally, the three plots indicate that for this particular example, using a uniform reference measure over the set of models for classification is suboptimal; as a result, small values of $\lambda$ are favored, as they rely more on the training data than on the prior.

\section{Conclusions}\label{SecConclusions}

This work has presented the solution to the ERM-$f$DR problem in~\eqref{EqOp_f_ERMRERNormal} and established conditions under which this solution also solves the optimization problem in~\eqref{EqOp_f_ERM_RND2}.
Additionally, the dual problem in~\eqref{EqOp_f_ERMRERDual} has been introduced, along with its solution, which is characterized by the normalization function and exhibits a zero duality gap. 
By proving a zero duality gap, it has been established that the ERM-$f$DR solution depends directly on the Legendre-Fenchel transform.
This connection enables the derivation of an implicit expression for the normalization function in the form of a nonlinear ODE, as well as the properties of the function.
These properties facilitated the design of Algorithm~\ref{Algo_RootFind}, which numerically approximates the normalization function for a given regularization factor.
The algorithm addresses a critical bottleneck in practical applications of the ERM-$f$DR solution: the intractability of analytically solving for the normalization function under general $f$-divergences. 
Without knowing the value of the normalization function, sampling methods such as rejection sampling or Markov Chain Monte Carlo (MCMC) cannot be applied due to the coupling of the log-likelihood ratio (for MCMC) and the need for scaling for suitable proposal distributions (for rejection sampling).
Furthermore, the Legendre-Fenchel transform has been used to characterize the difference between the expected empirical risk of the ERM-$f$DR solution and its reference measure.

These results pave the way for future exploration of nonconvex or nondifferentiable divergences (e.g., Wasserstein or total variation), where the dual optimization remains convex and thus provides lower bounds for the ERM-$f$DR problem.
Another possible direction for future work involves extending these expressions to characterize the difference between the ERM-$f$DR solution and an arbitrary probability measure (beyond the reference). 
Such an extension could enable explicit generalization bounds for ERM-$f$DR solutions using methods analogous to those in \cite{perlaza2024HAL}, as suggested by the equivalence result in Section~\ref{sec:equivalence}.
The equivalence result shows that the solution of an $f$-divergence regularizer can be equal to the solution of a different $f$-divergence regularizer via an appropriate transformation of the empirical risk (see~\eqref{EqFromf2gfDR}).
Additionally, this equivalence between regularizations clarifies the mutual absolute continuity between ERM-$f$DR solutions and the reference measure.
Finally, despite the equivalence, simulations indicate that there are advantages to choosing different $f$-divergence regularizers, though the conditions favoring one choice over another remain an open question. 
\IEEEtriggeratref{85} \bibliography{references.bib}
\newpage
\appendices
\onecolumn
\section{Preliminaries}
\label{sec:AppendixA}

\begin{lemma}
\label{lemm_canformFDiv}
Let $f:[0,\infty)\rightarrow \reals$ be a differentiable function inducing an $f$-divergence and satisfying $\dot{f}(1)\neq 0$. Then there exists a function 
$\tilde{f}:[0,\infty)\rightarrow \reals$ such that $\dot{\tilde{f}}(1)=0$ and, for all probability measures $P$ and $Q$, it holds that
\begin{equation}
\label{EqTiltedEquality}
\Divf{P}{Q}	=\Divf[\tilde{f}]{P}{Q}.
\end{equation}
\end{lemma}
\begin{IEEEproof}
Let the function	 $f:[0,\infty)\rightarrow \reals$ be differentiable and the function $\tilde{f}:[0,\infty)\rightarrow \reals$ be defined as
\begin{equation}
\label{EqfTilted}
	\tilde{f}(u) = f(u)-\dot{f}(1)(u-1).
\end{equation}
Note that, from \eqref{EqfTilted} it follows that
\begin{IEEEeqnarray}{rCl}
\label{EqfTiltedAt1}
	\tilde{f}(1) & = & f(1)-\dot{f}(1)(0)\\
	& = & 0.
\end{IEEEeqnarray}
Similarly, the derivative of \eqref{EqfTilted} yields,
\begin{equation}
	\dot{\tilde{f}}(u) = \dot{f}(u)-\dot{f}(1),
\end{equation}
which implies that
\begin{equation}
	\dot{\tilde{f}}(1) = 0.
\end{equation}
The proof continues by showing that the equality in \eqref{EqTiltedEquality} holds. 
From the definition of $f$-divergences, it follows that
\begin{IEEEeqnarray}{rCl}
 \Divf[\tilde{f}]{P}{Q}
 	& = & \int \tilde{f}(\frac{\diff P}{\diff Q}(\thetav))\diff Q(\thetav) \label{EqProofTiltEqual_s1}\\
 	& = &  \int f(\frac{\diff P}{\diff Q}(\thetav))- \dot{f}(1)(\frac{\diff P}{\diff Q}(\thetav)-1)\diff Q(\thetav)\label{EqProofTiltEqual_s2}\\
 	& = & \int f(\frac{\diff P}{\diff Q}(\thetav))\diff Q(\thetav)-\dot{f}(1)\int(\frac{\diff P}{\diff Q}(\thetav)-1)\diff Q(\thetav)\label{EqProofTiltEqual_s3}\\
 	& = & \Divf{P}{Q} - \dot{f}(1)(\int \diff P(\thetav)-\int \diff Q(\thetav))\label{EqProofTiltEqual_s4}\\
 	& = & \Divf{P}{Q},\label{EqProofTiltEqual_s5}
\end{IEEEeqnarray}
where \eqref{EqProofTiltEqual_s2} follows from \eqref{EqfTilted}; \eqref{EqProofTiltEqual_s3} follows from the change of measure; and \eqref{EqProofTiltEqual_s4} follows from the fact that both $P$ and $Q$ are probability measures.
This completes the proof of \eqref{EqTiltedEquality}.
\end{IEEEproof}

\begin{lemma}
\label{lemm_convxdotf}
Let $P$ and $Q$ be two probability measures on the measurable space $\msblspc{\set{M}}$. Let $f:[0,\infty)\rightarrow \reals$ be a twice differentiable and strictly convex function that induces a finite $f$-divergence and satisfies $\dot{f}(1)=0$. Then,
\begin{equation}
\label{EqIneqdotfdP}
\int \dot{f}(\frac{\diff P}{\diff Q}(\thetav)) \diff P(\thetav) \geq 0.
\end{equation}
\end{lemma}
\begin{IEEEproof}
From the $f$-divergence definition, $P$ is absolutely continuous with respect to $Q$. Consequently, the Radon-Nikodym derivative $u:\set{M}\to[0,\infty)$ exists, is $Q$-measurable and satisfies \mbox{$\int u(\thetav)\,\diff Q(\thetav)=1$}. Moreover, by the assumptions that $f$ is twice differentiable and strictly convex, it follows that $\ddot{f}\ge0$ on $(0,\infty)$. Finally, from the assumption that $D_f(P\|Q)<\infty$, the composition \(f\circ u\) is \(Q\)-integrable.

The first step is as follows. From the strict convexity of $f$ and the condition $f(1)=0$, for all $t>0$, the convexity and differentiability of $f$ yield the tangent-line inequality such that, for all $s>0$
\begin{equation}
f(s)\geq f(t) + \dot f(t)\,(s-t),
\end{equation}
with strict inequality for $s\neq t$.
Let $s=1$ and use the fact that $f(1)=0$ to obtain, for all $t>0$,
\begin{equation}\label{eq:tangent}
\dot{f}(t)(t-1)\geq f(t).
\end{equation}
Replacing $t$ by $u(\thetav)$ and integrating with respect to $Q$ yields
\begin{equation}\label{Eq_step1_int}
\int \dot f\big(u(\thetav)\big)\big(u(\thetav)-1\big)\,\diff Q(\thetav)
\geq \int f\big(u(\thetav)\big)\,\diff Q(\thetav)
= \Divf{P}{Q}.
\end{equation}

The second step is as follows. Consider the partition of the set $[0,\infty)$ formed by the sets $\set{A}_0$ and $\set{A}_1$, which satisfy the following:
\begin{IEEEeqnarray}{rCl}
A_0 & \triangleq & [0,1),\\
A_1 & \triangleq & [1,\infty).
\end{IEEEeqnarray}
For $t\in \set{A}_1$ the Fundamental Theorem of Calculus together with $f(1)=\dot{f}(1)=0$ give
\begin{align}
\dot f(t) &= \int_{1}^{t} \ddot{f}(s) \diff s, \label{Eq_dotf_case1}\\
f(t) &= \int_{1}^{t} (t-s)\,\ddot{f}(s) \diff s. \label{Eq_f_case1}
\end{align}
For $t\in \set{A}_0$ the same conditions give
\begin{align}
\dot f(t) &= -\int_{t}^{1} \ddot{f}(s) \diff s, \label{Eq_dotf_case0}\\
f(t) &= \int_{t}^{1} (s-t)\,\ddot{f}(s) \diff s. \label{Eq_f_case0}
\end{align}
Adding \eqref{Eq_dotf_case1} and \eqref{Eq_f_case1}, and \eqref{Eq_dotf_case0} and \eqref{Eq_f_case0} for each set $\set{A}_0$ and $\set{A}_1$ yields the representation
\begin{equation}\label{Eq_FTC_rep}
f(t)+\dot f(t)
=
\begin{cases}
\displaystyle \int_{1}^{t} (t-s+1)\,\ddot{f}(s) \diff s, & t\in A_1,\\
\displaystyle \int_{t}^{1} (t-s+1)\,\ddot{f}(s) \diff s, & t\in A_0.
\end{cases}
\end{equation}
Observe that for all $t \in [0,\infty)$, the integrand term $(t-s+1)\ddot{f}(s)$ is nonnegative. Indeed, since $\ddot{f}(s)\geq 0$, it suffices to verify that $t-s+1\geq 0$. If $t\geq 1$ and $s\in[1,t]$, then $t-s+1\geq 0$. If $t<1$ and $s\in[t,1]$, then $t-s+1\geq t>0$. Therefore, $t-s+1\geq 0$ holds throughout the interval $[0,\infty)$. Using this, and the fact that $t\geq 0$, \eqref{Eq_FTC_rep} can be expressed as follows,
\begin{equation}\label{Eq_FTC_rep1}
f(t)+\dot f(t) = \displaystyle \int_{0}^{\infty} \ind{s: (s-1)(t-1)>0}(t-s+1)\,\ddot{f}(s) \diff s
\end{equation}

The third step is as follows. The right-hand side of \eqref{Eq_FTC_rep1} can be written as an integral over the product space $\set{M}\times[0,\infty)$ of the nonnegative measurable function
\begin{IEEEeqnarray}{rCl}
\IEEEmulticolR{
\IEEEmulticolD{
\int \bigl(f(u(\thetav))+\dot f(u(\thetav))\bigr)\diff Q(\thetav)
}}
& = & \int (\int(u(\thetav)-s+1)\ddot{f}(s)\ind{(s-1)(u(\thetav)-1)>0}\diff s) \diff Q(\thetav)\label{Eq_swap_1}\\
& = & \int (\int(u(\thetav)-s+1)\ind{\thetav\in \set{M}: (s-1)(u(\thetav)-1)>0}\diff Q(\thetav)) \ddot{f}(s)\diff s.\label{Eq_swap_2}
\end{IEEEeqnarray}
where \eqref{Eq_swap_2} follows from Tonelli's theorem \cite[Theorem~2.5.3]{ash2000probability}, which follows by observing that the set $\{\thetav\in \set{M}:(s-1)\right.$ $\left.(u(\thetav)-1)>0\}$ is measurable as a result of $u$ being measurable.
For fixed $s>0$, define the measurable subset
\begin{IEEEeqnarray}{rCl}
\set{A}_s & \triangleq & \{\thetav\in \set{M}: (s-1)(u(\thetav)-1)>0\}.
\end{IEEEeqnarray}
Then
\begin{IEEEeqnarray}{rCl}
\int_{\set{A}_s} (u(\thetav)-s+1) \diff Q(\thetav)
	& = & \int_{\set{A}_s} (u(\thetav)-s) \diff Q(\thetav) + Q(\set{A}_s)
	 \geq 0.\label{EqPosInerInt}
\end{IEEEeqnarray}
From the strict convexity of $f$, it follows that $\ddot{f}(s)\geq 0$ for all $s$. Together with \eqref{EqPosInerInt}, this implies that the integral in \eqref{Eq_swap_2} is nonnegative for all $s \geq 0$. This yields
\begin{equation}
\label{Eq_step3_conclude}
\int \bigl(f(u(\thetav))+\dot{f}(u(\thetav))\bigr)\diff Q(\thetav) \geq 0,
\end{equation}
or equivalently
\begin{IEEEeqnarray}{rCl}
\label{Eq_dot_bound}
\int \dot{f}(u(\thetav))\diff Q(\thetav) 
& \geq & -\int f(u(\thetav))\diff Q(\thetav) \\
&   =  &  -\Divf{P}{Q}.
\end{IEEEeqnarray}
The proof continues by using the algebraic identity
\begin{equation}
 u(\thetav)\dot{f}(u(\thetav))
= \dot{f}(u(\thetav)) + \dot{f}(u(\thetav))( u(\thetav)-1),
\end{equation}
which together with \eqref{Eq_step1_int} and \eqref{Eq_dot_bound} imply that 
\begin{IEEEeqnarray}{rCl}
\!\!\!\!\!\int u(\thetav)\dot{f}(u(\thetav))\diff Q(\thetav)
 & = & \int\dot{f}(u(\thetav))( u(\thetav)-1)\diff Q(\thetav) +\int \dot{f}(u(\thetav))\diff Q(\thetav) \\
 & \geq & \Divf{P}{Q}  +\int \dot{f}(u(\thetav))\diff Q(\thetav)\\
 & \geq & \Divf{P}{Q} - \Divf{P}{Q}\\
 & = & 0. \label{Eq_Term2IsPoss}
\end{IEEEeqnarray}
It follows from \eqref{Eq_Term2IsPoss}, that
\begin{IEEEeqnarray}{rCl}
 	\int \dot{f}(\frac{\diff P}{\diff Q}(\thetav)) \diff P(\thetav) & \geq & 0,
\end{IEEEeqnarray}
which completes the proof.
\end{IEEEproof}

\begin{theorem}
\label{Theo_ERM_fDR_Leibniz}
Given a probability measure space $\mspc{\set{M}}{P}$	and an open subset $\set{A}\subseteq\reals$, let the function $f: \set{A}\times\set{M} \rightarrow \reals$ be measurable with respect to $\msblspc{\set{A}\times\set{M}}$ and $\bormsblspc{\reals}$. If for all $\nuv \in \set{M}$, the function $f(\cdot,\nuv):\set{A} \rightarrow \reals$ is \Lipschitz continuous and for some $s \in \set{A}$, $\int f(s,\nuv) \diff P(\nuv) < \infty$, then
\begin{IEEEeqnarray}{rCl}
\label{EqLeibnizInt}
	\left.\frac{\diff }{\diff t} \int f(t, \nuv) \diff P(\nuv)\right|_{t = s}
	& = & \left.\int \frac{\diff }{\diff t} f(t, \nuv)\right|_{t = s} \diff P(\nuv),
\end{IEEEeqnarray}
\end{theorem}
\begin{IEEEproof}
Note that
\begin{IEEEeqnarray}{rCl}
	\left.\frac{\diff }{\diff t} \int f(t, \nuv) \diff P(\nuv)\right|_{t = s}
	& = & \lim_{\delta \rightarrow 0} \frac{\int f(s + \delta, \nuv) \diff P(\nuv)-\int f(s, \nuv) \diff P(\nuv)}{\delta}
	\label{EqLeibnizLim_s1}\\
	& = & \lim_{\delta \rightarrow 0} \int \frac{f(s+\delta, \nuv) - f(s, \nuv) }{\delta} \diff P(\nuv),
	\label{EqLeibnizLim_s2}
\end{IEEEeqnarray}
where~\eqref{EqLeibnizLim_s2} follows from \cite[Theorem 1.6.3]{ash2000probability}. The assumption that for all $\nuv \in \set{M}$, the function $f(\cdot,\nuv)$ is \Lipschitz continuous implies that for all $s \in \set{A}$ and some $\delta \in \reals$ and $L \in \reals$,
\begin{IEEEeqnarray}{rCl}
	\abs{f(s + \delta, \nuv) - f(s, \nuv)}
	& < & L \abs{\delta}.
	\label{EqLeibnizIsLipschitz_s1}
\end{IEEEeqnarray}
And thus, dividing the right hand side and left hand side of~\eqref{EqLeibnizIsLipschitz_s1} by $\abs{\delta}$ yields
\begin{IEEEeqnarray}{rCl}
	\abs{\frac{f(s + \delta, \nuv) - f(s, \nuv)}{\delta}}
	& < & L ,
	\label{EqLeibnizIsLipschitz_s2}\\
\end{IEEEeqnarray}
which implies that
\begin{IEEEeqnarray}{rCl}
	\int \abs{\frac{f(s + \delta, \nuv) - f(s, \nuv)}{\delta}} \diff P(\nuv)
	& < & \infty.
	\label{EqLeibnizIsLipschitz_s3}\\
\end{IEEEeqnarray}
This allows using the dominated convergence theorem \cite[Theorem 1.6.9]{ash2000probability} as follows. From~\eqref{EqLeibnizLim_s2}, the following holds
\begin{IEEEeqnarray}{rCl}
	\left.\frac{\diff }{\diff t} \int f(t, \nuv) \diff P(\nuv)\right|_{t = s}
	& = & \lim_{\delta \rightarrow 0} \int \frac{ f(s + \delta, \nuv) - f(s, \nuv) }{\delta} \diff P(\nuv)
	\label{EqLeibnizLim2_s1}\\
	& = & \int  \lim_{\delta \rightarrow 0} \frac{ f(s + \delta, \nuv) - f(s, \nuv) }{\delta} \diff P(\nuv)
	\label{EqLeibnizLim2_s2}\\
	& = & \left.\int \frac{\diff }{\diff t} f(t, \nuv)\right|_{t = s} \diff P(\nuv),
\end{IEEEeqnarray}
where~\eqref{EqLeibnizLim2_s2} follows from the dominated convergence theorem \cite[Theorem 1.6.9]{ash2000probability}. This completes the proof.
\end{IEEEproof}

\begin{lemma}
\label{lemm_ERM_fDR_DiffZero}
Let $\field{M}$ be the set of measurable functions that maps $\set{M} \rightarrow \reals$, with respect to the measurable space $\msblspc{\set{M}}$ and $\bormsblspc{\reals}$. Let $\field{S}$ be the subset of $\field{M}$, including all nonnegative functions that are absolutely integrable with respect to a probability measure $Q$. That is, for all $h \in \field{S}$, it holds that
\begin{IEEEeqnarray}{rCl}
\int \abs{h(\thetav)}\diff Q(\thetav) & < & \infty.
\end{IEEEeqnarray}
Given a strictly convex function $f:\reals \rightarrow \reals$, let the function $\hat{r}: \reals \rightarrow \reals$ be such that
\begin{IEEEeqnarray}{rCl}
\label{EqHatR_lemmAppx}
	\hat{r}(\alpha) & = & \int f(g(\thetav) + \alpha h(\thetav)) \diff Q(\thetav),
\end{IEEEeqnarray}
for some function $g$ and $h$ in $\field{S}$ and $\alpha \in (-\epsilon, \epsilon)$, with $\epsilon > 0$ arbitrarily small. Then, the function $\hat{r}$ in~\eqref{EqHatR_lemmAppx} is differentiable at zero.
\end{lemma}
\begin{IEEEproof}
The objective is to prove that the function $\hat{r}$ in~\eqref{EqHatR_lemmAppx} is differentiable at zero, which boils down to proving that the limit
\begin{IEEEeqnarray}{rCl}
\label{EqLimf}
	& \lim_{\delta \rightarrow 0} \frac{1}{\delta}(\hat{r}(\alpha+\delta) -\hat{r}(\alpha))&
\end{IEEEeqnarray}
exists for $\alpha \in (-\epsilon, \epsilon)$, with $\epsilon > 0$ arbitrarily small. 
The proof of the existence of the limit in~\eqref{EqLimf} relies on the fact that the function $f$ in~\eqref{EqHatR_lemmAppx} is strictly convex and differentiable, which implies that $f$ is also Lipschitz continuous. Hence, it follows that
\begin{IEEEeqnarray}{rCl}
\label{EqfisLipschitz}
	\abs{f(g(\thetav) + (\alpha+\delta) h(\thetav))-f(g(\thetav) + \alpha h(\thetav))} & \leq & c\abs{h(\thetav)}\abs{\delta},
\end{IEEEeqnarray}
for some positive and finite constant $c$, which implies that
\begin{IEEEeqnarray}{rCl}
\label{EqDfDxLipschitz}
	\frac{\abs{f(g(\thetav) + (\alpha+\delta) h(\thetav))-f(g(\thetav) + \alpha h(\thetav))}}{\abs{\delta}} & \leq & c\abs{h(\thetav)},
\end{IEEEeqnarray}
and thus, given that $g \in \field{S}$, integrating both sides of~\eqref{EqDfDxLipschitz} with respect to $Q$ yields,
\begin{IEEEeqnarray}{rCl}
\label{EqLimitExists}
	\int \frac{\abs{f(g(\thetav) + (\alpha+\delta) h(\thetav))-f(g(\thetav) + \alpha h(\thetav))}}{\abs{\delta}} \diff Q(\thetav) & \leq & \infty.
\end{IEEEeqnarray}
This allows using the dominated convergence theorem as follows. From the fact that the function $f$ is differentiable, let $\dot{f}:(0,\infty) \rightarrow \reals$ be the first derivative of $f$. The limit in~\eqref{EqLimf} satisfies for all $\alpha \in (-\epsilon,\epsilon)$, with $\epsilon > 0$ arbitrarily small,

\begin{IEEEeqnarray}{rCl}
	\IEEEmulticolR{
	\IEEEmulticolD{
	\!\!\lim_{\delta \rightarrow 0} \frac{1}{\delta}(\hat{r}(\alpha+\delta) -\hat{r}(\alpha)) 
	}}
	& = & \lim_{\delta \rightarrow 0} \frac{1}{\delta}(\int f(g(\thetav) + (\alpha+\delta) h(\thetav)) \diff Q(\thetav)\!-\!\int f(g(\thetav) + \alpha h(\thetav)) \diff Q(\thetav))\ 
	\label{EqLimHatrDCT_s1}\\
	& = & \lim_{\delta \rightarrow 0} \int \frac{1}{\delta}(f(g(\thetav) + (\alpha+\delta) h(\thetav)) - f(g(\thetav) + \alpha h(\thetav))) \diff Q(\thetav)
	\label{EqLimHatrDCT_s2}\\
	& = & \int \lim_{\delta \rightarrow 0} \frac{1}{\delta}(f(g(\thetav) + (\alpha+\delta) h(\thetav)) - f(g(\thetav) + \alpha h(\thetav))) \diff Q(\thetav)
	\label{EqLimHatrDCT_s3}\\
	& = & \int \dot{f}(g(\thetav) + (\alpha+\delta) h(\thetav))  \diff Q(\thetav)
	\label{EqLimHatrDCT_s4}\\
	& < &  \infty,
	\label{EqLimHatrDCT_s5}
\end{IEEEeqnarray}
where the equalities in~\eqref{EqLimHatrDCT_s3} and~\eqref{EqLimHatrDCT_s5} follow from the dominated convergence theorem \cite[Theorem 1.6.9]{ash2000probability}. From~\eqref{EqLimHatrDCT_s5}, it follows that the function $\hat{r}$ in~\eqref{EqHatR_lemmAppx} is differentiable at zero. This completes the proof.
\end{IEEEproof}

\begin{lemma}
\label{lemm_f_invIsInc}
Given a strictly convex and differentiable function $f:\set{I}\rightarrow \reals$, the inverse of the derivative of $f$, denoted by the function $\dot{f}^{-1}:\set{J} \rightarrow \set{I}$, exists and is strictly increasing.
\end{lemma}
\begin{IEEEproof}
Under the assumption that the function $f:\set{I}\rightarrow \reals$ is strictly convex and differentiable, it follows from the definition of strict convexity (see \cite[Definition 7.8]{luenberger1997bookOptimization}) that its derivative $\dot{f}: \set{I} \rightarrow \set{J}$ is strictly increasing and continuous.
By the continuous inverse theorem (see \cite[Theorem 5.6]{bartle2000introduction}), the inverse function $\dot{f}^{-1}:\set{J} \rightarrow \set{I}$ exists and is also strictly increasing and continuous.
This completes the proof.
\end{IEEEproof}

\begin{lemma}
\label{lemm_f_NormInv}
Given a strictly convex and twice differentiable function $f:\set{I}\rightarrow \reals$, and a differentiable function $h:\set{I} \rightarrow \set{J}$, for all $u \in \set{I}$ it holds that
\begin{IEEEeqnarray}{rCl}
 \frac{\diff }{\diff u}\dot{f}^{-1}(h(u))& = & \frac{\dot{h}(u)}{\ddot{f}(\dot{f}^{-1}(h(u)))}.
\end{IEEEeqnarray}

\end{lemma}
\begin{IEEEproof}
Let the function $g:\set{I}\rightarrow \set{J}$ be defined for all $u \in \set{I}$ by 
\begin{IEEEeqnarray}{rCl}
\label{EqDefgasInv}
g(u) & = & \dot{f}^{-1}(h(u))
\end{IEEEeqnarray}
By the definition of the inverse function, it follows that
\begin{IEEEeqnarray}{rCl}
\label{EqInvofInvIsX}
\dot{f}(g(u))& = & h(u).	
\end{IEEEeqnarray}
Differentiating~\eqref{EqInvofInvIsX} with respect to $u$ yields
\begin{IEEEeqnarray}{rCl}
\label{EqDiffInvofInvIs1}
\frac{\diff }{\diff u}\dot{f}(g(u))
& = & \ddot{f}(g(u))\,\dot{g}(u)\\
& = & \dot{h}(u).	
\end{IEEEeqnarray}
From~\eqref{EqDiffInvofInvIs1}, the derivative of the function $g$ in~\eqref{EqInvofInvIsX} is given by
\begin{IEEEeqnarray}{rCl}
\dot{g}(u)
& = & \frac{\dot{h}(u)}{\ddot{f}(g(u))}\label{EqDiffg} \\
& = & \frac{\dot{h}(u)}{\ddot{f}(\dot{f}^{-1}(h(u)))},\label{EqDiffg_s2}
\end{IEEEeqnarray}
where~\eqref{EqDiffg_s2} follows from~\eqref{EqDefgasInv}. Hence, \eqref{EqDefgasInv} and~\eqref{EqDiffg_s2} imply that
\begin{IEEEeqnarray}{rCl}
\frac{\diff }{\diff u}\dot{f}^{-1}(h(u))
& = & \frac{\dot{h}(u)}{\ddot{f}(\dot{f}^{-1}(h(u)))}.
\end{IEEEeqnarray}
This completes the proof.
\end{IEEEproof}

\begin{lemma}
\label{lemm_f_LFTequality}
	The \emph{Legendre-Fenchel} transform of a strictly convex and differentiable function $f$, satisfies for all $v \in \set{J}$, with $\set{J}$ in~\eqref{EqDefJinLFT}, 
\begin{IEEEeqnarray}{rCl}
	f^{*}(v) & = &  v\dot{f^{*}}(v)- f\big(\dot{f^{*}}(v)\big).
\end{IEEEeqnarray}
\end{lemma}
\begin{IEEEproof}
From the \emph{Legendre-Fenchel} transform in Definition~\ref{DefLT_cnvxcnj}, it holds that for all $v \in \set{J}$, with $\set{J}$ in~\eqref{EqDefJinLFT},
\begin{IEEEeqnarray}{rCl}
\label{EqpfDefLFT}
   f^*(v) & = & \sup_{u \in \set{I}} \left( vu  - f(u) \right),
\end{IEEEeqnarray}
which implies that, for all $u \in \set{I}$,
\begin{IEEEeqnarray}{rCl}
\label{EqpfFenchelIneq}
   f^*(v) & \geq &  vu - f(u).
\end{IEEEeqnarray}
Note that~\eqref{EqpfFenchelIneq} holds with equality if $u$ achieves the supremum in~\eqref{EqpfDefLFT}, which will be denoted by $u_v$. In other words, $u_v$ is the maximizing argument
\begin{IEEEeqnarray}{rCl}
\label{EqpfMaxArgLFT}
   u_v & = & \arg \max_{u \in \set{I}}\ uv-f(u),
\end{IEEEeqnarray}
such that
\begin{IEEEeqnarray}{rCl}
\label{EqpfMaxLFT_uv}
    f^{*}(v)  & = &  v u_v - f(u_v).
\end{IEEEeqnarray}
Note that under Assumption~\ref{assume:a} in Theorem~\ref{Theo_f_ERMRadNik}, the maximization problem 
\begin{IEEEeqnarray}{rCl}
\max_{u \in \set{I}} & \ & uv-f(u),
\end{IEEEeqnarray}
is concave and possesses a unique solution that satisfies
\begin{IEEEeqnarray}{rCCCl}
\left.\frac{\diff}{\diff u} (uv-f(u))\right|_{u = u_v} & = & v - \dot{f}(u_v)
& = & 0, \label{EqpfZeroLFT_Dfproof}
\end{IEEEeqnarray}
and thus,
\begin{IEEEeqnarray}{rCl}
u_v & = & \dot{f}^{-1}(v)\\
	& = & \dot{f^{*}}(v),
	\label{EqpfSolutionDfDt}
\end{IEEEeqnarray}
where~\eqref{EqpfSolutionDfDt} follows from \cite[Corollary~23.5.1]{rockafellar1970conjugate}.
Hence, substituting~\eqref{EqpfSolutionDfDt} into \eqref{EqpfMaxLFT_uv} yields that for all $v \in \set{J}$,
\begin{IEEEeqnarray}{rCl}
	f^{*}(v) & = & v\dot{f^{*}}(v)- f\big(\dot{f^{*}}(v)\big),
\end{IEEEeqnarray}
which completes the proof.
\end{IEEEproof}

\begin{lemma}
\label{lemm_UpperboundRQ}
Let $P$ and $Q$ be probability measures on the measurable space $\msblspc{\set{M}}$, such that $P$ and $Q$ are mutually absolutely continuous. Under the assumption that the Radon-Nikodym derivative $\frac{\diff P}{\diff Q}$ is strictly monotonic with respect to the empirical risk $\foo{L}_{\dset{z}}$ defined in~\eqref{EqLxy}. Then the corresponding expected empirical risks satisfy
\begin{equation}
    \foo{R}_{\dset{z}}(Q) < \foo{R}_{\dset{z}}(P),
\end{equation}
in the case in which $\frac{\diff P}{\diff Q}$ is strictly monotonically decreasing with respect to $\foo{L}_{\dset{z}}$, and
\begin{equation}
    \foo{R}_{\dset{z}}(Q) > \foo{R}_{\dset{z}}(P),
\end{equation}
in the case in which $\frac{\diff P}{\diff Q}$ is strictly monotonically increasing with respect to $\foo{L}_{\dset{z}}$.
\end{lemma}
\begin{IEEEproof}
The proof treats the case in which $\frac{\diff P}{\diff Q}$ is strictly increasing; the case of strict decrease follows by the same argument with the inequalities reversed.
Denote the Radon-Nikodym derivative $\frac{\diff P}{\diff Q}$ by the function $g:[0,\infty)\to [0,\infty)$, such that for all $\thetav \in \supp Q$,
\begin{IEEEeqnarray}{rCl}
\label{EqTemDeffRNDg}
	\frac{\diff P}{\diff Q}(\thetav) & = &  g(\foo{L}_{\dset{z}}(\thetav)),
\end{IEEEeqnarray}
Let the real value $k \in \reals$ be defined as 
\begin{IEEEeqnarray}{rCl}
\label{EqRNdPdQat1}
	k & = &  {g}^{-1}(1),
\end{IEEEeqnarray}
and consider the partition of the set $\set{M}$ formed by the sets $\set{M}_0$ and $\set{M}_1$, which satisfy the following:
\begin{IEEEeqnarray}{rCl}
	\set{M}_0 & = & \{\thetav \in \set{M}: \foo{L}_{\dset{z}}(\thetav)  <  k \},
	\label{EqproofSetM0}\\
	\set{M}_1 & = & \{\thetav \in \set{M}: \foo{L}_{\dset{z}}(\thetav)\geq k \}.
	\label{EqproofSetM1}
\end{IEEEeqnarray}
Note that from~\eqref{EqRNdPdQat1} for all $\thetav \in \set{M}$, such that $\foo{L}_{\dset{z}}(\thetav) = k$, it follows
\begin{IEEEeqnarray}{rCl}
\label{EqgStarIs1}
 	g(\foo{L}_{\dset{z}}(\thetav)) = 1.
\end{IEEEeqnarray}
The monotonicity of $g$ and \eqref{EqgStarIs1} imply that for all $\thetav \in \set{M}_0$,
\begin{IEEEeqnarray}{rCl}
\label{EqgStarIsLess1}
 	g(\foo{L}_{\dset{z}}(\thetav)) < 1,
\end{IEEEeqnarray}
and for all $\thetav \in \set{M}_1$,
\begin{IEEEeqnarray}{rCl}
\label{EqgStarIsMore1}
 	g(\foo{L}_{\dset{z}}(\thetav)) \geq 1,
\end{IEEEeqnarray}
Let $P$ denote the probability measure defined by the function $g$ in \eqref{EqTemDeffRNDg}, such that
\begin{IEEEeqnarray}{rCl}
 P(\set{M}) & = & \int_{\set{M}_0} g(\foo{L}_{\dset{z}}(\thetav)) \diff Q(\thetav) + \int_{\set{M}_1} g(\foo{L}_{\dset{z}}(\thetav)) \diff Q(\thetav).
\end{IEEEeqnarray}
Under the assumption that $g$ is increasing with respect to $\foo{L}_{\dset{z}}$, the measure $P$ over the set $\set{M}_0$ in~\eqref{EqproofSetM0} satisfies
\begin{subequations}
\label{EqIneqP_M0}
\begin{IEEEeqnarray}{rCl}
	P(\set{M}_0) 
	& = & \int_{\set{M}_0} g(\foo{L}_{\dset{z}}(\thetav)) \diff Q(\thetav) 
	\label{EqIneqP_M0_s1} \\  
	& < & \int_{\set{M}_0} \diff Q(\thetav)
	\label{EqIneqP_M0_s2}\\
	& = & Q(\set{M}_0),
	\label{EqIneqP_M0_s3}
\end{IEEEeqnarray}
\end{subequations}
where~\eqref{EqIneqP_M0_s2} follows from~\eqref{EqgStarIsLess1}.
Hence, from \eqref{EqgStarIsLess1} and \eqref{EqIneqP_M0} it follows that
\begin{IEEEeqnarray}{rCl}
	\int_{\set{M}_0} \foo{L}_{\dset{z}}(\thetav) g(\foo{L}_{\dset{z}}(\thetav)) \diff Q(\thetav) 
	& < & \int_{\set{M}_0} \foo{L}_{\dset{z}}(\thetav) \diff Q(\thetav).
	\label{EqIneqRz_M0}
\end{IEEEeqnarray}
Similarly, under the assumption $g$ is increasing with respect to $\foo{L}_{\dset{z}}$, the measure $P$ over the set $\set{M}_1$ in~\eqref{EqproofSetM1} satisfies
\begin{subequations}
\label{EqIneqP_M1}
\begin{IEEEeqnarray}{rCl}
	P(\set{M}_1)
	& = & \int_{\set{M}_1} g(\foo{L}_{\dset{z}}(\thetav)) \diff Q(\thetav) 
	\label{EqIneqP_M1_s1} \\  
	& \geq & \int_{\set{M}_1} \diff Q(\thetav)
	\label{EqIneqP_M1_s2}\\
	& = & Q(\set{M}_1).
\end{IEEEeqnarray}
\end{subequations}
where~\eqref{EqIneqP_M1_s2} follows from~\eqref{EqgStarIsMore1}.
Hence, from \eqref{EqgStarIsMore1} and \eqref{EqIneqP_M1} it follows that
\begin{IEEEeqnarray}{rCl}
	\int_{\set{M}_1} \foo{L}_{\dset{z}}(\thetav) g(\foo{L}_{\dset{z}}(\thetav)) \diff Q(\thetav) 
	& \geq & \int_{\set{M}_1} \foo{L}_{\dset{z}}(\thetav) \diff Q(\thetav).
	\label{EqIneqRz_M1}
\end{IEEEeqnarray}
Note that from~\eqref{EqproofSetM0} and~\eqref{EqIneqRz_M0}, it holds that
\begin{subequations}
	\label{EqIneqP1P2diff_M0}
\begin{IEEEeqnarray}{rCl}
	0 & < & \int_{\set{M}_0} \foo{L}_{\dset{z}}(\thetav) \diff Q(\thetav) - \int_{\set{M}_0} \foo{L}_{\dset{z}}(\thetav) \diff P(\thetav)\\
	& < & k(Q(\set{M}_0)-P(\set{M}_0)).
\end{IEEEeqnarray}
\end{subequations}
Similarly, from~\eqref{EqproofSetM1} and~\eqref{EqIneqRz_M1}, it holds that
\begin{subequations}
	\label{EqIneqP1P2diff_M1}
\begin{IEEEeqnarray}{rCl}
	\int_{\set{M}_1} \foo{L}_{\dset{z}}(\thetav) \diff P(\thetav) 
	- \int_{\set{M}_1} \foo{L}_{\dset{z}}(\thetav) \diff Q(\thetav) 
	& > & k(P(\set{M}_1)-Q(\set{M}_1))\\
	& = & k((1-P(\set{M}_0))-(1-Q(\set{M}_0)))\\
	& = & k(Q(\set{M}_0)-P(\set{M}_0))
\end{IEEEeqnarray}
\end{subequations}
Substituting, \eqref{EqIneqP1P2diff_M0} into \eqref{EqIneqP1P2diff_M1} yields
\begin{IEEEeqnarray}{rCl}
	\int_{\set{M}_0} \foo{L}_{\dset{z}}(\thetav) \diff Q(\thetav) 
	- \int_{\set{M}_0} \foo{L}_{\dset{z}}(\thetav) \diff P(\thetav) & < &\int_{\set{M}_1} \foo{L}_{\dset{z}}(\thetav) \diff P(\thetav) 
	- \int_{\set{M}_1} \foo{L}_{\dset{z}}(\thetav) \diff Q(\thetav),
\end{IEEEeqnarray}
which can be rearranged into,
\begin{IEEEeqnarray}{rCl}
	\int_{\set{M}_0} \foo{L}_{\dset{z}}(\thetav) \diff Q(\thetav) +\int_{\set{M}_1} \foo{L}_{\dset{z}}(\thetav) \diff Q(\thetav)
	& < &\int_{\set{M}_1} \foo{L}_{\dset{z}}(\thetav) \diff P(\thetav) +\int_{\set{M}_0} \foo{L}_{\dset{z}}(\thetav) \diff P(\thetav).
\end{IEEEeqnarray}
Thus, the expected empirical risk of the measures $P$ under the assumption $g$ is increasing with respect to $\foo{L}_{\dset{z}}$ implies that
\begin{IEEEeqnarray}{rCl}
\label{EqIneqLxyQ_lemma}
	\foo{R}_{\dset{z}}(Q) & < & \foo{R}_{\dset{z}}(P), 
\end{IEEEeqnarray}
which completes the proof.
\end{IEEEproof}

\begin{lemma}
\label{lemm_RP1LessRP2}
Let $Q$ be a probability measure on $\msblspc{\set{M}}$, and let $P_{1}$ and $P_{2}$ be two probability measures on $\msblspc{\set{M}}$ that are absolutely continuous with respect to $Q$, with Radon-Nikodym derivatives that satisfy, for all $\thetav \in \supp Q$,
\begin{IEEEeqnarray}{rCl}
\label{EqRNDasFunG}
\frac{\diff P_{i}}{\diff Q}(\thetav) &=& g(-\frac{\foo{L}_{\dset{z}}(\thetav)+\beta_i}{\lambda_i}),
\end{IEEEeqnarray}
where the function $g$ is a strictly increasing function, the function $\foo{L}_{\dset{z}}$ is the empirical risk in \eqref{EqLxy} and the real values satisfy $0<\lambda_1 <\lambda_2$, and $\beta_i \in \reals$, with $i \in \{1,2\}$. 
Then, the expected empirical risk satisfies
\begin{IEEEeqnarray}{rCl}
\foo{R}_{\dset{z}}(P_{1}) < \foo{R}_{\dset{z}}(P_{2}).
\end{IEEEeqnarray}
\end{lemma}
\begin{IEEEproof}
From the strict monotonicity of the function $g$ in \eqref{EqRNDasFunG}, the empirical risk induces
\begin{IEEEeqnarray}{rCl}
	\frac{\diff P_1}{\diff Q}(\thetav) \geq \frac{\diff P_2}{\diff Q}(\thetav) & \iff  & -\frac{\foo{L}_{\dset{z}}(\thetav)+\beta_1}{\lambda_1} \geq -\frac{\foo{L}_{\dset{z}}(\thetav)+\beta_2}{\lambda_2}.
\end{IEEEeqnarray}
Using the fact that $0<\lambda_1<\lambda_2$, let the real value $c \in \reals$ be 
\begin{IEEEeqnarray}{rCl}
	c & = & \frac{\beta_1\lambda_2-\beta_2\lambda_1}{\lambda_1-\lambda_2},
\end{IEEEeqnarray}
and consider the partition of the set $\set{M}$ formed by the sets $\set{A}_0$, $\set{A}_1$ and $\set{A}_2$, which satisfy the following:
\begin{IEEEeqnarray}{rCl}
	\set{A}_0 & \triangleq & \{ \thetav \in \set{M}: \foo{L}_{\dset{z}}(\thetav) = c\},
	\label{EqproofSetA0}\\
	\set{A}_1 & \triangleq  &\{ \thetav \in \set{M}: \foo{L}_{\dset{z}}(\thetav) < c\},
	\label{EqproofSetA1}\\
	\set{A}_2 & \triangleq  &\{ \thetav \in \set{M}: \foo{L}_{\dset{z}}(\thetav) > c\}.
	\label{EqproofSetA2}
\end{IEEEeqnarray}
Note that for all $\thetav \in \set{A}_0$, the pair $(\lambda_2,\beta_2)$ satisfies
\begin{IEEEeqnarray}{rCl}
	-\frac{1}{\lambda_2}(\foo{L}_{\dset{z}}(\thetav) + \beta_2) 
	& = & -\frac{1}{\lambda_2}(c + \beta_2)
	\label{EqprehatEqualLxy_s1}\\
	& = & -\frac{1}{\lambda_2}(\frac{\beta_1\lambda_2-\beta_2\lambda_1}{\lambda_1-\lambda_2}+\frac{ \beta_2(\lambda_1-\lambda_2)}{\lambda_1-\lambda_2})
	\label{EqprehatEqualLxy_s2}\\
	& = & -\frac{1}{\lambda_2}(\frac{\beta_1\lambda_2-\beta_2 \lambda_2}{\lambda_1-\lambda_2})
	\label{EqprehatEqualLxy_s3}\\
	& = & -(\frac{\beta_1-\beta_2}{\lambda_1-\lambda_2}).
	\label{EqprehatEqualLxy_s4}
\end{IEEEeqnarray}
Similarly, for all $\thetav \in \set{A}_0$, the pair $(\lambda_1,\beta_1)$ satisfies
\begin{IEEEeqnarray}{rCl}
	-\frac{1}{\lambda_1}(\foo{L}_{\dset{z}}(\thetav) + \beta_1) 
	& = & -\frac{1}{\lambda_1}(c + \beta_1)
	\label{EqpreEqualLxy_s1}\\
	& = & -\frac{1}{\lambda_1}(\frac{\beta_1\lambda_2-\beta_2\lambda_1}{\lambda_1-\lambda_2}+\frac{ \beta_1(\lambda_1-\lambda_2)}{\lambda_1-\lambda_2})
	\label{EqpreEqualLxy_s2}\\
	& = & -\frac{1}{\lambda_1}(\frac{\beta_1\lambda_1-\beta_2 \lambda_1}{\lambda_1-\lambda_2})
	\label{EqpreEqualLxy_s3}\\
	& = & -(\frac{\beta_1-\beta_2}{\lambda_1-\lambda_2}).
	\label{EqpreEqualLxy_s4}
\end{IEEEeqnarray}
Hence, from~\eqref{EqprehatEqualLxy_s4} and~\eqref{EqpreEqualLxy_s4} for all $\thetav \in \set{A}_0$, it holds that
\begin{IEEEeqnarray}{rCl}
\label{EqEqualandhatl_Lxy}
	-\frac{\foo{L}_{\dset{z}}(\thetav) + \beta_2}{\lambda_2} 
	& = & -\frac{\foo{L}_{\dset{z}}(\thetav) + \beta_1}{\lambda_1}.
\end{IEEEeqnarray}
Then, from~\eqref{EqRNDasFunG} and~\eqref{EqEqualandhatl_Lxy}, it follows that for all  $\thetav \in \set{A}_0$,
\begin{subequations}
\label{EqEqualLxy}
\begin{IEEEeqnarray}{rCl}
	\frac{\diff P_1}{\diff Q}(\thetav) & = & g(-\frac{\foo{L}_{\dset{z}}(\thetav) + \beta_1}{\lambda_1})\label{EqEqualLxy_s1}\\
	& = &g(-\frac{\foo{L}_{\dset{z}}(\thetav) + \beta_2}{\lambda_2})\label{EqEqualLxy_s2}\\
	& = & \frac{\diff P_2}{\diff Q}(\thetav)\label{EqEqualLxy_s3}.
\end{IEEEeqnarray}
\end{subequations}
Furthermore, from the fact that $g$ is strictly increasing, for all~$\thetav \in \set{A}_1$, it holds that
\begin{IEEEeqnarray}{rCl}
\label{EqIneqRDN_f_A1}
	\frac{\diff P_1}{\diff Q}(\thetav) & > & \frac{\diff P_2}{\diff Q}(\thetav), 
\end{IEEEeqnarray}
and for all~$\thetav \in \set{A}_2$, it holds that
\begin{IEEEeqnarray}{rCl}
\label{EqIneqRDN_f_A2}
	\frac{\diff P_1}{\diff Q}(\thetav) & < & \frac{\diff P_2}{\diff Q}(\thetav).
\end{IEEEeqnarray}
Let $P_1$ and $P_2$ denote the probability measures defined by the pairs $(\lambda_1,\beta_1)$ and $(\lambda_2, \beta_2)$, respectively. From~\eqref{EqRNDasFunG}, it follows that 
\begin{IEEEeqnarray}{rCl}
	P_1(\set{A}_1) 
	& = & \int_{\set{A}_1} \frac{\diff P_1}{\diff Q}(\thetav) \diff Q(\thetav) 
	\label{EqpreIneqP_A1_s1},
\end{IEEEeqnarray}
and
\begin{IEEEeqnarray}{rCl}
	P_2(\set{A}_1) 
	& = & \int_{\set{A}_1} \frac{\diff P_2}{\diff Q}(\thetav) \diff Q(\thetav) 
	\label{EqpreIneqhatP_A1_s1}.
\end{IEEEeqnarray}
From~\eqref{EqpreIneqP_A1_s1} and~\eqref{EqpreIneqhatP_A1_s1}, the measures $P_1$ and $P_2$ over the set $\set{A}_1$ in~\eqref{EqproofSetA1} satisfy
\begin{subequations}
\label{EqIneqP_A1}
\begin{IEEEeqnarray}{rCl}
	P_1(\set{A}_1) 
	& = & \int_{\set{A}_1} \frac{\diff P_1}{\diff Q}(\thetav) \diff Q(\thetav) 
	\label{EqIneqP_A1_s1} \\  
	& > & \int_{\set{A}_1} \frac{\diff P_2}{\diff Q}(\thetav)  \diff Q(\thetav)
	\label{EqIneqP_A1_s2}\\
	& = & P_2(\set{A}_1),
	\label{EqIneqP_A1_s3}
\end{IEEEeqnarray}
\end{subequations}
where~\eqref{EqIneqP_A1_s2} follows from~\eqref{EqIneqRDN_f_A1}.
Hence, from \eqref{EqIneqP_A1} it follows that
\begin{IEEEeqnarray}{rCl}
	\int_{\set{A}_1} \foo{L}_{\dset{z}}(\thetav) \frac{\diff P_1}{\diff Q}(\thetav) \diff Q(\thetav) 
	& > & \int_{\set{A}_1} \foo{L}_{\dset{z}}(\thetav) \frac{\diff P_2}{\diff Q}(\thetav)  \diff Q(\thetav).
	\label{EqIneqRz_A1}
\end{IEEEeqnarray}
Similarly, from~\eqref{EqpreIneqP_A1_s1} and~\eqref{EqpreIneqhatP_A1_s1}, the measures $P_1$ and $P_2$ over the set $\set{A}_2$ in~\eqref{EqproofSetA2} satisfy
\begin{subequations}
\label{EqIneqP_A2}
\begin{IEEEeqnarray}{rCl}
	P_1(\set{A}_2)
	& = & \int_{\set{A}_2} \frac{\diff P_1}{\diff Q}(\thetav) \diff Q(\thetav) 
	\label{EqIneqP_A2_s1} \\  
	& < & \int_{\set{A}_2} \frac{\diff P_2}{\diff Q}(\thetav)  \diff Q(\thetav)
	\label{EqIneqP_A2_s2}\\
	& = & P_2(\set{A}_2).
\end{IEEEeqnarray}
\end{subequations}
where~\eqref{EqIneqP_A2_s2} follows from~\eqref{EqIneqRDN_f_A2}.
Hence, from \eqref{EqIneqP_A2} it follows that
\begin{IEEEeqnarray}{rCl}
	\int_{\set{A}_2} \foo{L}_{\dset{z}}(\thetav) \frac{\diff P_1}{\diff Q}(\thetav) \diff Q(\thetav) 
	& < & \int_{\set{A}_2} \foo{L}_{\dset{z}}(\thetav) \frac{\diff P_2}{\diff Q}(\thetav)  \diff Q(\thetav).
	\label{EqIneqRz_A2}
\end{IEEEeqnarray}
Note that from~\eqref{EqproofSetA1} and~\eqref{EqIneqRz_A1}, it holds that
\begin{subequations}
	\label{EqIneqP1P2diff_A1}
\begin{IEEEeqnarray}{rCl}
	0 & < &\int_{\set{A}_1} \foo{L}_{\dset{z}}(\thetav) \diff P_1(\thetav) 
	- \int_{\set{A}_1} \foo{L}_{\dset{z}}(\thetav) \diff P_2(\thetav)\\
	& < & c(P_1(\set{A}_1)-P_2(\set{A}_1)).
\end{IEEEeqnarray}
\end{subequations}
Similarly, from~\eqref{EqproofSetA2} and~\eqref{EqIneqRz_A2}, it holds that
\begin{subequations}
	\label{EqIneqP1P2diff_A2}
\begin{IEEEeqnarray}{rCl}
	\int_{\set{A}_2}\!\! \foo{L}_{\dset{z}}(\thetav) \diff P_2(\thetav) 
	\!-\! \int_{\set{A}_2}\!\! \foo{L}_{\dset{z}}(\thetav) \diff P_1(\thetav)
	& > & c(P_2(\set{A}_2)-P_1(\set{A}_2))\\
	& = & c((1-P_2(\set{A}_1))-(1-P_1(\set{A}_1)))\qquad\\
	& = & c(P_1(\set{A}_1)-P_2(\set{A}_1))
\end{IEEEeqnarray}
\end{subequations}
Substituting, \eqref{EqIneqP1P2diff_A1} into \eqref{EqIneqP1P2diff_A2} yields
\begin{IEEEeqnarray}{rCl}
	\int_{\set{A}_1}\!\! \foo{L}_{\dset{z}}(\thetav) \diff P_1(\thetav) 
	\!-\!\! \int_{\set{A}_1}\!\! \foo{L}_{\dset{z}}(\thetav) \diff P_2(\thetav) & < &\!\int_{\set{A}_2}\!\! \foo{L}_{\dset{z}}(\thetav) \diff P_2(\thetav) 
	- \int_{\set{A}_2}\!\! \foo{L}_{\dset{z}}(\thetav) \diff P_1(\thetav),\qquad
\end{IEEEeqnarray}
which can be rearranged into,
\begin{IEEEeqnarray}{rCl}
	\int_{\set{A}_1}\!\! \foo{L}_{\dset{z}}(\thetav) \diff P_1(\thetav)\!+\!\int_{\set{A}_2}\!\! \foo{L}_{\dset{z}}(\thetav) \diff P_1(\thetav)
	& < &\int_{\set{A}_2}\!\! \foo{L}_{\dset{z}}(\thetav) \diff P_2(\thetav)\!+\!\int_{\set{A}_1}\!\! \foo{L}_{\dset{z}}(\thetav) \diff P_2(\thetav).\qquad
\end{IEEEeqnarray}
Thus, the expected empirical risk of the measures $P_1$ and $P_2$ satisfies
\begin{IEEEeqnarray}{rCl}
\label{EqIneqLxy_lemm}
	\foo{R}_{\dset{z}}(P_1) & < & \foo{R}_{\dset{z}}(P_2). 
\end{IEEEeqnarray}
This completes the proof.
\end{IEEEproof}

\begin{lemma}
\label{lemm_monotonic_integral}
Let $Q$ be a probability measure on a measurable space $\msblspc{\set{M}}$. Consider the pair $(a_1,b_1) \in (0,\infty) \times \reals$ and let $g: \reals \to (0, \infty)$ be a strictly increasing function such that
\begin{IEEEeqnarray}{rCl}
\int g(-\frac{\foo{L}_{\dset{z}}(\thetav)+b_1}{a_1}) \diff Q(\thetav) & < & \infty,
\label{eq:integrability_condition}
\end{IEEEeqnarray}
where $\foo{L}_{\dset{z}}$ is defined in~\eqref{EqLxy}.
Then the following monotonicity properties hold:
For all $b_2 \in \mathbb{R}$ with $b_1 < b_2$:
    \begin{equation}
        \int g\left(-\frac{\foo{L}_{\dset{z}}(\thetav)+b_1}{a_1}\right) \diff Q(\thetav)
        > \int g\left(-\frac{\foo{L}_{\dset{z}}(\thetav)+b_2}{a_1}\right) \diff Q(\thetav).
        \label{eq:b_monotonicity}
    \end{equation}
\end{lemma}
\begin{IEEEproof}
For all $(b_1, b_2) \in \reals^2$, such that $b_1 < b_2$, it holds that for all $\thetav \in \supp Q$,
\begin{IEEEeqnarray}{rCCCl}
	-\frac{\foo{L}_{\dset{z}}(\thetav) + b_1}{a_1}
	& > & -\frac{\foo{L}_{\dset{z}}(\thetav) + b_2}{a_1}.
\end{IEEEeqnarray}
Then, from the assumption that $g$ is monotonically increasing it follows that for all $\thetav \in \supp Q$,
\begin{equation}
\label{EqIneq_cnvxcnj_2_appc}
	g\Big(\!-\frac{\foo{L}_{\dset{z}}(\thetav)\! +\! b_1}{a_1}\Big)\! > \! g\Big(\!\!-\frac{\foo{L}_{\dset{z}}(\thetav)\! +\! b_2}{a_1}\Big).
\end{equation}
Thus, integrating~\eqref{EqIneq_cnvxcnj_2_appc} with respect to $Q$ yields
\begin{equation}
\!\int\!\! g\Big(\!\!-\frac{\foo{L}_{\dset{z}}(\thetav)\! +\! b_1}{a} \Big)\! \diff Q(\thetav)\! > \!\! \int\!\! g\Big(\!\!-\frac{\foo{L}_{\dset{z}}(\thetav)\! +\! b_2}{a}\Big)\! \diff Q(\thetav),
\end{equation}
which completes the proof.\end{IEEEproof}

\section{Proof of Theorem~\ref{Theo_f_ERMRadNik}}

\begin{IEEEproof}
\label{app_theo_f_ERMRadNik2}
The optimization problems in~\eqref{EqOp_f_ERMRERNormal} and~\eqref{EqOp_f_ERM_RND2} can be re-written in terms of the \RadonNikodym derivative of the optimization measure $P$ with respect to the reference measure $Q$, denoted by $\frac{\diff P}{\diff Q}: \set{M} \rightarrow [0,\infty)$, which yields:
\begin{subequations}
\label{EqOp_f_ERM_RND2_pf}
\begin{IEEEeqnarray}{rCl}
	\min_{P \in \bigtriangleup_{Q}(\set{M})}
	& \quad & \int \foo{L}_{\dset{z}}(\thetav) \frac{\diff P}{\diff Q}(\thetav) \diff Q(\thetav)\\
	\text{s.t.} 
 	& &  \int f( \frac{\diff P}{\diff Q}(\thetav)) \diff Q(\thetav) \leq \eta \label{EqOp_f_ERM_RND2_pf_c_s1}\\
 	& & \int \frac{\diff P}{\diff Q}(\thetav) \diff Q (\thetav) =1.\label{EqOp_f_ERM_RND2_pf_c_s2}
\end{IEEEeqnarray}
\end{subequations}
The remainder of the proof focuses on the problem in which the optimization is over the Radon-Nikodym derivative $\frac{\diff P}{\diff Q}$ instead of the measures $P$. This is due to the fact that for all $P \in \bigtriangleup_{Q}(\set{M})$, the Radon-Nikodym derivative $\frac{\diff P}{\diff Q}$  is unique up to sets of measure zero with respect to $Q$.
The first part is as follows. Let $\field{M}$ be the set of measurable functions $\set{M}\rightarrow \reals$ with respect to the measurable space $\msblspc{\set{M}}$ and $(\reals, \borelsigma)$. Let $\field{S}$ be the subset of $\field{M}$, including all nonnegative functions that are absolutely integrable with respect to $Q$. That is, for all $\hat{g}\in \field{S}$, it holds that
\begin{IEEEeqnarray}{rCl}
	\int \abs{\hat{g}(\thetav)}\diff Q(\thetav) & < & \infty.	
\end{IEEEeqnarray}
Note that the set $\field{M}$ forms a real vector space and the set $\field{S}$ is a convex subset of $\field{M}$. Note also that the constraints~\eqref{EqOp_f_ERM_RND2_pf_c_s1} and~\eqref{EqOp_f_ERM_RND2_pf_c_s2} are satisfied by the probability measure $Q$, which also satisfies $Q \in \bigtriangleup_{Q}( \set{M} )$.
Hence, the constraints do not induce an empty feasible set. Finally, note that without loss of generality the minimization in~\eqref{EqOp_f_ERM_RND2_pf} can be written as a minimization problem of the form:
\begin{subequations}
\label{EqOp_f_ERM_Min_all}
\begin{IEEEeqnarray}{rCl}
	\min_{g \in \field{S}} 
	& \quad & \int \foo{L}_{\dset{z}}(\thetav) g(\thetav) \diff Q(\thetav)\\
	\text{s.t.} 
	& &  \frac{1}{\eta}\int f(g(\thetav)) \diff Q(\thetav) \leq 1 \label{EqOp_f_ERM_Min_c_s1}\\
 	& & \int g (\thetav) \diff Q(\thetav) = 1 \label{EqOp_f_ERM_Min_c_s2},
\end{IEEEeqnarray}
\end{subequations}
where the expressions $ \int \foo{L}_{\dset{z}}(\thetav) g(\thetav) \diff Q(\thetav)$ and $\int g (\thetav) \diff Q(\thetav)$ are linear with $g$; the expression $ \frac{1}{\eta}\int f(g(\thetav)) \diff Q(\thetav)$ is convex with $g$.
The proof continues by assuming that the problem in~\eqref{EqOp_f_ERM_Min_all} possesses a solution, which is denoted by $g^{\star} \in \field{S}$. Let $\mu_0 \in [0,\infty)$ be 
\begin{subequations}
\label{EqOp_f_ERM_Min_all_mu}
\begin{IEEEeqnarray}{rcCl}
	\mu_0 &\ \triangleq\ & \min_{g \in \field{S}} 
	 & \int \foo{L}_{\dset{z}}(\thetav) g(\thetav) \diff Q(\thetav)\\
	&   &\text{s.t.}&
	\frac{1}{\eta}\int f(g(\thetav)) \diff Q(\thetav) \leq 1 \label{EqOp_f_ERM_Min_mu_c_s1}\\
 	&   &  & \int g (\thetav) \diff Q(\thetav) = 1 \label{EqOp_f_ERM_Min_mu_c_s2}\\
 	& = &  & \int \foo{L}_{\dset{z}}(\thetav) g^{\star}(\thetav) \diff Q(\thetav).
\end{IEEEeqnarray}
\end{subequations}
From~\cite[Theorem~1, Section~8.3]{luenberger1997bookOptimization}, it holds that there exists two tuples $(a_1,b_1)$ and $(a_2,b_2)$ in $\reals^2$ such that
\begin{subequations}
\label{EqOp_f_ERM_Min_all_anci}
\begin{IEEEeqnarray}{rCl}
	\mu_0 
	& = & \min_{g \in \field{S}} \{ \int \foo{L}_{\dset{z}}(\thetav) g(\thetav) \diff Q(\thetav) +  \frac{a_1}{\eta}\int f(g(\thetav)) \diff Q(\thetav) 
	\splitR[1]
	+ b_1 + a_2\int g (\thetav) \diff Q(\thetav) + b_2  \},
\end{IEEEeqnarray}
and moreover,
\begin{IEEEeqnarray}{rCl}
\label{EqOp_f_ERM_Min_all_anci_c_s2}
	0 & = & \frac{a_1}{\eta}\int f(g^{\star}(\thetav)) \diff Q(\thetav) + b_1, \text{ and}\\
\label{EqOp_f_ERM_Min_all_anci_c_s3}
	0 & = & a_2\int g^{\star} (\thetav) \diff Q(\thetav) + b_2. 
\end{IEEEeqnarray}
\end{subequations}
Hence, the proof continues by solving the ancillary optimization problem in~\eqref{EqOp_f_ERM_Min_all_anci}, which allows the reformulation of the optimization problem in an unconstrained dual problem. This reformulation is possible as the tuples $(a_1,b_1)$ and $(a_2,b_2)$ are such that equalities~\eqref{EqOp_f_ERM_Min_all_anci_c_s2} and~\eqref{EqOp_f_ERM_Min_all_anci_c_s3} are satisfied, by definition.

Let the function $L:\field{S} \rightarrow \reals$ be such that
\begin{IEEEeqnarray}{rCl}
L(g)
& = & \int\! \foo{L}_{\dset{z}}(\thetav) g(\thetav) \diff Q(\thetav) \!+\! \frac{a_1}{\eta}\!\int\! f(g(\thetav))\diff Q(\thetav) \!+\! b_1 \!+\! a_2\!\int\! g (\thetav) \diff Q(\thetav) \!+\! b_2. \qquad
	\label{Eq_Lagrange_Min}
\end{IEEEeqnarray}
Let $\hat{g}:\set{M} \rightarrow \reals$ be a function in $\field{S}$. 
The Gateaux differential of the functional $L$ in~\eqref{Eq_Lagrange_Min} at $\left(g, \beta\right) \in \mathscr{S}\times \reals$ in the direction of $\hat{g}$ is
\begin{IEEEeqnarray}{rcl}
\label{EqNecessaryCondtionDivff}
\partial L(g; \hat{g} ) & \triangleq & \left.\frac{\diff}{\diff \alpha}  L(g + \alpha \hat{g}, \beta) \right|_{\alpha = 0}.
\end{IEEEeqnarray}
Let the function $r:\reals \rightarrow \reals$ be defined for some fixed functions $g$ and $\hat{g}$ and some fixed $a_1$, $b_1$, $a_2$ and $b_2$ such that for all $\alpha \in (-\epsilon, \epsilon)$, with $\epsilon$ arbitrarily small, 
\begin{IEEEeqnarray}{rcl}
\label{Eq_r_is_Lgg}
    r(\alpha) & = & L(g + \alpha \hat{g}).
\end{IEEEeqnarray}
The proof follows by showing that the function $r$ in \eqref{Eq_r_is_Lgg} is differentiable at zero, in order to prove the existence of the Gateaux differential in \eqref{EqNecessaryCondtionDivff} for the functions $g$ and $\hat{g}$ and real values $a_1$, $b_1$, $a_2$ and $b_2$.
To this end, note that
\begin{IEEEeqnarray}{rcl}
r(\alpha)
& = & \int \foo{L}_{\vect{z}}(\thetav)(g (\thetav) \!+\! \alpha \hat{g}(\thetav))\,\diff Q(\thetav)
\splitD
+ \> \frac{a_1}{\eta} \int f(g(\thetav) \!+\! \alpha \hat{g}(\thetav))\,\diff Q(\thetav)+b_1 
\splitR \splitD
+ \> a_2\int(g(\thetav)  \!+\! \alpha \hat{g}(\thetav))\diff Q(\thetav) + b_2 
\end{IEEEeqnarray}
which can be rewritten as follows,
\begin{IEEEeqnarray}{rcl}
r(\alpha) 
& = & \alpha \int \hat{g}(\thetav)(a_2 +\foo{L}_{\vect{z}}(\thetav))   \diff Q(\thetav)
\splitD
+ \> \frac{a_1}{\eta} \int f(g(\thetav) + \alpha \hat{g}(\thetav))\diff Q(\thetav) 
\splitR \splitD
+ \int g(\thetav)(a_2 + \foo{L}_{\vect{z}}(\thetav))\,\diff Q(\thetav)
\splitD
+ b_1  + b_2.\label{EqpreGateaux_r}
\end{IEEEeqnarray}
Note that the first term in~\eqref{EqpreGateaux_r} is linear with $\alpha$; the second term can be written using the function $\hat{r}: \reals \to \reals$ such that for all $\alpha \in (-\epsilon, \epsilon)$, with $\epsilon$ arbitrarily small, it holds that
\begin{IEEEeqnarray}{rcl}
\label{EqrHatForDaunas}
\hat{r}(\alpha) & = &  \lambda \int f(g(\thetav) + \alpha \hat{g}(\thetav))\,\diff Q(\thetav);
\end{IEEEeqnarray}
and the remaining terms are independent of $\alpha$.
Hence, based on the fact that the function $\hat{r}$ in~\eqref{EqrHatForDaunas} is differentiable at zero (see Lemma~\ref{lemm_ERM_fDR_DiffZero}~in Appendix~\ref{sec:AppendixA}), so is the function $r$ in~\eqref{EqpreGateaux_r}, which implies that the G{\^a}teaux differential of $\partial L (g,\hat{g})$ in~\eqref{EqNecessaryCondtionDivff} exists.

The proof proceeds by calculating the Gateaux differential  $\partial L(g, \hat{g} )$ in \eqref{EqNecessaryCondtionDivff}, which requires calculating the derivative of the real function $r$ in~\eqref{EqpreGateaux_r}. 
That is,
\begin{IEEEeqnarray}{rCl}
	\frac{\diff}{\diff \alpha} r(\alpha) 
	& = & \frac{\diff}{\diff \alpha} \left(\alpha \int \hat{g}(\thetav)(a_2 +\foo{L}_{\vect{z}}(\thetav))   \diff Q(\thetav)
+ \> \frac{a_1}{\eta} \int f(g(\thetav) + \alpha \hat{g}(\thetav))\diff Q(\thetav) 
	\right. \nonumber \\ &    & \left. 
	+ \int g(\thetav)(a_2 + \foo{L}_{\vect{z}}(\thetav) \right)\,\diff Q(\thetav)
+ b_1  + b_2)
	\label{Eq_r_dalpha_Min_s1}\\
	& = &  \int \hat{g}(\thetav)(a_2 +\foo{L}_{\vect{z}}(\thetav))   \diff Q(\thetav) 
+  \frac{a_1}{\eta} \int \frac{\diff}{\diff \alpha} f(g(\thetav) + \alpha \hat{g}(\thetav))\diff Q(\thetav)
	\label{Eq_r_dalpha_Min_s2}\\
	& = &  \int \hat{g}(\thetav)(a_2 +\foo{L}_{\vect{z}}(\thetav))   \diff Q(\thetav) 
+  \frac{a_1}{\eta} \int \hat{g}(\thetav)\dot{f}(g(\thetav) + \alpha \hat{g}(\thetav))\diff Q(\thetav)
	\label{Eq_r_dalpha_Min_s3}
\end{IEEEeqnarray}
where~\eqref{Eq_r_dalpha_Min_s2} follows from Theorem~\ref{Theo_ERM_fDR_Leibniz}.
From equations~\eqref{EqNecessaryCondtionDivff} and~\eqref{Eq_r_dalpha_Min_s3}, it follows that
\begin{subequations}
\label{Eq_dL_Min}
\begin{IEEEeqnarray}{rCl}
	\partial L(g;\hat{g}) 
	& = &  \int \hat{g}(\thetav)(a_2 +\foo{L}_{\vect{z}}(\thetav))   \diff Q(\thetav) 
+  \frac{a_1}{\eta} \int \hat{g}(\thetav)\dot{f}(g(\thetav))\diff Q(\thetav)
	\label{Eq_dL_Min_s1}\\
	& = &  \int \hat{g}(\thetav)(a_2 +\foo{L}_{\vect{z}}(\thetav) 
+  \frac{a_1}{\eta}  \dot{f}(g(\thetav)))\diff Q(\thetav)
	\label{Eq_dL_Min_s2}.
\end{IEEEeqnarray}
\end{subequations}
From \cite[Theorem~1, Chapter 7]{luenberger1997bookOptimization}, a necessary condition to use for the functional $L$ in~\eqref{Eq_Lagrange_Min} to have a minimum at $g^{\star}$ is that for all functions $\hat{g}\in \field{S}$,
\begin{IEEEeqnarray}{rCl}
\label{Eq_dL_Min_luen_Th1}
\partial L(g^{\star};\hat{g}) & = & 0.	
\end{IEEEeqnarray}
From~\eqref{Eq_dL_Min_s2} and \eqref{Eq_dL_Min_luen_Th1}, it follows that $g^{\star}$ must satisfy for all $\thetav \in \supp Q$, that
\begin{IEEEeqnarray}{rCl}
	\label{Eq_dL_Min_zero}
	\foo{L}_{\dset{z}}(\thetav)  + \frac{a_1}{\eta}\dot{f}(g^{\star}(\thetav)) + a_2 & = & 0.	
\end{IEEEeqnarray}
Assuming that
\begin{IEEEeqnarray}{rCl}
\label{Eq_dL_Min_zero_c}
	a_1 \neq 0,	
\end{IEEEeqnarray}
it follows from~\eqref{Eq_dL_Min_zero} that
\begin{IEEEeqnarray}{rCl}
	\label{Eq_Op_RND_Min}
	g^{\star}(\thetav) & = & \dot{f}^{-1}(-\frac{\eta}{a_1}(\foo{L}_{\dset{z}}(\thetav) + a_2)),	
\end{IEEEeqnarray}
where the values $a_1$ and $a_2$ satisfy~\eqref{EqOp_f_ERM_Min_all_anci_c_s2} and~\eqref{EqOp_f_ERM_Min_all_anci_c_s3} and~\eqref{Eq_dL_Min_zero_c}.

The remainder of the proof focuses on determining the values of $a_1$, $a_2$, $b_1$, and $b_2$, which must also be such that $g^{\star}$ in~\eqref{Eq_Op_RND_Min} satisfies the constraints~\eqref{EqOp_f_ERM_Min_c_s1} and~\eqref{EqOp_f_ERM_Min_c_s2} under the assumption that $\foo{L}_{\dset{z}}$ in~\eqref{EqLxy} is separable. 
For instance, from constraints~\eqref{EqOp_f_ERM_Min_c_s1} and~\eqref{EqOp_f_ERM_Min_all_anci_c_s3}, it follows that
\begin{IEEEeqnarray}{rCl}
\label{EqConstrainA2eqB2}
	a_2 & = & -b_2.
\end{IEEEeqnarray}
From~\eqref{EqConstrainA2eqB2}, the constraint in~\eqref{EqOp_f_ERM_Min_all_anci_c_s3} implies that the choice of $a_2$ satisfies
\begin{IEEEeqnarray}{rCl}
\label{EqConstA2isOne}
	1 & = &	\int g^{\star}(\thetav)\diff Q(\thetav).
\end{IEEEeqnarray}
Similarly, the function $g^{\star}$ in~\eqref{Eq_Op_RND_Min} is the Radon-Nikodym derivative with respect to $Q$ of the solution $P^{\star} \in \bigtriangleup( \set{M} )$ to the problem in~\eqref{EqOp_f_ERM_RND2_pf}. Hence, \eqref{EqOp_f_ERM_Min_all_anci_c_s2} can be written as follows
\begin{IEEEeqnarray}{rCl}
\label{Eq_Op_KL_Min_Jpz}
	\frac{a_1}{\eta}\KLf{P^{\star}}{Q}  + b_1 & = & 0,	
\end{IEEEeqnarray}
which implies
\begin{IEEEeqnarray}{rCl}
	b_1 & = & - \frac{a_1}{\eta} \KLf{P^{\star}}{Q}.	
\end{IEEEeqnarray}
Note that if $a_1 < 0$, then the function $g^{\star}$ in~\eqref{Eq_Op_RND_Min} is strictly increasing with respect to the empirical risk $\foo{L}_{\dset{z}}$ in \eqref{EqLxy}. Then, from Lemma~\ref{lemm_UpperboundRQ} in Appendix~\ref{sec:AppendixA}, it follows that
\begin{IEEEeqnarray}{rCl}
\label{EqIneqLxyQ_pf}
	\foo{R}_{\dset{z}}(P^{\star})>\foo{R}_{\dset{z}}(Q).
\end{IEEEeqnarray}
Note that from Definition~\ref{Def_fDivergence}, for all $P \in \bigtriangleup_Q(\set{M})$, it follows that 
\begin{IEEEeqnarray}{rCl}
\label{EqDivIneqQ_pf}
	\Divf{Q}{Q} & \leq &  \Divf{P}{Q}, 
\end{IEEEeqnarray}
with equality, if and only if $P=Q$. Hence, \eqref{EqIneqLxyQ_pf} and \eqref{EqDivIneqQ_pf} imply 
\begin{IEEEeqnarray}{rCl}
\label{EqCostFunIneqQ_pf}
	\foo{R}_{\dset{z}}(Q) + \lambda\Divf{Q}{Q} & < &  \foo{R}_{\dset{z}}(P^{\star})+ \lambda \Divf{P^{\star}}{Q}, 
\end{IEEEeqnarray}
which is a contradiction.
Thus, the focus in the remainder of the proof is on the case in which
\begin{IEEEeqnarray}{rCl}
	a_1 & > & 0,	
\end{IEEEeqnarray}
which implies that for models $\thetav_1$ and $\thetav_2$ in $\supp Q$ satisfying $\foo{L}_{\dset{z}}(\thetav_1) < \foo{L}_{\dset{z}}(\thetav_2)$, it holds that
\begin{IEEEeqnarray}{rCl}
\label{EqIneq_g_decreasing}
	g^{\star}(\thetav_1) & > & g^{\star}(\thetav_2).	
\end{IEEEeqnarray}
Given the pairs $(a_1,a_2)$ and $(\hat{a}_1,\hat{a}_2)$ in $\reals^2$ such that each pair satisfies the constraints in~\eqref{EqOp_f_ERM_Min_all_anci_c_s2} and~\eqref{EqOp_f_ERM_Min_all_anci_c_s3}, from~\eqref{Eq_Op_RND_Min} there exists a solution for each pair given by
\begin{IEEEeqnarray}{rCl}
	\label{Eq_Op_RND_Min_sub1}
	g^{\star}(\thetav) & = & \dot{f}^{-1}(-\frac{\eta}{a_1}(\foo{L}_{\dset{z}}(\thetav) + a_2)),	
\end{IEEEeqnarray}
and
\begin{IEEEeqnarray}{rCl}
	\label{Eq_Op_RND_Min_sub2}
	\hat{g}^{\star}(\thetav) & = & \dot{f}^{-1}(-\frac{\eta}{\hat{a}_1}(\foo{L}_{\dset{z}}(\thetav) + \hat{a}_2)),	
\end{IEEEeqnarray}
where the functions $g^{\star}$ and $\hat{g}^{\star}$ are the Radon-Nikodym derivative of the solutions $P^{\star}$ and $\hat{P}^{\star}$ with respect to $Q$ for each pair $(a_1,a_2)$ and $(\hat{a}_1,\hat{a}_2)$, respectively.
Since $f$ is strictly convex, Lemma~\ref{lemm_f_invIsInc} in Appendix~\ref{sec:AppendixA} implies that $\dot{f}^{-1}$ is strictly increasing. Moreover, by Lemma~\ref{lemm_RP1LessRP2} in Appendix~\ref{sec:AppendixA} it folllows that 
\begin{IEEEeqnarray}{rCl}
\label{EqIneqLxy_pf}
	\foo{R}_{\dset{z}}(P^{\star}) & < & \foo{R}_{\dset{z}}(\hat{P}^{\star}). 
\end{IEEEeqnarray}
From~\eqref{EqIneqLxy_pf} and the assumption of $a_1 < \hat{a}_1$, it follows that 
\begin{subequations}
\label{Eq_da1_g_star_increase}
\begin{IEEEeqnarray}{rCl}
	\frac{\diff }{\diff a_1} \foo{R}_{\dset{z}}(P^{\star}) & = & \frac{\diff }{\diff a_1} \int \foo{L}_{\dset{z}}(\thetav) \diff P^{\star}(\thetav)
	\label{Eq_da1_g_star_increase_s1}\\
	& = & \lim_{\hat{a}_1 \rightarrow a_1} \frac{\int \foo{L}_{\dset{z}}(\thetav) \diff \hat{P}^{\star}(\thetav)-\int \foo{L}_{\dset{z}}(\thetav) \diff P^{\star}(\thetav)}{\hat{a}_1 - a_1}
	\label{Eq_da1_g_star_increase_s2}\\
	& > & 0,\label{Eq_da1_g_star_increase_s3}
\end{IEEEeqnarray}
\end{subequations}
where~\eqref{Eq_da1_g_star_increase_s3} follows from~\eqref{EqIneqLxy_pf}.
The proof continues by showing the induced $f$-divergence by the function $f$ is increasing with respect to $a_1$, which is equivalent to showing that the derivative of $\foo{D}_{f}$ with respect to $a_1$ is always positive.
To do so, note that from Assumption~\ref{assume:a} in Theorem~\ref{Theo_f_ERMRadNik} specifically, the strict convexity, it follows from Jensen inequality that
\begin{subequations}
\label{EqDfIneqPandPs}
\begin{IEEEeqnarray}{rCl}
	\IEEEmulticolR{
	\IEEEmulticolD{
	\KLf{P^{\star}}{Q}-\KLf{\hat{P}^{\star}}{Q}
	}}
	& = & \int f(g^{\star}(\thetav))\diff Q(\thetav) - \int f(\hat{g}^{\star}(\thetav)) \diff Q(\thetav)\label{EqDfIneqPandPs_s1}\\
	& = & \int f(g^{\star}(\thetav))-f(\hat{g}^{\star}(\thetav)) \diff Q(\thetav)
	\label{EqDfIneqPandPs_s2}\\
	& > & \int \dot{f}(\hat{g}^{\star}(\thetav))(g^{\star}(\thetav)-\hat{g}^{\star}(\thetav))\diff Q(\thetav)
	\label{EqDfIneqPandPs_s3}\\
	& = & \int \dot{f}(\dot{f}^{-1}(-\frac{\eta}{\hat{a}_1}(\foo{L}_{\dset{z}}(\thetav) + \hat{a}_2)))(g^{\star}(\thetav)-\hat{g}^{\star}(\thetav)) \diff Q(\thetav)
	\label{EqDfIneqPandPs_s4}\\
	& = & \int -\frac{\eta}{\hat{a}_1}(\foo{L}_{\dset{z}}(\thetav) + \hat{a}_2)(g^{\star}(\thetav)-\hat{g}^{\star}(\thetav)) \diff Q(\thetav)
	\label{EqDfIneqPandPs_s5}\\
	& = & \int \frac{\eta}{\hat{a}_1}(\foo{L}_{\dset{z}}(\thetav) + \hat{a}_2)\hat{g}^{\star}(\thetav)-\frac{\eta}{\hat{a}_1}(\foo{L}_{\dset{z}}(\thetav) + \hat{a}_2)g^{\star}(\thetav) \diff Q(\thetav)
	\label{EqDfIneqPandPs_s6}\\
	& = & \int \frac{\eta}{\hat{a}_1}(\foo{L}_{\dset{z}}(\thetav) + \hat{a}_2)\hat{g}^{\star}(\thetav) \diff Q(\thetav) \!-\! \int\frac{\eta}{\hat{a}_1}(\foo{L}_{\dset{z}}(\thetav) + \hat{a}_2)g^{\star}(\thetav) \diff Q(\thetav)
	\label{EqDfIneqPandPs_s7}\\
	& = & \int \frac{\eta}{\hat{a}_1}(\foo{L}_{\dset{z}}(\thetav) + \hat{a}_2) \diff \hat{P}^{\star}(\thetav)- \int\frac{\eta}{\hat{a}_1}(\foo{L}_{\dset{z}}(\thetav) + \hat{a}_2) \diff P^{\star}(\thetav)
	\label{EqDfIneqPandPs_s8}\\
	& = & \frac{\eta}{\hat{a}_1}(\int \foo{L}_{\dset{z}}(\thetav) \diff \hat{P}^{\star}(\thetav)+ \hat{a}_2 - \int\foo{L}_{\dset{z}}(\thetav) \diff P^{\star}(\thetav)- \hat{a}_2)
	\label{EqDfIneqPandPs_s9}\\
	& = & \frac{\eta}{\hat{a}_1}(\int \foo{L}_{\dset{z}}(\thetav) \diff \hat{P}^{\star}(\thetav) - \int\foo{L}_{\dset{z}}(\thetav) \diff P^{\star}(\thetav))
	\label{EqDfIneqPandPs_s10}\\
	& = & \frac{\eta}{\hat{a}_1}(\foo{R}_{\dset{z}}(\hat{P}^{\star}) - \foo{R}_{\dset{z}}(P^{\star}))
	\label{EqDfIneqPandPs_s11}\\
	& > & 0,\label{EqDfIneqPandPs_s12}
\end{IEEEeqnarray}
\end{subequations}
where~\eqref{EqDfIneqPandPs_s3} follows from first-order condition (see \cite[Section 3.1.3]{boyd2004convex};~\eqref{EqDfIneqPandPs_s4} follows from \eqref{Eq_Op_RND_Min_sub2};~\eqref{EqDfIneqPandPs_s8} follows from~\eqref{EqpreIneqP_A1_s1} and~\eqref{EqpreIneqhatP_A1_s1}; and \eqref{EqDfIneqPandPs_s12} follows from \eqref{EqIneqLxy_pf} and the facts that $0 < \eta, \hat{a}_1$.
Observe that from~\eqref{EqDfIneqPandPs} and the assumption of $a_1 < \hat{a}_1$, it follows that
\begin{IEEEeqnarray}{rCl}
\frac{\diff }{\diff a_1} \Divf{P^{\star}}{Q} 
& = &\frac{\diff }{\diff a_1} \int f(g^{\star}(\thetav)) \diff Q(\thetav)\label{Eq_Op_da1_Divf_s1}\\
	& = & \lim_{\hat{a}_1 \rightarrow a_1} \frac{\Divf{P^{\star}}{Q}-\Divf{\hat{P}^{\star}}{Q}}{a_1-\hat{a}_1}\\
	& < & 0,\label{Eq_Op_da1_Divf_s7}
\end{IEEEeqnarray}
where~\eqref{Eq_Op_da1_Divf_s7} follows from~\eqref{EqDfIneqPandPs_s12} and that $a_1-\hat{a}_1<0$.
Hence, from~\eqref{Eq_da1_g_star_increase} the term $\int g^{\star}(\thetav)\foo{L}_{\dset{z}}(\thetav) \diff Q(\thetav)$ in~\eqref{EqOp_f_ERM_RND2_pf_c_s1} is strictly increasing with $a_1$ and from~\eqref{Eq_Op_da1_Divf_s7} the term $\int f(g^{\star}(\thetav)) \diff Q(\thetav)$ in~\eqref{EqOp_f_ERM_RND2_pf_c_s2} is strictly decreasing with $a_1$. This implies that $a_1 > 0$ shall be chosen such that
\begin{IEEEeqnarray}{rCl}
	\Divf{P^{\star}}{Q} = \eta,
\end{IEEEeqnarray}
and justify the uniqueness of the solution.

For the case in which the empirical risk function $\foo{L}_{\dset{z}}$ in~\eqref{EqLxy} is nonseparable (see Definition \ref{Def_SeparableLxy}), the objective function in~\eqref{EqOp_f_ERM_RND2}~is a constant. Hence, the optimization problems in~\eqref{EqOp_f_ERMRERNormal} and~\eqref{EqOp_f_ERM_RND2} do not share the same solutions when for all $\thetav \in \supp Q$, $\foo{L}_{\dset{z}}(\thetav) = c$, for some $c > 0$ and $\foo{L}_{\dset{z}}$ in~\eqref{EqLxy}.
More specifically, the set of solutions to~\eqref{EqOp_f_ERMRERNormal} is the singleton $\{Q\}$, while the set of solutions to the problem in~\eqref{EqOp_f_ERM_RND2} is  $\{P \in \bigtriangleup_{Q}(\set{M}): \Divf{P}{Q} \leq \eta\}$. Thus, the problem is ill-posted and justifies the need for Assumption~\ref{assume:c} in Theorem~\ref{Theo_f_ERMRadNik}.

Essentially, choosing a real value $\lambda = \frac{a_1}{\eta}$, the real $a_2$ to satisfy 
\begin{equation}
\label{EqSetB_a_2}
a_2 \in \left\lbrace t\in \reals: \forall \vect{\theta} \in \supp Q , 0 <  \dot{f}^{-1} \left( -\frac{t + \foo{L}_{\vect{z}}(\thetav)}{\lambda} \right)\right\rbrace, 
\end{equation}
and denoting the solution $P^{\star}$ as $\Pgibbs{P}{Q}$, it holds that $g^{\star}$ in~\eqref{Eq_Op_RND_Min} can be written as $\frac{\diff \Pgibbs{P}{Q}}{\diff Q}$, and thus, for all $(\thetav) \in \supp Q$,
\begin{IEEEeqnarray}{rCl}
	\frac{\diff \Pgibbs{P}{Q}}{\diff Q} (\thetav) 
	& = & \dot{f}^{-1}(-\frac{\foo{L}_{\dset{z}}(\thetav)+a_2}{\lambda}),
\end{IEEEeqnarray}
where $\lambda$ is such that $\Divf{\Pgibbs{P}{Q}}{Q}=\eta$.
This completes the proof.
\end{IEEEproof}

\section{Proof of Lemma~\ref{lemm_mutuallyAbsCont}}
\begin{IEEEproof}
\label{AppProoflemm_mutuallyAbsCont}
	The probability measure $\Pgibbs{P}{Q}$ is absolutely continuous with respect to $Q$ from the fact that $\Pgibbs{P}{Q}\in\bigtriangleup_{Q}(\set{M})$. 
	Additionally, for all sets $\set{A} \in \field{F}$, such that $\Pgibbs{P}{Q}(\set{A}) = 0$, it holds that 
\begin{IEEEeqnarray}{rCl}
		0 & = &	\int_{\set{A}} \diff \Pgibbs{P}{Q}(\thetav)\\
		  & = & \int_{\set{A}} \frac{\diff \Pgibbs{P}{Q}}{\diff Q} ( \thetav )  \diff Q(\thetav),
	\end{IEEEeqnarray}
which implies, from the fact that $\frac{\diff \Pgibbs{P}{Q}}{\diff Q} >0 $ for all $\thetav \in \supp Q$, that  $Q(\set{A}) = 0$.
Thus, the probability measures $Q$ and $\Pgibbs{P}{Q}$ are mutually absolutely continuous. This completes the proof.
\end{IEEEproof}

\section{Proof of Theorem~\ref{Theo_InfDevKfDR}}
\begin{IEEEproof}
\label{AppProofLemmaInfDevKDivf}
The proof is divided into two parts.
The first part uses the properties of the $f$-divergence regularization to prove the properties of an ancillary function.
The second part proves that the normalization function $N_{Q, \dset{z}}:\set{A}_{Q, \dset{z}} \rightarrow \set{B}_{Q, \dset{z}}$, with $N_{Q, \dset{z}}$ implicitly defined in~\eqref{EqDefNormFunction}, has a unique solution and is continuous by using the implicit function theorem on the ancillary function properties.

Under the assumptions $(a)$, $(b)$ and $(c)$ from Theorem~\ref{Theo_f_ERMRadNik}, the sets $\set{A}_{Q, \dset{z}}$ and $\set{B}_{Q, \dset{z}}$ in~\eqref{EqDefNormFunction} are non-empty such that 
\begin{IEEEeqnarray}{rCl}
	\bar{a} & = & \sup \set{A}_{Q, \dset{z}},\\
	\underline{a} & = & \inf \set{A}_{Q, \dset{z}},\\
	\bar{b} & = & \sup \set{B}_{Q, \dset{z}}, \text{ and}\\
	\underline{b} & = & \inf \set{B}_{Q, \dset{z}},
\end{IEEEeqnarray}
such that 
\begin{subequations}
\label{EqDefABforF}
\begin{IEEEeqnarray}{rCl}
	\set{A} & = & (\underline{a},\bar{a}) \subseteq (0,\infty), \text{ and} \\
	\set{B} & = & (\underline{b},\bar{b}) \subseteq \reals.
\end{IEEEeqnarray}
\end{subequations}
Let the function $F: \set{A}\times \set{B} \to \reals$ be
\begin{IEEEeqnarray}{rcl}
\label{EqF4NQzLem5}
F(a,b) & = &  \int \dot{f}^{-1}(-\frac{b + \foo{L}_{\dset{z}}(\thetav)}{a}) \diff Q (\vect{\theta})-1.
\end{IEEEeqnarray}
The first part is as follows.
Given the function $F$ in~\eqref{EqF4NQzLem5}, the continuity of $F$ over the sets $\set{A}$ and $\set{B}$, defined in~\eqref{EqDefABforF}, is established by showing that $F$ exhibits a limit at every point in $\set{A}$ and $\set{B}$, respectively.
Note that due to the strict convexity and differentiability of $f$, satisfying the normalization constraint~\eqref{EqEqualToABigOne} requires the argument of $\dot{f}^{-1}$ (equivalently $\dot{f}^{\star}$) to lie strictly within the interior of the conjugate's domain. This guarantees the density remains finite. In other words, for all $(a,b) \in \set{A}\times \set{B}$ and $\vect{\theta} \in \supp Q$, it holds that
\begin{IEEEeqnarray}{rcl}
\label{EqThoseEyesISawToday}
\dot{f}^{-1}(\frac{-b - \foo{L}_{\dset{z}}(\thetav)}{a}) & \leq  & \dot{f}^{-1}(-\frac{b + \delta^\star_{Q, \dset{z}}}{a}) < \infty, 
\end{IEEEeqnarray}
where equality holds if and only if $ \foo{L}_{\dset{z}}(\thetav)  = \delta^\star_{Q, \dset{z}}$.
Now, from \cite[Corollary~24.5.1]{rockafellar1970conjugate} the function $\dot{f}^{-1}$ is continuous, such that for all $b \in \set{B}$, it holds that
\begin{IEEEeqnarray}{rcl}
\label{EqLimRNparB}
\lim_{b \to \beta} \dot{f}^{-1}(\frac{-b - \foo{L}_{\dset{z}}(\thetav)}{a}) & = &  \dot{f}^{-1}(\frac{-\beta - \foo{L}_{\dset{z}}(\thetav)}{a}).
\end{IEEEeqnarray}
Hence, from the dominated convergence theorem~\cite[Theorem~$1.6.9$]{ash2000probability}, the following limit exists and satisfies
\begin{IEEEeqnarray}{rcl}
\label{EqLimContFbeta}
\lim_{b \to \beta} F(a,b) & = & \lim_{b \to \beta}  \int \dot{f}^{-1}(-\frac{b + \foo{L}_{\dset{z}}(\thetav)}{a} ) \diff Q (\vect{\theta})-1\\
& = &  \int  (\lim_{b \to \beta} \dot{f}^{-1}(-\frac{b   + \foo{L}_{\dset{z}}(\thetav)}{a} ) )\diff Q (\vect{\theta})-1\\
& = & \int  \dot{f}^{-1}(-\frac{\beta + \foo{L}_{\dset{z}}(\thetav)}{a}  )\diff Q (\vect{\theta})-1\\
& = & F(a, \beta),
\end{IEEEeqnarray}
which proves that the function $F$ in~\eqref{EqF4NQzLem5} is continuous in $\set{B}$.
Similarly, from \cite[Corollary~24.5.1]{rockafellar1970conjugate} the function $\dot{f}^{-1}$ is continuous, such that for all $a \in \set{A}$, it holds that
\begin{IEEEeqnarray}{rcl}
\label{EqLimRNparA}
\lim_{a \to \lambda} \dot{f}^{-1}(\frac{-b - \foo{L}_{\dset{z}}(\thetav)}{a}) & = &  \dot{f}^{-1}(\frac{-b - \foo{L}_{\dset{z}}(\thetav)}{\lambda}).
\end{IEEEeqnarray}
Hence, from the dominated convergence theorem~\cite[Theorem~$1.6.9$]{ash2000probability}, the following limit exists and satisfies
\begin{IEEEeqnarray}{rcl}
\label{EqLimContFlambda}
\lim_{a \to \lambda} F(a,b) & = & \lim_{a \to \lambda}  \int \dot{f}^{-1}(\frac{-b - \foo{L}_{\dset{z}}(\thetav)}{a} ) \diff Q (\vect{\theta})-1\\
& = &  \int  (\lim_{a \to \lambda} \dot{f}^{-1}(\frac{-b - \foo{L}_{\dset{z}}(\thetav)}{a} ) )\diff Q (\vect{\theta})-1\\
& = & \int  \dot{f}^{-1}(\frac{-b - \foo{L}_{\dset{z}}(\thetav)}{\lambda}  )\diff Q (\vect{\theta})-1 \\
& = & F(\lambda,b),
\end{IEEEeqnarray}
which proves that the function $F$ in~\eqref{EqF4NQzLem5} is continuous in $\set{A}$.
Given a pair $(a,b) \in \set{A}_{Q, \dset{z}}\times\set{B}_{Q, \dset{z}}$, with $\set{A}_{Q, \dset{z}}$ in~\eqref{EqDefNormFunction}, assume that
\begin{equation}
\label{EqProofKrescalingL1}
N_{Q, \dset{z}}(a) = b.
\end{equation}
This implies that
\begin{IEEEeqnarray}{rCl}
0 
& = & \int \frac{\diff \Pgibbs[\dset{z}][b]{P}{Q}}{\diff Q}(\thetav) \diff Q(\thetav) - 1
\label{Eq_ProofLambdaIsTheInvOfNQz_s1}\\
& = & \int \dot{f}^{-1}(-\frac{b + \foo{L}_{\dset{z}}(\thetav)}{a})\diff Q(\thetav).
\label{Eq_ProofLambdaIsTheInvOfNQz_s2}\\
& = & F(a,b).
\end{IEEEeqnarray}
Note that the inverse $\dot{f}^{-1}$ exists from the fact that $f$ is strictly convex, which implies that $\dot{f}$ is a strictly increasing function. Hence, $\dot{f}^{-1}$ is also a strictly increasing function in $\set{B}_{Q, \dset{z}}$~\cite[Theorem 5.6.9]{bartle2000introduction}.
Moreover, from the assumption that $f$ is strictly convex and differentiable, it holds that $\dot{f}$ is continuous~\cite[Proposition $5.44$]{douchet2010Analyse}. This implies that $\dot{f}^{-1}$ is continuous.
From~Lemma~\ref{lemm_f_invIsInc} the function $\dot{f}^{-1}$ is strictly increasing such that for all $b \in \set{B}_{Q, \dset{z}}$ and for all $\thetav \in \supp Q$, it holds that
\begin{IEEEeqnarray}{rCl}
\label{Eq_ProofFinitenes_pf}	
\dot{f}^{-1}(- \frac{b +\foo{L}_{\dset{z}}(\thetav)}{a})
& \leq  & \dot{f}^{-1}(- \frac{b +\delta^\star_{Q, \dset{z}}}{a}),
\end{IEEEeqnarray}
with $\delta^\star_{Q, \dset{z}}$ defined in~\eqref{EqDefDeltaStar}.
Then, from~\eqref{Eq_ProofFinitenes_pf} it follows that
\begin{IEEEeqnarray}{rCl}
\int \dot{f}^{-1}(- \frac{b +\foo{L}_{\dset{z}}(\thetav)}{a})\diff Q(\thetav)
& < & \int \dot{f}^{-1}(- \frac{b + \delta^\star_{Q, \dset{z}}}{a})\diff Q(\thetav)
\label{Eq_ProofKbarNoLambdaIsFinite_pf}\\
& = & \dot{f}^{-1}(- \frac{b + \delta^\star_{Q, \dset{z}}}{a})\\
& < & \infty,\label{Eq_ProofKbarNoLambdaIsFinite_pf_s3}
\end{IEEEeqnarray}
where~\eqref{Eq_ProofKbarNoLambdaIsFinite_pf_s3} follows from $\set{A}_{Q, \dset{z}} \subseteq (0,\infty)$, which implies $a > 0$.
For all $(b_1, b_2) \in \set{B}_{Q, \dset{z}}^2$, such that $b_1 < b < b_2$, it holds that for all $\thetav \in \supp Q$,
\begin{IEEEeqnarray}{rCCCl}
	-\frac{1}{a}(\foo{L}_{\dset{z}}(\thetav) + b_1) 
	& > &-\frac{1}{a}(\foo{L}_{\dset{z}}(\thetav) + b) 
	& > & -\frac{1}{a}(\foo{L}_{\dset{z}}(\thetav) + b_2),
\end{IEEEeqnarray}
which from Lemma~\ref{lemm_f_invIsInc} implies that
\begin{IEEEeqnarray}{rCCCl}
\label{EqIneq_dotF4b1andb2}
	\dot{f}^{-1}(-\frac{1}{a}(\foo{L}_{\dset{z}}(\thetav) \!+\! b_1) ) \!& > &\! \dot{f}^{-1}(-\frac{1}{a}(\foo{L}_{\dset{z}}(\thetav) \!+\! b) )  \!& > &\! \dot{f}^{-1}(-\frac{1}{a}(\foo{L}_{\dset{z}}(\thetav) \!+\! b_2) ). \qquad
\end{IEEEeqnarray}
From~\eqref{EqIneq_dotF4b1andb2}, it holds that
\begin{equation}
	F(a,b_1) >  0 > F(a,b_2),
\end{equation}
which implies that the function $F$ in \eqref{EqF4NQzLem5} is strictly monotonic with respect to~$b$. 

The second part is as follows.
From the definition of $\set{A}$ and $\set{B}$ in~\eqref{EqDefABforF} there exists at least one point $(\lambda, \beta) \in \set{A}\times\set{B}$, such that 
\begin{IEEEeqnarray}{rcl}
 (\lambda, \beta) \in \set{A}_{Q, \dset{z}}\times \set{B}_{Q, \dset{z}}, 
\end{IEEEeqnarray}
which implies that
\begin{IEEEeqnarray}{rcl}
 F(\lambda,\beta) 
 	& = &  \int \dot{f}^{-1}(-\frac{\beta + \foo{L}_{\dset{z}}(\thetav)}{\lambda} ) \diff Q (\vect{\theta})-1\\
 	& = & \int \frac{\diff \Pgibbs{P}{Q}}{\diff Q}(\thetav) \diff Q (\vect{\theta})-1\\
 	& = & \int \diff \Pgibbs{P}{Q} (\vect{\theta})-1\\
 	& = & 0.
\end{IEEEeqnarray}
Note that from~\eqref{EqLimContFbeta} and~\eqref{EqLimContFlambda} the function $F$ is continuous and thus the partial derivative of $F$ satisfy
\begin{IEEEeqnarray}{rcl}
 \label{EqDeffaFab}
 \frac{\partial}{\partial a}F(a,b) 
 	& = &  \frac{\partial}{\partial a} (\int \dot{f}^{-1}(-\frac{b + \foo{L}_{\dset{z}}(\thetav)}{a} ) \diff Q (\vect{\theta})-1)\\
 	& = & \int \frac{\partial}{\partial a} \dot{f}^{-1}(-\frac{b + \foo{L}_{\dset{z}}(\thetav)}{a} ) \diff Q (\vect{\theta})\\
 	& = & \int \frac{\diff }{\diff a} \dot{f}^{-1}(-\frac{b + \foo{L}_{\dset{z}}(\thetav)}{a} ) \diff Q (\vect{\theta})\\
 	& = & \int \frac{b + \foo{L}_{\dset{z}}(\thetav)}{a^{2}}\frac{1}{\ddot{f}(\dot{f}^{-1}(-\frac{b + \foo{L}_{\dset{z}}(\thetav)}{a}))} \diff Q (\vect{\theta}), \label{EqDeffaFab_s4}
\end{IEEEeqnarray}
where~\eqref{EqDeffaFab_s4} follows from Lemma~\ref{lemm_f_NormInv}; and
\begin{IEEEeqnarray}{rcl}
 \label{EqDeffbFab}
 \frac{\partial}{\partial b}F(a,b) 
 	& = &  \frac{\partial}{\partial b} (\int \dot{f}^{-1}(-\frac{b + \foo{L}_{\dset{z}}(\thetav)}{a} ) \diff Q (\vect{\theta})-1)\\
 	& = & \int \frac{\partial}{\partial b} \dot{f}^{-1}(-\frac{b + \foo{L}_{\dset{z}}(\thetav)}{a} ) \diff Q (\vect{\theta})\\
 	& = & \int \frac{\diff}{\diff b} \dot{f}^{-1}(-\frac{b + \foo{L}_{\dset{z}}(\thetav)}{a} ) \diff Q (\vect{\theta})\\
 	& = & \int -\frac{1}{a}\frac{1}{\ddot{f}(\dot{f}^{-1}(-\frac{b + \foo{L}_{\dset{z}}(\thetav)}{a}))} \diff Q (\vect{\theta}),\label{EqDeffbFab_s4}
\end{IEEEeqnarray}
where~\eqref{EqDeffbFab_s4} follows from Lemma~\ref{lemm_f_NormInv}.
Note that, for all $(a,b)\times \set{A}\times \set{B}$, the partial derivative $\frac{\partial}{\partial b}F$, satisfies,
\begin{equation}
	\frac{\partial}{\partial b}F(a,b) < 0,
\end{equation}
The fact that $a>0$, together with the strict convexity and twice differentiability of $f$, implies that for all $u \in \reals$, the function $\ddot{f}$ satisfies, $\ddot{f}(u) > 0 $.
Then, from \emph{The Implicit Function Theorem} in \cite[Theorem 4]{oswaldo2014TIFT} the function $N_{Q, \dset{z}}$ exists and is unique in the open interval $\set{A}$ with $\set{A}$ in~\eqref{EqDefABforF} and for all $a \in \set{A}$ satisfies
\begin{IEEEeqnarray}{rcl}
N_{Q, \dset{z}}(a) & = &  b, 
\end{IEEEeqnarray}
such that 
\begin{IEEEeqnarray}{rcl}
F(a,N_{Q, \dset{z}}(a)) & = &  0, 
\end{IEEEeqnarray}
which completes the proof of continuity and uniqueness for the normalization function $N_{Q, \dset{z}}$.

\end{IEEEproof}

\section{Proof of Theorem~\ref{Theo_ODE_NQz}}
\label{app_proof_theo_ODE_NQz}
\begin{IEEEproof}
\label{AppProofTheo_ODE_NQz}
Recall the function $F$ defined in~\eqref{EqF4NQzLem5}, together with its partial derivatives $\frac{\partial F}{\partial a}$ and $\frac{\partial F}{\partial b}$, introduced in~\eqref{EqDeffaFab} and~\eqref{EqDeffbFab}, respectively. Then, from Theorem~\ref{Theo_InfDevKfDR} and~\cite[Theorem~4]{oswaldo2014TIFT}, the derivative of the normalization function $N_{Q, \dset{z}}$ satisfies
\begin{IEEEeqnarray}{rcl}
\!\!\frac{\diff }{\diff a}N_{Q, \dset{z}}(a)
& = & - \! (\!\frac{\partial}{\partial b}F(a,b)\!)^{-1}\!\!\frac{\partial}{\partial a}\!F(a,b\!), \qquad\\
& = &-\frac{\displaystyle\int \frac{b + \foo{L}_{\dset{z}}(\thetav)}{a^{2}}\frac{1}{\ddot{f}(\dot{f}^{-1}(-\frac{b + \foo{L}_{\dset{z}}(\thetav)}{a}))} \diff Q (\thetav)}{\displaystyle\int -\frac{1}{a}\frac{1}{\ddot{f}(\dot{f}^{-1}(-\frac{b + \foo{L}_{\dset{z}}(\nuv)}{a}))} \diff Q (\nuv)} \\
& = &\frac{\displaystyle\int \frac{b + \foo{L}_{\dset{z}}(\thetav)}{a}\frac{1}{\ddot{f}(\dot{f}^{-1}(-\frac{b + \foo{L}_{\dset{z}}(\thetav)}{a}))} \diff Q (\thetav)}{\displaystyle\int \frac{1}{\ddot{f}(\dot{f}^{-1}(-\frac{b + \foo{L}_{\dset{z}}(\nuv)}{a}))} \diff Q (\nuv)} \\
& = &\frac{\displaystyle\int \frac{b + \foo{L}_{\dset{z}}(\thetav)}{a}(\ddot{f}(\frac{\diff \Pgibbs[\dset{z}][a]{P}{Q}}{\diff Q}(\thetav)))^{-1} \diff Q (\thetav)}{\displaystyle\int (\ddot{f}(\frac{\diff \Pgibbs[\dset{z}][a]{P}{Q}}{\diff Q}(\nuv)))^{-1} \diff Q (\nuv)} \\
& = &\frac{b}{a}+ \frac{1}{a}\frac{\displaystyle\int\foo{L}_{\dset{z}}(\thetav)(\ddot{f}(\frac{\diff \Pgibbs[\dset{z}][a]{P}{Q}}{\diff Q}(\thetav)))^{-1} \diff Q (\thetav)}{\displaystyle\int (\ddot{f}(\frac{\diff \Pgibbs[\dset{z}][a]{P}{Q}}{\diff Q}(\nuv)))^{-1} \diff Q (\nuv)}\\
& = &\frac{b}{a}+ \frac{1}{a}\frac{\displaystyle\int\foo{L}_{\dset{z}}(\thetav)(\ddot{f}(\frac{\diff \Pgibbs[\dset{z}][a]{P}{Q}}{\diff Q}(\thetav)))^{-1} \diff Q (\thetav)}{\displaystyle\int (\ddot{f}(\frac{\diff \Pgibbs[\dset{z}][a]{P}{Q}}{\diff Q}(\nuv)))^{-1} \diff Q (\nuv)}\\
& = & \frac{b}{a}+ \displaystyle\int \frac{\foo{L}_{\dset{z}}(\thetav)}{a}g_a(\thetav)\diff Q (\thetav),\label{EqProofNzFuncDef_s7}\\
& = & \frac{N_{Q, \dset{z}}(a)}{a}+ \displaystyle\int \frac{\foo{L}_{\dset{z}}(\thetav)}{a}g_a(\thetav)\diff Q (\thetav),\label{EqProofNzFuncDef_s8}
\end{IEEEeqnarray}
with the function $g_a:\set{M} \to \reals$, such that for all $\thetav \in \supp Q$
\begin{IEEEeqnarray}{rcl}
\label{EqRNnewRm}
	g_a(\thetav) & = & \frac{\Bigg(\displaystyle\ddot{f}\Bigg(\frac{\diff \Pgibbs[\dset{z}][a]{P}{Q}}{\diff Q}(\thetav)\Bigg)\Bigg)^{-1}}{\displaystyle\int\! \Bigg(\ddot{f}\Bigg(\frac{\diff \Pgibbs[\dset{z}][a]{P}{Q}}{\diff Q}(\nuv)\Bigg)\Bigg)^{-1}\!\!\!\!\! \diff Q (\nuv)}.
\end{IEEEeqnarray}
Note that from the assumption that $f$ is strictly convex and twice differentiable, the derivative $\dot{f}$ is increasing, and the second derivative $\ddot{f}$ is positive for all $\thetav \in \supp Q$. 
The denominator is the integral of the numerator with respect to $Q$, and thus, it serves as a normalization constant. 
This ensures that $g_a$ constitutes a valid Radon-Nikodym derivative with respect to $Q$ that satisfies
\begin{IEEEeqnarray}{rcl}
	\int g_a(\thetav) \diff Q (\thetav)& = & 1.
\end{IEEEeqnarray}
Therefore, the function $g_a$ in~\eqref{EqRNnewRm} can be interpreted as the Radon-Nikodym derivative of a new probability measure $P^{(a)}$, parameterized by the regularization factor $a$ with respect to $Q$. Specifically, if the measure $P^{(a)}$ is defined such that for any set $\set{A} \in \field{F}_{\set{M}}$,
\begin{IEEEeqnarray}{rcl}
\label{EqGaIsMeasureP}
	P^{(a)}(\set{A}) = \int_{\set{A}} g_a(\thetav) \diff Q(\thetav).
\end{IEEEeqnarray}
From~\eqref{EqProofNzFuncDef_s8} and~\eqref{EqGaIsMeasureP}, it follows that
\begin{IEEEeqnarray}{rcl}
	N_{Q, \dset{z}}(a) 
	& = & a \frac{\diff }{\diff a}N_{Q, \dset{z}}(a) -\displaystyle\int\foo{L}_{\dset{z}}(\thetav)\frac{\diff P^{(a)}}{\diff Q}(\thetav) \diff Q (\thetav)\\
	& = & a \frac{\diff }{\diff a}N_{Q, \dset{z}}(a) - \foo{R}_{\dset{z}}(P^{(a)}),
\end{IEEEeqnarray}
with $\foo{R}_{\dset{z}}$ defined in~\eqref{EqRxy}.
This completes the proof.
\end{IEEEproof}

\section{Proof of Lemma~\ref{lemm_NQz_decreas}}
\label{app_proof_lemm_monotologcvx}
\begin{IEEEproof}
	From the implicit function theorem the derivative of the normalization function satisfies 
	\begin{IEEEeqnarray}{rcl}
\!\!\frac{\diff }{\diff \lambda}N_{Q, \dset{z}}(\lambda)
& = & - \! (\!\frac{\partial}{\partial \beta}F(\lambda,\beta)\!)^{-1}\!\!\frac{\partial}{\partial \lambda}\!F(\lambda,\beta\!), \qquad\\
	& = &\frac{\displaystyle\int \frac{\beta + \foo{L}_{\dset{z}}(\thetav)}{\lambda}(\ddot{f}(\frac{\diff \Pgibbs[\dset{z}][\lambda]{P}{Q}}{\diff Q}(\thetav)))^{-1} \diff Q (\thetav)}{\displaystyle\int (\ddot{f}(\frac{\diff \Pgibbs[\dset{z}][\lambda]{P}{Q}}{\diff Q}(\nuv)))^{-1} \diff Q (\nuv)} \\
	& = &\frac{\displaystyle\int -\dot{f}(\frac{\diff \Pgibbs[\dset{z}][\lambda]{P}{Q}}{\diff Q}(\thetav))(\ddot{f}(\frac{\diff \Pgibbs[\dset{z}][\lambda]{P}{Q}}{\diff Q}(\thetav)))^{-1} \diff Q (\thetav)}{\displaystyle\int (\ddot{f}(\frac{\diff \Pgibbs[\dset{z}][\lambda]{P}{Q}}{\diff Q}(\nuv)))^{-1} \diff Q (\nuv)}\\
	& = & \frac{\displaystyle\int -\dot{f}(\frac{\diff \Pgibbs[\dset{z}][\lambda]{P}{Q}}{\diff Q}(\thetav))\ddot{f^{*}}(\dot{f}(\frac{\diff \Pgibbs[\dset{z}][\lambda]{P}{Q}}{\diff Q}(\thetav))) \diff Q (\thetav)}{\displaystyle\int \ddot{f^{*}}(\dot{f}(\frac{\diff \Pgibbs[\dset{z}][\lambda]{P}{Q}}{\diff Q}(\nuv))) \diff Q (\nuv)}\label{EqNderivativeInLFtransfom}
	\end{IEEEeqnarray}
	Let the function $\psi:\reals \to \reals$ be 
	\begin{equation}
	\label{EqLogDerivative}
	\psi(y) = \frac{\ddot{f^{*}}(y)}{\dot{f^{*}}(y)} = \frac{\diff}{\diff y} \log(\dot{f^{*}}(y)).
\end{equation}
	Then, from \eqref{EqNderivativeInLFtransfom} and \eqref{EqLogDerivative}, \begin{IEEEeqnarray}{rcl}
\!\!\frac{\diff }{\diff \lambda}N_{Q, \dset{z}}(\lambda)
& = &  \frac{\displaystyle\int \frac{\foo{L}_{\dset{z}}(\thetav)+\beta}{\lambda}\psi(-\frac{\foo{L}_{\dset{z}}(\thetav)+\beta}{\lambda})\diff \Pgibbs[\dset{z}][\lambda]{P}{Q}(\thetav)}{\displaystyle\int \psi(-\frac{\foo{L}_{\dset{z}}(\thetav)+\beta}{\lambda})\diff \Pgibbs[\dset{z}][\lambda]{P}{Q}(\thetav)}.\label{EqNormFunctionNewForm}
	\end{IEEEeqnarray}
	Note that under Assumptions \ref{assume:a}, it holds that for all $v \in \set{J}$, the functions $\dot{f}^{-1}$ and $\dot{f^{*}}$ satisfy $\dot{f}^{-1}(t)=\dot{f^{*}}(t)$. Hence, together with the assumption that $\dot{f}^{-1}$ is log convex, it implies that the function $\psi$ is increasing. Therefore, the map $\foo{L}_{\dset{z}}(\thetav) \mapsto \psi(-\frac{\foo{L}_{\dset{z}}(\thetav)+\beta}{\lambda})$ is decreasing with respect to the empirical risk $\foo{L}_{\dset{z}}(\thetav)$.
	Now consider the integral 
	\begin{equation}
		 \int \frac{\foo{L}_{\dset{z}}(\thetav)+\beta}{\lambda}\psi(-\frac{\foo{L}_{\dset{z}}(\thetav)+\beta}{\lambda})\diff \Pgibbs[\dset{z}][\lambda]{P}{Q}(\thetav).
	\end{equation}
	This quantity can be analyzed and upper-bounded by evaluating the covariance between the functions $\frac{\foo{L}_{\dset{z}}(\thetav)+\beta}{\lambda}$ and $\psi(-\frac{\foo{L}_{\dset{z}}(\thetav)+\beta}{\lambda})$ under the measure $\Pgibbs{P}{Q}$.
	Since models $\thetav \in \set{M}$ are sampled independently and identically according to the probability measure $\Pgibbs{P}{Q}$, it follows that 
	\begin{IEEEeqnarray}{rCl}
		\IEEEmulticolR{
		\frac{1}{2}\!\int\!\int\!(\frac{\foo{L}_{\dset{z}}(\thetav)\!+\!\beta}{\lambda}\!-\!\frac{\foo{L}_{\dset{z}}(\nuv)\!+\!\beta}{\lambda}) \Bigg(\!\psi(-\frac{\foo{L}_{\dset{z}}(\thetav)\!+\!\beta}{\lambda})\! 
		}
		\IEEEmulticolR{
		-\!\psi(-\frac{\foo{L}_{\dset{z}}(\nuv)\!+\!\beta}{\lambda})\Bigg)\diff \Pgibbs{P}{Q}(\thetav)\diff \Pgibbs[\dset{z}][\lambda]{P}{Q}(\nuv)
		}
		& = & \frac{1}{2}(\int \frac{\foo{L}_{\dset{z}}(\thetav)+\beta}{\lambda}\psi(-\frac{\foo{L}_{\dset{z}}(\thetav)+\beta}{\lambda})\diff \Pgibbs{P}{Q}(\thetav) 
		\splitR[1]
		+\int \frac{\foo{L}_{\dset{z}}(\nuv)+\beta}{\lambda}\psi(-\frac{\foo{L}_{\dset{z}}(\nuv)+\beta}{\lambda})\diff \Pgibbs{P}{Q}(\nuv)
		\splitR[1]
		-\int\int \frac{\foo{L}_{\dset{z}}(\thetav)+\beta}{\lambda}\psi(-\frac{\foo{L}_{\dset{z}}(\nuv)+\beta}{\lambda}) \diff \Pgibbs{P}{Q}(\thetav) \diff \Pgibbs{P}{Q}(\nuv) 
		\splitR[1]
		- \int\int \frac{\foo{L}_{\dset{z}}(\nuv)+\beta}{\lambda}\psi(-\frac{\foo{L}_{\dset{z}}(\thetav)+\beta}{\lambda}) \diff \Pgibbs{P}{Q}(\thetav) \diff \Pgibbs{P}{Q}(\nuv))\\
		& = & \int \frac{\foo{L}_{\dset{z}}(\thetav)+\beta}{\lambda}\psi(-\frac{\foo{L}_{\dset{z}}(\thetav)+\beta}{\lambda})\diff \Pgibbs{P}{Q}(\thetav)
		\splitR
		- \int\frac{\foo{L}_{\dset{z}}(\thetav)+\beta}{\lambda}\diff\Pgibbs{P}{Q}(\thetav) \int \psi(-\frac{\foo{L}_{\dset{z}}(\thetav)+\beta}{\lambda})\Pgibbs{P}{Q}(\thetav).
		\label{EqLogCovariance}
	\end{IEEEeqnarray}
	Observe that for all $(\thetav, \nuv) \in \supp Q$ satisfying $\foo{L}_{\dset{z}}(\nuv) \geq \foo{L}_{\dset{z}}(\thetav)$, it follows that
	\begin{equation}
		\psi(-\frac{\foo{L}_{\dset{z}}(\thetav)+\beta}{\lambda})\leq \psi(-\frac{\foo{L}_{\dset{z}}(\nuv)+\beta}{\lambda}),
	\end{equation} 
	which implies 
	\begin{equation}
	\label{EqIneqCovCase1IsZero}
		(\frac{\foo{L}_{\dset{z}}(\thetav)+\beta}{\lambda}-\frac{\foo{L}_{\dset{z}}(\nuv)+\beta}{\lambda})(\psi(-\frac{\foo{L}_{\dset{z}}(\thetav)+\beta}{\lambda})-\psi(-\frac{\foo{L}_{\dset{z}}(\nuv)+\beta}{\lambda})) \leq 0.
	\end{equation} 
	Similarly, for all $(\thetav, \nuv) \in \supp Q$ such that $\foo{L}_{\dset{z}}(\thetav) < \foo{L}_{\dset{z}}(\nuv)$, it holds that 
	\begin{equation}
		\psi(-\frac{\foo{L}_{\dset{z}}(\thetav)+\beta}{\lambda})>\psi(-\frac{\foo{L}_{\dset{z}}(\nuv)+\beta}{\lambda}),
	\end{equation}
	which implies
	\begin{equation}
	\label{EqIneqCovCase2IsZero}
		(\frac{\foo{L}_{\dset{z}}(\thetav)+\beta}{\lambda}-\frac{\foo{L}_{\dset{z}}(\nuv)+\beta}{\lambda})(\psi(-\frac{\foo{L}_{\dset{z}}(\thetav)+\beta}{\lambda})-\psi(-\frac{\foo{L}_{\dset{z}}(\nuv)+\beta}{\lambda})) < 0.
	\end{equation}
	Using \eqref{EqLogCovariance}, \eqref{EqIneqCovCase1IsZero} and \eqref{EqIneqCovCase2IsZero}
	\begin{IEEEeqnarray}{rCl}
		\IEEEmulticolR{
		\int \frac{\foo{L}_{\dset{z}}(\thetav)+\beta}{\lambda}\psi(-\frac{\foo{L}_{\dset{z}}(\thetav)+\beta}{\lambda})\diff \Pgibbs{P}{Q}(\thetav)
		}
		& \leq & \int\frac{\foo{L}_{\dset{z}}(\thetav)+\beta}{\lambda}\diff\Pgibbs{P}{Q}(\thetav) \int \psi(-\frac{\foo{L}_{\dset{z}}(\thetav)+\beta}{\lambda})\Pgibbs{P}{Q}(\thetav).
		\label{EqIneqLogConvexForm}
	\end{IEEEeqnarray}
	From \eqref{EqNormFunctionNewForm} and \eqref{EqIneqLogConvexForm} it follows that 
	\begin{IEEEeqnarray}{rcl}
\!\!\frac{\diff }{\diff \lambda}N_{Q, \dset{z}}(\lambda)
& \leq &  \frac{\displaystyle\int \frac{\foo{L}_{\dset{z}}(\thetav)\!+\!\beta}{\lambda}\diff \Pgibbs{P}{Q}(\thetav)\!\int\!\psi(-\frac{\foo{L}_{\dset{z}}(\thetav)\!+\!\beta}{\lambda})\diff \Pgibbs[\dset{z}][\lambda]{P}{Q}(\thetav)}{\displaystyle\int \psi(-\frac{\foo{L}_{\dset{z}}(\thetav)\!+\!\beta}{\lambda})\diff \Pgibbs[\dset{z}][\lambda]{P}{Q}(\thetav)}\\
	& = & \int \frac{\foo{L}_{\dset{z}}(\thetav)+\beta}{\lambda}\diff \Pgibbs{P}{Q}(\thetav)\\
	& = & \int -\dot{f}(\frac{\diff \Pgibbs{P}{Q}}{\diff Q}(\thetav))\diff \Pgibbs{P}{Q}(\thetav)\\
	& = &- \int \frac{\diff \Pgibbs{P}{Q}}{\diff Q}(\thetav) \dot{f}(\frac{\diff \Pgibbs{P}{Q}}{\diff Q}(\thetav))\diff Q(\thetav).\label{EqIntdPdQdotF}
	\end{IEEEeqnarray}
	From Lemma~\ref{lemm_convxdotf} in Appendix~\ref{sec:AppendixA} and \eqref{EqIntdPdQdotF}, it follows that
	\begin{IEEEeqnarray}{rcl}
\!\!\frac{\diff }{\diff \lambda}N_{Q, \dset{z}}(\lambda)
& \leq & 0.
	\end{IEEEeqnarray}
	which completes the proof.
\end{IEEEproof}

\section{Proof of Lemma~\ref{lemm_fDR_kset}}
\label{app_proof_lemm_fDR_kset}
\begin{IEEEproof}
\label{proof_lemm_fDR_kset}
The proof is divided into two parts. 
In the first part, the strictly increasing property of the Radon-Nikodym derivative, obtained from the connection established with the Legendre-Fenchel transform of $f$, is used to evaluate the real values of $\lambda$ under which assumption~\eqref{EqDefSetB} holds.
In the second part, the strictly increasing property is used to evaluate the real values of $\lambda$ under which assumption~\eqref{EqEqualToABigOne} holds.

The first part is as follows.
Lemma~\ref{lemm_f_LFTequality} in Appendix~\ref{sec:AppendixA}, together with \cite[Corollary~23.5.1]{rockafellar1970conjugate}, implies that for all $\thetav \in \supp Q$,
\begin{IEEEeqnarray}{rCl}
\label{Eq_convcnj4_pf}
	\frac{\diff \Pgibbs{P}{Q}}{\diff Q}(\thetav) & = & \dot{f^{\star}}(-\frac{\foo{L}_{\dset{z}}(\thetav)+\beta}{\lambda})\\
	& = &\dot{f}^{-1}(-\frac{\foo{L}_{\dset{z}}(\thetav)+\beta}{\lambda}).
\end{IEEEeqnarray}
Evaluating the real values of $\lambda$ under which assumption in~\eqref{EqDefSetB} holds requires to show that the function $\frac{\diff \Pgibbs{P}{Q}}{\diff Q}$ belongs to the set of nonnegative measurable functions.
From the $f$-divergence in Defintion~\ref{Def_fDivergence} and the fact that $\dot{f}^{-1}$ strictly increasing and bijective, the proof follows by showing that the limit 
\begin{IEEEeqnarray}{rCl}
\label{Eq_limdf_at_0}
	\lim_{u\rightarrow 0^{+}}\dot{f}(u) & = & v_0,
\end{IEEEeqnarray}
satisfies for all $\thetav \in \supp Q$,
\begin{IEEEeqnarray}{rCl}
\label{Eq_limdf_argument}
	-\frac{\foo{L}_{\dset{z}}(\thetav)+\beta}{\lambda} & > & v_0.
\end{IEEEeqnarray}
Note that~\eqref{Eq_limdf_argument} is suficient from the fact that the monotonicity of $\dot{f}^{-1}$ implies that for all $v > v_0$, 
\begin{IEEEeqnarray}{rCl}
	\dot{f}^{-1}(v) & > & 0.
\end{IEEEeqnarray}
To evaluate the real values of $\lambda$ under which assumption in~\eqref{EqDefSetB} holds, two cases must be considered for the limit in~\eqref{Eq_limdf_at_0}.

{\bf Case 1:} Assume that
\begin{IEEEeqnarray}{rCl}
\label{Eq_limdf_at_0_a}
	\lim_{u\rightarrow 0^{+}}\dot{f}(u) & = & a,
\end{IEEEeqnarray}
where $a \in \reals$. Under the above assumption, consider the set
\begin{IEEEeqnarray}{rCl}
\label{Eq_limdf_at_0_a_set}
	\set{D} & = & \{\thetav \in \supp Q: -\foo{L}_{\dset{z}}(\thetav) < a\lambda + \beta\}.
\end{IEEEeqnarray}
On one hand, note that if the function $\foo{L}_{\dset{z}}$ in~\eqref{EqLxy} is unbounded in $\supp Q$, from~\eqref{Eq_limdf_at_0_a} the set $\set{D}$ in~\eqref{Eq_limdf_at_0_a_set} is nonegligble and measurable, such that for all $\thetav \in \set{D}$,
\begin{IEEEeqnarray}{rCl}
	-\frac{\foo{L}_{\dset{z}}(\thetav)+\beta}{\lambda} & < & a,
\end{IEEEeqnarray}
which implies that 
\begin{IEEEeqnarray}{rCl}
	\dot{f}^{-1}(-\frac{\foo{L}_{\dset{z}}(\thetav)+\beta}{\lambda}) & < & 0.
\end{IEEEeqnarray}
Hence, Assumption $(b)$ of Theorem~\ref{Theo_f_ERMRadNik} is not satisfied and nothing can be stated about the solution.
On the other hand, if the function $\foo{L}_{\dset{z}}$ in~\eqref{EqLxy} is bounded in $\supp Q$, such that
\begin{IEEEeqnarray}{rCl}
\label{Eq_supMcase2}
	M & = & \sup_{\thetav \in \supp Q} \foo{L}_{\dset{z}}(\thetav).
\end{IEEEeqnarray}
Then, there exists a $\lambda_{Q, \dset{z}} \in (0,\infty)$ such that
\begin{IEEEeqnarray}{rCl}
\label{EqlambdaStarp1}
	-M & = & a\lambda_{Q, \dset{z}}+\beta.
\end{IEEEeqnarray}
From~\eqref{EqlambdaStarp1} for all $\lambda > \lambda_{Q, \dset{z}}$, it holds that for all $\thetav \in \supp Q$,
\begin{IEEEeqnarray}{rCl}
\label{EqconstNonegCase2}
	\dot{f}^{-1}(-\frac{\foo{L}_{\dset{z}}(\thetav)+\beta}{\lambda}) & > & 0.
\end{IEEEeqnarray}
From~\eqref{EqconstNonegCase2}, consider the following conditions:
If there exists a model $\bar{\thetav} \in \supp Q$ such that $\foo{L}_{\dset{z}}(\bar{\thetav}) = M$, where $M$ is defined in~\eqref{Eq_supMcase2}, then the set of regularization factors $\lambda$ for which the function $\frac{\diff \Pgibbs{P}{Q}}{\diff Q}$ is nonnegative is $(\lambda_{Q, \dset{z}}, \infty)$.
Alternatively, if for all models $\thetav \in \supp Q$, it holds that $\foo{L}_{\dset{z}}(\thetav) < M$, where $M$ is defined in~\eqref{Eq_supMcase2}, then the set of regularization factors $\lambda$ for which the function $\frac{\diff \Pgibbs{P}{Q}}{\diff Q}$ is nonnegative is $[\lambda_{Q, \dset{z}}, \infty)$.

{\bf Case 2:} Assume that
\begin{IEEEeqnarray}{rCl}
\label{Eq_limdf_at_0_ninfty}
	\lim_{u\rightarrow 0^{+}}\dot{f}(u) & = & -\infty.
\end{IEEEeqnarray}
Under the above assumption, for all $\thetav \in \supp Q$,
\begin{IEEEeqnarray}{rCl}
\label{Eq_limdf_argumentTrue}
	-\frac{\foo{L}_{\dset{z}}(\thetav)+\beta}{\lambda} & > & -\infty,
\end{IEEEeqnarray}
which implies that
\begin{IEEEeqnarray}{rCl}
	\dot{f}^{-1}(-\frac{\foo{L}_{\dset{z}}(\thetav)+\beta}{\lambda}) & > & 0.
\end{IEEEeqnarray}
Hence, for all $\lambda \in (0,\infty)$, Assumption~\ref{assume:b} of Theorem~\ref{Theo_f_ERMRadNik} is satisfied, guaranteeing the nonnegativity of the function $\frac{\diff \Pgibbs{P}{Q}}{\diff Q}$.
This completes the first part of the proof.

The second part is as follows.
Evaluating the values $\lambda$ under which assumption~\eqref{EqEqualToABigOne} holds requires to show there exists a real value $\beta \in \reals$ such that the integral of $\frac{\diff \Pgibbs{P}{Q}}{\diff Q}$ with respect to $Q$ is one. 
From Lemma~\ref{Theo_InfDevKfDR} the monotonicity of the normalization function $N_{Q, \dset{z}}$ in~\eqref{EqDefNormFunction} there is a minimum regularization factor $\lambda^{\star}_{Q, \dset{z}}$ defined in~\eqref{EqDefLambdaStar}. Furthemore, from Lemma~\ref{Theo_InfDevKfDR} the continuity of the function $N_{Q, \dset{z}}$ implies that for all $\lambda \in (\lambda^{\star}_{Q, \dset{z}}, \infty)$, there exists a unique $\beta \in \set{B}_{Q, \dset{z}}$ such that Assumption $(b)$ of Theorem~\ref{Theo_f_ERMRadNik} is satisfied.
From Lemma~\ref{Theo_InfDevKfDR}, it holds that
\begin{IEEEeqnarray}{rCl}
\lim_{\lambda \rightarrow {\lambda^{\star}_{Q, \dset{z}}}^{+}} N_{Q, \dset{z}}(\lambda) & = & N_{Q, \dset{z}}(\lambda^{\star}_{Q, \dset{z}}),
\end{IEEEeqnarray}
with the function $N_{Q, \dset{z}}$ defined in~\eqref{EqDefNormFunction} and the limit from the right is well defined from the fact that the set $\set{A}_{Q, \dset{z}}$ is convex. 
To determine whether the infimum in~\eqref{EqDefLambdaStar} belongs to the set $\set{A}_{Q, \dset{z}}$ two cases are considered.

{\bf Case 1:} Assume that $\beta > N_{Q, \dset{z}}(\lambda^{\star}_{Q, \dset{z}})$, such that
\begin{IEEEeqnarray}{rCl}
\label{EqIntAssumption1}
	\int \dot{f}^{-1}(-\frac{\beta + \foo{L}_{\dset{z}}(\thetav)}{\lambda^{\star}_{Q, \dset{z}}})\diff Q(\thetav)
	& = & \infty.
\end{IEEEeqnarray}
Notice that from~\eqref{EqIntAssumption1} for all $\beta_1 \in [ N_{Q, \dset{z}}(\lambda^{\star}_{Q, \dset{z}}), \beta)$ and for all $\thetav \in \supp Q$, it holds that
\begin{IEEEeqnarray}{rCl}
\label{EqIntAssumption1_prt2}
	\dot{f}^{-1}(-\frac{\beta_1 + \foo{L}_{\dset{z}}(\thetav)}{\lambda^{\star}_{Q, \dset{z}}})
	& > & \dot{f}^{-1}(-\frac{\beta + \foo{L}_{\dset{z}}(\thetav)}{\lambda^{\star}_{Q, \dset{z}}}).
\end{IEEEeqnarray}
Hence, under this assumption, $N_{Q, \dset{z}}(\lambda^{\star}_{Q, \dset{z}}) \notin \set{B}_{Q, \dset{z}}$. This implies that the set $\set{A}_{Q, \dset{z}}$ of regularization parameters that satisfy~\eqref{EqEqualToABigOne} is given by $\set{A}_{Q, \dset{z}} = (\lambda^{\star}_{Q, \dset{z}},\infty)$.

{\bf Case 2:} Assume that $\beta > N_{Q, \dset{z}}(\lambda^{\star}_{Q, \dset{z}})$, such that
\begin{IEEEeqnarray}{rCl}
\label{EqIntAssumption2}
	\int \dot{f}^{-1}(-\frac{\beta + \foo{L}_{\dset{z}}(\thetav)}{\lambda^{\star}_{Q, \dset{z}}})\diff Q(\thetav)
	& < & \infty.
\end{IEEEeqnarray}
From the monotonicity of the solution in part one and continuity of the function $N_{Q, \dset{z}}$ in~\eqref{EqDefNormFunction} from Lemma~\ref{Theo_InfDevKfDR}, there exists a $\beta^{\star}_{Q, \dset{z}} \in \set{B}_{Q, \dset{z}}$  such that $N_{Q, \dset{z}}(\lambda^{\star}_{Q, \dset{z}})=\beta^{\star}_{Q, \dset{z}}$, which implies that the set of all regularization factors $\set{A}_{Q, \dset{z}} = [\lambda^{\star}_{Q, \dset{z}},\infty)$.

Finally, from parts two and three of the proof the set $\set{A}_{Q, \dset{z}}$ is a convex set such that the regularization factors for which the assumptions of Theorem~\ref{Theo_f_ERMRadNik} hold and are given by 
\begin{IEEEeqnarray}{rCl}
 (\lambda^{\star}_{Q, \dset{z}}, \infty) \subseteq \set{A}_{Q, \dset{z}} \subseteq [\lambda^{\star}_{Q, \dset{z}}, \infty),
\end{IEEEeqnarray}
which completes the proof.

\end{IEEEproof}
\section{Proof of Lemma~\ref{lemm_fDR_No_minRegF_nneg}}
\label{app_proof_lemm_fDR_No_minRegF_nneg}
\begin{IEEEproof}
\label{proof_lemm_fDR_No_minRegF_nneg}
The following proof is divided into two parts. In the first part, an auxiliary function is introduced and proven to be continuous. In the second part, relying on the assumption that $\dot{f}^{-1}$ is nonnegative and the previously established continuity of the auxiliary function, the intermediate value theorem is applied to prove the existence of a unique solution for any given regularization factor. Consequently, it is shown that the set of admissible regularization factors comprises the positive reals.
The first part is as follows.
Let the function $k: \reals \to (0, \infty)$, be such that
\begin{IEEEeqnarray}{rcl}
\label{EqkWA}
k(b) & = &  \int \dot{f}^{-1}(\frac{-b - \foo{L}_{\dset{z}}(\thetav)}{\lambda} ) \diff Q (\vect{\theta}).
\end{IEEEeqnarray}
The first step is to prove that the function $k$ in~\eqref{EqkWA} is continuous in $\reals$. This is proved by showing that $k$ always exhibits a limit. 
Note that from Lemma~\ref{lemm_f_invIsInc}~in Appendix~\ref{sec:AppendixA}, the function $\dot{f}^{-1}$ is strictly increasing; thus, it holds that for all $b \in \set{B}_{Q, \dset{z}}$ with $\set{B}_{Q, \dset{z}}$ defined in~\eqref{EqDefNormFunction} and for all $\vect{\theta} \in \supp Q$, it holds that 
\begin{IEEEeqnarray}{rcl}
\label{EqIDidntSee}
\dot{f}^{-1}(\frac{-b - \foo{L}_{\dset{z}}(\thetav)}{\lambda}) & \leq  & \dot{f}^{-1}(-\frac{b}{\lambda}), \end{IEEEeqnarray}
where equality holds if and only if $ \foo{L}_{\dset{z}}(\thetav)  = 0$.
Now, from the \cite[Corollary~24.5.1]{rockafellar1970conjugate} the function $\dot{f}^{-1}$ is continuous, such that for all $\beta \in \set{B}$, it holds that
\begin{IEEEeqnarray}{rcl}
\label{EqSuchABeautiflFace2}
\lim_{b \to \beta} \dot{f}^{-1}(\frac{-b - \foo{L}_{\dset{z}}(\thetav)}{\lambda}) & = &  \dot{f}^{-1}(\frac{-\beta - \foo{L}_{\dset{z}}(\thetav)}{\lambda}).
\end{IEEEeqnarray}
Hence, from the dominated convergence theorem~\cite[Theorem~$1.6.9$]{ash2000probability}, the following limit exists and satisfies
\begin{IEEEeqnarray}{rcl}
\label{EqSuchACrazyDay}
\lim_{b \to \beta} k(b) & = & \lim_{b \to \beta}  \int \dot{f}^{-1}(\frac{-b - \foo{L}_{\dset{z}}(\thetav)}{\lambda} ) \diff Q (\vect{\theta})\\
& = &  \int  (\lim_{b \to \beta} \dot{f}^{-1}(\frac{-b   - \foo{L}_{\dset{z}}(\thetav)}{\lambda} ) )\diff Q (\vect{\theta})\\
& = & \int  \dot{f}^{-1}(\frac{-\beta - \foo{L}_{\dset{z}}(\thetav)}{\lambda}  )\diff Q (\vect{\theta})\\
& = & k(\beta),
\end{IEEEeqnarray}
which proves that the function $k$ in~\eqref{EqkWA} is continuous.

The second part proceeds as follows.
Consider an arbitrary $\lambda \in (0, \infty)$. From \cite[Corollary~24.5.1]{rockafellar1970conjugate} and Lemma~\ref{lemm_f_invIsInc}~in Appendix~\ref{sec:AppendixA}, the function $\dot{f}^{-1}$ is continuous and strictly increasing. Consequently, the function $k(b)$ defined in~\eqref{EqkWA} is strictly decreasing with respect to $b$.
Given the range of $\dot{f}^{-1}$, there exist values $b_1 \in (b^{\star}_{Q, \dset{z}},b)$ and $b_2 \in (b,\infty)$ such that the integral is bounded above and below unity
\begin{equation}
\label{EqfDRforTs}
\int \dot{f}^{-1}\left(-\frac{b_1+\foo{L}_{\dset{z}}(\thetav)}{\lambda}\right) \diff Q(\thetav)  >  1 
>  \int \dot{f}^{-1}\left(-\frac{b_2+\foo{L}_{\dset{z}}(\thetav)}{\lambda}\right) \diff Q(\thetav).
\end{equation}
Since $k(b)$ is continuous (as proven in the first part) and satisfies the bounds in~\eqref{EqfDRforTs}, the intermediate-value theorem~\cite[Theorem~$4.23$]{rudin1953bookPrinciples} guarantees the existence of a real $b \in \set{B}_{Q, \dset{z}}$ such that $k(b) = 1$. Furthermore, the strict monotonicity of $k(b)$ ensures that this solution $b$ is unique for the chosen $\lambda$.
Finally, since this result holds for any arbitrary $\lambda \in (0, \infty)$, it follows that the set of admissible regularization factors $\set{A}_{Q, \dset{z}}$ in~\eqref{EqDefMapNormFunction} is identical to $(0,\infty)$, which completes the proof.
\end{IEEEproof}
\section{Proof of Lemma~\ref{lemm_ERM_RER_Divf_ineq}}
\label{app_lemm_ERM_RER_Divf_ineq}
\begin{IEEEproof}\label{Proof_lemm_ERM_RER_Divf_ineq}
For all~$(\thetav_1, \thetav_2) \in \left( \supp Q \right)^2$, such that
\begin{equation}
\label{EqApril21at20h43in2025atHome}
\foo{L}_{\dset{z}}(\thetav_1)\leq \foo{L}_{\dset{z}}(\thetav_2),
\end{equation}
it follows that 
\begin{IEEEeqnarray}{rCl}
\label{EqLinearRelation}
	-\frac{\foo{L}_{\dset{z}}(\thetav_1) + N_{Q, \dset{z}}(\lambda)}{\lambda} 
	& \leq & - \frac{\foo{L}_{\dset{z}}(\thetav_2) + N_{Q, \dset{z}}(\lambda)}{\lambda},
\end{IEEEeqnarray}
where the function $N_{Q, \dset{z}}$ is defined in~\eqref{EqDefNormFunction}; and 
 equality holds if and only if \eqref{EqLinearRelation} holds with equality. 
From Lemma \ref{lemm_f_invIsInc}~in Appendix~\ref{sec:AppendixA}, the function $\dot{f}^{-1}$ is strictly increasing. Then, the monotonicity of $\dot{f}^{-1}$ and \eqref{EqLinearRelation} imply that
\begin{IEEEeqnarray}{rCl}
\label{EqGenpdfDivfWithN}
	\dot{f}^{-1}(- \frac{\foo{L}_{\dset{z}}(\thetav_1) + \beta}{\lambda}) 
	& \geq & \dot{f}^{-1}(- \frac{\foo{L}_{\dset{z}}(\thetav_2) + \beta}{\lambda}).
\end{IEEEeqnarray}
The proof is completed by noticing that from \eqref{EqGenpdfDivfWithN}, the inequality above can be rewritten as
\begin{equation}
	\frac{\diff \Pgibbs{P}{Q}}{\diff Q}(\thetav_1) 
	\geq \frac{\diff \Pgibbs{P}{Q}}{\diff Q}(\thetav_2),
\end{equation}
which completes the proof.
\end{IEEEproof}
\section{Proof of Lemma~\ref{lemm_ERM_RER_Divf_RNdBounds}}
\begin{IEEEproof}
\label{app_lemm_ERM_RER_Divf_RNdBounds}
From Lemma~\ref{lemm_ERM_RER_Divf_ineq}, it follows that for all $\lambda > 0$ such that $\beta \in \set{B}$, with $\set{B}$ defined in~\eqref{EqDefSetB}, it holds that for all $\thetav \in \supp Q$, and for all $\phiv \in \set{L}^{\star}_{Q, \dset{z}} \cap \supp Q$, 
\begin{align}
\label{Eq_lem_DivfRNdBounds1}
	\frac{\diff \Pgibbs{P}{Q}}{\diff Q}(\thetav) \leq & \frac{\diff \Pgibbs{P}{Q}}{\diff Q}(\phiv)\\
	& =  \dot{f}^{-1}(-\frac{\foo{L}_{\dset{z}}(\phiv) + N_{Q, \dset{z}}(\lambda)}{\lambda})
	\label{Eq_lem_DivfRNdBounds2}\\
	& =  \dot{f}^{-1}(-\frac{\delta^\star_{Q, \dset{z}} + N_{Q, \dset{z}}(\lambda)}{\lambda})
	\label{Eq_lem_DivfRNdBounds3}\\
	& <  \infty, \label{Eq_lem_DivfRNdBounds4}
\end{align}
where~\eqref{Eq_lem_DivfRNdBounds1} follows from~\eqref{EqGenpdffDv};~\eqref{Eq_lem_DivfRNdBounds2} follows from the fact that $\foo{L}_{\dset{z}}(\phiv) \geq \delta^\star_{Q, \dset{z}}$;
and~\eqref{Eq_lem_DivfRNdBounds4} follows from the fact that for all $\lambda > 0$, the function $N_{Q, \dset{z}}(\lambda) < \infty$. 
From the definition of $\delta^\star_{Q, \dset{z}}$ in~\eqref{EqDefDeltaStar} and $\set{L}^{\star}_{Q, \dset{z}}$ in~\eqref{EqDefSetLStarQz} equality in~\eqref{Eq_lem_DivfRNdBounds1} holds if and only if $\thetav \in \set{L}^{\star}_{Q, \dset{z}} \cap \supp Q$. 
This completes the proof of finiteness.

For the proof of positivity, observe that from Lemma~\ref{lemm_mutuallyAbsCont}, it holds that for all $ \thetav \in \supp Q$,  
\begin{IEEEeqnarray}{rCl}
	\frac{\diff \Pgibbs{P}{Q}}{\diff Q}(\thetav) 
	& > &\> 0, 
\end{IEEEeqnarray}
which completes the proof. 
\end{IEEEproof}

\section{Proof of Theorem~\ref{Theo_ERM_fDR_LT}}
\begin{IEEEproof}
\label{app_theo_ERM_fDR_LT}
The Legendre-Fenchel transform of a strictly convex function $f:\set{I}\rightarrow \reals$ satisfies
\begin{IEEEeqnarray}{rCl}
	f^{*}(t) & \triangleq & \sup_{s\in \set{I}}( ts- f(s)).
\end{IEEEeqnarray}
From~\cite[Theorem 23.5]{rockafellar1970conjugate} if $f$ is strictly convex then maximizing argument of the convex conjugate $f^{*}$ satisfies
\begin{IEEEeqnarray}{rCl}
\label{Eq_convcnj_pf}
	f^{*}(t) & = &  t\frac{\diff }{\diff t}f^{*}(t)- f(\frac{\diff }{\diff t}f^{*}(t)).
\end{IEEEeqnarray}
Furthermore, from~\cite[Corollary~23.5.1]{rockafellar1970conjugate} the function $\dot{f}^{-1}$ is the derivative of the convex conjugate of~$f$ in~\eqref{Eq_convcnj_pf}, which implies that
Differential Equations
\begin{IEEEeqnarray}{rCl}
\label{EqLegendreFencheclID}
	f^{*}(t) & = & t\dot{f}^{-1}(t)- f(\dot{f}^{-1}(t)).
\end{IEEEeqnarray}
From Theorem~\ref{Theo_f_ERMRadNik}, let $t = -\frac{\foo{L}_{\dset{z}}(\thetav)+\beta}{\lambda}$ in \eqref{EqLegendreFencheclID}, then it holds that for all $\thetav \in \supp Q$,
\begin{IEEEeqnarray}{rCl}
\label{Eq_conjugate_between_pf}
	f^{*}(-\frac{\foo{L}_{\dset{z}}(\thetav)+\beta}{\lambda}) 
	& = &  -\frac{\foo{L}_{\dset{z}}(\thetav)+\beta}{\lambda}\frac{\diff \Pgibbs{P}{Q}}{\diff Q}(\thetav)- f(\frac{\diff \Pgibbs{P}{Q}}{\diff Q}(\thetav)).
\end{IEEEeqnarray}
Taking the integral of~\eqref{Eq_conjugate_between_pf} with respect to the reference measure $Q$, yields
\begin{IEEEeqnarray}{rCl}
	\IEEEmulticolR{
	\int\!f^{*}(-\frac{\foo{L}_{\dset{z}}(\thetav)\!+\!\beta}{\lambda}) \diff Q(\thetav)
	}
	& = &  \int(-\frac{\foo{L}_{\dset{z}}(\thetav)\!+\!\beta}{\lambda}\frac{\diff \Pgibbs{P}{Q}}{\diff Q}(\thetav)- f(\frac{\diff \Pgibbs{P}{Q}}{\diff Q}(\thetav)\!))\diff Q(\thetav)\label{Eq_conjugate_between_pf_s1}\\
	& = & -\frac{1}{\lambda}(\foo{R}_{\dset{z}}(\Pgibbs{P}{Q})+\beta)-\Divf{\Pgibbs{P}{Q}}{Q}\label{Eq_conjugate_between_pf_s2}.
\end{IEEEeqnarray}
Arranging~\eqref{Eq_conjugate_between_pf_s2} results in
\begin{IEEEeqnarray}{rCl}
	\foo{R}_{\dset{z}}(\Pgibbs{P}{Q}) + \lambda\Divf{\Pgibbs{P}{Q}}{Q}
	& = & -\lambda\int f^{*}(-\frac{\foo{L}_{\dset{z}}(\thetav)+\beta}{\lambda}) \diff Q(\thetav)-\beta ,
\end{IEEEeqnarray}
which completes the proof.

The second part of the proof is as follows.
From Lemma~\ref{lemm_mutuallyAbsCont}, equality~\eqref{Eq_conjugate_between_pf} can be rewritten as
\begin{IEEEeqnarray}{rCl}
\IEEEmulticolR{
\frac{\foo{L}_{\dset{z}}(\thetav)+\beta}{\lambda}
}
& = & -f^{*}(-\frac{\foo{L}_{\dset{z}}(\thetav)+\beta}{\lambda})\frac{\diff Q}{\diff \Pgibbs{P}{Q}}(\thetav)- \frac{\diff Q}{\diff \Pgibbs{P}{Q}}(\thetav)f(\frac{\diff \Pgibbs{P}{Q}}{\diff Q}(\thetav)).\quad
\label{Eq_conjugate_between_pf2}
\end{IEEEeqnarray}
Taking the integral of~\eqref{Eq_conjugate_between_pf2} with respect to the reference measure $Q$, yields
\begin{IEEEeqnarray}{rCl}
\label{Eq_conjugate_deff_pf2}
	\frac{1}{\lambda}\foo{R}_{\dset{z}}(Q) + \frac{\beta}{\lambda} 
	& = & - \int f^{*}(-\frac{\foo{L}_{\dset{z}}(\thetav)+\beta}{\lambda})\frac{\diff Q}{\diff \Pgibbs{P}{Q}}(\thetav) \diff Q(\thetav) 
	\splitR 
	- \int \frac{\diff Q}{\diff \Pgibbs{P}{Q}}(\thetav)f(\frac{\diff \Pgibbs{P}{Q}}{\diff Q}(\thetav))\diff Q(\thetav).
\end{IEEEeqnarray}
Arranging~\eqref{Eq_conjugate_deff_pf2} results in
\begin{IEEEeqnarray}{rCl}
	\IEEEmulticolR{
	\foo{R}_{\dset{z}}(Q) +\lambda\int \frac{\diff Q}{\diff \Pgibbs{P}{Q}}(\thetav)f(\frac{\diff \Pgibbs{P}{Q}}{\diff Q}(\thetav)) \diff Q(\thetav)
	}
	& = & -\lambda\int f^{*}(-\frac{\foo{L}_{\dset{z}}(\thetav)+\beta}{\lambda}) \frac{\diff Q}{\diff \Pgibbs{P}{Q}}(\thetav) \diff Q(\thetav)-\beta ,
\end{IEEEeqnarray}
which completes the proof.
\end{IEEEproof}

\section{Proof of Theorem~\ref{Theo_ERMfDRType1To2}}
\label{app_lemm_ERM_fDR_f2g}
\begin{IEEEproof}
Note that from Theorem~\ref{Theo_f_ERMRadNik}, the functions $f$ and $g$ are differentiable and strictly convex. Hence, the functional inverse of the derivative is well-defined from the fact that $\dot{f}$ and $\dot{g}$ are strictly increasing and bijective.
Denote by $\Pgibbs{\hat{P}}{Q}$ the solution to the optimization problem in~\eqref{EqOpERMLink_fDiv}. Then, from Theorem~\ref{Theo_f_ERMRadNik}, for all $\thetav\in \supp{Q}$, it follows that
\begin{IEEEeqnarray}{rcl}
\frac{\diff \Pgibbs{\hat{P}}{Q}}{\diff Q}(\thetav)
& = & \dot{g}^{-1}(-\frac{\hat{N}_{Q, \dset{z}}(\lambda) + v(\foo{L}_{\dset{z}}(\thetav))}{\lambda})
\label{EqProof_DivT1_T2_s0}\\
& = & \dot{g}^{-1}(\dot{g}(\dot{f}^{-1}(-\frac{N_{Q, \dset{z}}(\lambda) + \foo{L}_{\dset{z}}(\thetav)}{\lambda}))+\frac{c-\hat{N}_{Q, \dset{z}}(\lambda)}{\lambda} ),\label{EqProof_DivT1_T2_s1}
\end{IEEEeqnarray}
where $\hat{N}_{Q, \dset{z}}$ is the normalization function of the optimization problem in \eqref{EqOpERMLink_fDiv}.
Then, under the assumption that $\hat{N}_{Q, \dset{z}}$ satisfies $\hat{N}_{Q, \dset{z}}(\lambda)>c$, from the monotonicity of $\dot{g}^{-1}$, for all $\thetav \in \supp Q$,
\begin{IEEEeqnarray}{rcl}
\IEEEeqnarraymulticol{3}{l}{
	\dot{g}^{-1}(\dot{g}(\dot{f}^{-1}(-\frac{N_{Q, \dset{z}}(\lambda) + \foo{L}_{\dset{z}}(\thetav)}{\lambda}))-\frac{\hat{N}_{Q, \dset{z}}(\lambda) + c}{\lambda} )
	}\nonumber \\ & < & \dot{f}^{-1}(-\frac{N_{Q, \dset{z}}(\lambda) + \foo{L}_{\dset{z}}(\thetav)}{\lambda}).
	\label{EqFromf2gfDRLess1}
\end{IEEEeqnarray}
Similarly, under the assumption that $\hat{N}_{Q, \dset{z}}$ satisfies $\hat{N}_{Q, \dset{z}}(\lambda)<c$, from the monotonicity of $\dot{g}^{-1}$, for all $\thetav \in \supp Q$,
\begin{IEEEeqnarray}{rcl}
\IEEEeqnarraymulticol{3}{l}{
	\dot{g}^{-1}(\dot{g}(\dot{f}^{-1}(-\frac{N_{Q, \dset{z}}(\lambda) + \foo{L}_{\dset{z}}(\thetav)}{\lambda}))-\frac{\hat{N}_{Q, \dset{z}}(\lambda) + c}{\lambda} )
	}\nonumber \\ & > & \dot{f}^{-1}(-\frac{N_{Q, \dset{z}}(\lambda) + \foo{L}_{\dset{z}}(\thetav)}{\lambda}).
	\label{EqFromf2gfDRMore1}
\end{IEEEeqnarray}
Note that $\hat{N}_{Q, \dset{z}}$ needs to be such that,
\begin{equation}
	1 = \int \frac{\diff \Pgibbs{\hat{P}}{Q}}{\diff Q}(\thetav) \diff Q(\thetav).
\end{equation}
Under the assumption that $\hat{N}_{Q, \dset{z}}(\lambda)>c$, from \eqref{EqFromf2gfDRLess1}, it follows that
\begin{IEEEeqnarray}{rCl}
	\int \frac{\diff \Pgibbs{\hat{P}}{Q}}{\diff Q}(\thetav) \diff Q(\thetav) & < & \int \dot{f}^{-1}(-\frac{N_{Q, \dset{z}}(\lambda) + \foo{L}_{\dset{z}}(\thetav)}{\lambda}) \diff Q(\thetav)\\
	& = & 1.
\end{IEEEeqnarray}
Similarly, under the assumption that $\hat{N}_{Q, \dset{z}}(\lambda)<c$, from \eqref{EqFromf2gfDRMore1}, it follows that
\begin{IEEEeqnarray}{rCl}
	\int \frac{\diff \Pgibbs{\hat{P}}{Q}}{\diff Q}(\thetav) \diff Q(\thetav) & > & \int \dot{f}^{-1}(-\frac{N_{Q, \dset{z}}(\lambda) + \foo{L}_{\dset{z}}(\thetav)}{\lambda}) \diff Q(\thetav)\\
	& = & 1.
\end{IEEEeqnarray}
Therefore, the normalization function $\hat{N}_{Q, \dset{z}}$ satisfies
\begin{equation}
\label{EqNqzHatIsC}
	\hat{N}_{Q, \dset{z}}(\lambda) = c.
\end{equation}
Thus, from \eqref{EqProof_DivT1_T2_s1} and \eqref{EqNqzHatIsC}, for all $\thetav \in \supp Q$ that 
\begin{IEEEeqnarray}{rcl}
\label{EqProof_DivT1_T2_s2}
\frac{\diff \Pgibbs{\hat{P}}{Q}}{\diff Q}(\thetav)
& = & \dot{f}^{-1}(-\frac{N_{Q, \dset{z}}(\lambda) + \foo{L}_{\dset{z}}(\thetav)}{\lambda})
\label{EqProof_DivT1_T2_s3}\\
& = & \frac{\diff \Pgibbs{P}{Q}}{\diff Q}(\thetav) \label{EqProof_DivT1_T2_s4},
\end{IEEEeqnarray}
where~\eqref{EqProof_DivT1_T2_s2} follows from~\eqref{EqFromf2gfDR}, which completes the proof.
\end{IEEEproof}

\section{Numerical Simulation}
\label{AppNumericalSimulation}

The MNIST dataset consists of $60{,}000$ images for training and $10{,}000$ images for testing. Out of the $60{,}000$ training images, $12{,}183$ are labeled as the digits six or seven, while $1{,}986$ out of the $10{,}000$ test images correspond to these digits. Each image is a $28 \times 28$ grayscale picture and is represented by the matrix $I \in [0,1]^{28\times28}$.

\subsection{Feature Extraction of the Histogram of Oriented Gradients}
\label{SubsectionHOG}

The grayscale images are processed by calculating their corresponding \emph{histogram of oriented gradients} (HOG) \cite{dalal2005histograms}. The HOG for each image is computed through the following steps:

$1.\big)$
For each pixel location $(i, j) \in \{1,2,...,28\}^{2}$ in the image, the gradients in the $w$- and $h$-directions (\emph{width, height}) are computed using finite differences given by the functions $\foo{G}_w:\{1,2,...,28\}^{2}\rightarrow \reals$ and $\foo{G}_h:\{1,2,...,28\}^{2}\rightarrow \reals$, which are defined as
\begin{IEEEeqnarray}{rCl}
\foo{G}_w(i, j) & = & \begin{cases}
	I(i+1, j) - I(i-1, j)& \!\!\!\text{if } i \in \{2,\ldots,27\} \\
	I(i+1, j) - I(i, j)  & \!\!\!\text{if } i =1 \\
	I(i, j) - I(i-1, j)  & \!\!\!\text{if } i =28
\end{cases}\!\!,\qquad \\
\noalign{\noindent and \vspace{2\jot}}
\foo{G}_h(i, j) & = & \begin{cases}
	I(i, j+1) - I(i, j-1)& \!\!\!\text{if } j \in \{2,\ldots,27\} \\
	I(i, j+1) - I(i, j)  & \!\!\!\text{if } j =1 \\
	I(i, j) - I(i, j-1)  & \!\!\!\text{if } j =28
\end{cases}\!\!\!, \qquad
\end{IEEEeqnarray}
where $I(i, j) \in [0,1]$ represents the pixel intensity at location $(i, j)$.

$2.\big)$
Given a pixel location $(i, j) \in \{1,2,...,28\}^{2}$, the magnitude and orientation of a pixel at location $(i, j)$ is given by the functions $\foo{M}:\{1,2,...,28\}^{2}\rightarrow \reals$ and  $\phi:\{1,2,...,28\}^{2}\rightarrow \reals$, such that
\begin{IEEEeqnarray}{rCl}
\label{EqDefMagnitudeM}
\foo{M}(i, j) & = & \sqrt{\foo{G}_w(i, j)^2 + \foo{G}_h(i, j)^2}, \text{ and} \\
\label{EqDefOrientationP}
\phi(i, j) & = & \arctan\left(\frac{\foo{G}_h(i, j)}{\foo{G}_w(i, j)}\right).
\end{IEEEeqnarray}

$3.\big)$ The matrix $I$ is divided into sub-matrices of size $4 \times 4$, such that the number of sub-matrices is $7$. 
These sub-matrices are referred to as \emph{cells} and are denoted, for all $(w,h) \in \{1,\cdots, 7\}^2$, by
\begin{IEEEeqnarray}{rCl}
\label{EqDefCellCm}
C_{w,h} & = & \begin{bmatrix}
 	 I(a_w, b_h)  & \cdots  & I(a_w+3, b_h)\\
 	 \vdots    & \ddots     & \vdots   \\
 	 I(a_w, b_h+3)  & \cdots  & I(a_w+3, b_h+3)
 \end{bmatrix},\quad
\end{IEEEeqnarray}
where the real values $a_w$ and $b_h$ are
\begin{IEEEeqnarray}{rCl}
\label{EqDef_aw}
a_w & = & 4(w-1)+1\\
b_h & = & 4(h-1)+1.
\label{EqDef_bh}
\end{IEEEeqnarray}
This implies that the matrix $I$ can be represented as
\begin{IEEEeqnarray}{rCl}
\label{EqDefImgAsCell}
I & = & \begin{bmatrix}
 	C_{1,1}    & \cdots  & C_{7,1}\\
 	 \vdots    & \ddots  & \vdots   \\
 	 C_{1,7} & \cdots  & C_{7,7}
 \end{bmatrix}.
\end{IEEEeqnarray}
From~\eqref{EqDefCellCm}, the set of all pairs $(i,j)$ of pixel coordinates in $I$ that lie within the cell $C_{w,h}$ is given by:
\begin{IEEEeqnarray}{rCl}
\set{A}_{w,h} & = & \{a_w,a_w+3\}\times\{b_h,b_h+3\},
\end{IEEEeqnarray}
with $a_w$ in~\eqref{EqDef_aw} and $b_h$ in~\eqref{EqDef_bh}.

$4.\big)$ For each cell $C_{w,h}$ in~\eqref{EqDefCellCm} the orientations $\phi(i, j)$ in~\eqref{EqDefOrientationP} are divided into $9$ bins. That is, the $n^{th}$ bin, with $1\leq n \leq 9$, satisfies that
\begin{IEEEeqnarray}{rCl}
\IEEEeqnarraymulticol{3}{l}{
\set{B}^{(n)}_{w,h} 
}\nonumber \\
	& = & \{\phi(i, j) \in \reals:
180(\frac{n-1}{9})\leq \phi(i, j) < 180(\frac{n}{9}):
	\right. \nonumber\\ &  & \left.
 	(i,j)\in \set{A}_{w,h} \vphantom{180(\frac{n-1}{9})} \}. 
 	\label{EqDefNthbin}
\end{IEEEeqnarray}
The contribution of each pixel to its corresponding bin is based on its gradient magnitude. That is, the value of the $n$-th bin from the $(w,h)$-th cell $C_{w,h}$ in~\eqref{EqDefCellCm} is given by the function $H_{w,h}(n): \{1,2, \ldots, 9\} \rightarrow \reals$, such that
\begin{IEEEeqnarray}{rCl}
\label{EqDefHistBinN}
H_{w,h}(n) & = & \sum_{(i, j) \in \set{A}_{w,h}} \foo{M}(i, j)\ind{\phi(i, j) \in \set{B}^{(n)}_{w,h}},
\end{IEEEeqnarray}
with $\foo{M}$ in~\eqref{EqDefMagnitudeM}; and $\set{B}^{(n)}_{w,h}$ in~\eqref{EqDefNthbin}. Thus, the histogram of gradient orientations of the cell $C_{w,h}$ is represented by the vector $\vect{H}_{{w,h}} \in \reals^{9}$, such that
\begin{IEEEeqnarray}{rCl}
\label{EqDefHistVect}
\vect{H}_{w,h} & = & [H_{w,h}(1),H_{w,h}(2),\cdots H_{w,h}(9)],
\end{IEEEeqnarray}
with the function $H_{w,h}$ in~\eqref{EqDefHistBinN}.

$5.\big)$
To account for illumination and contrast variations, the histogram $\vect{H}_{w,h}$ in~\eqref{EqDefHistVect} is normalized. To normalize the histograms for all cells $C_{w,h}$ in~\eqref{EqDefCellCm}, the cells are grouped into sub-matrices formed by $2 \times 2$ cells with a \emph{cell overlap} set to $1$ pixel, such that  the number of sub-matrices is:
\begin{IEEEeqnarray}{rCl}
\label{EqNumBlocks}
(7-1)\times (7-1)	& = & 36.
\end{IEEEeqnarray}
These sub-matrices of the matrix $I$ in~\eqref{EqDefImgAsCell}, with $(m,s) \in \{1, \cdots \sqrt{36}\}^2$ are referred to as \emph{blocks}, and denoted by
\begin{IEEEeqnarray}{rCl}
\label{EqDefBlockAsCell}
B_{m,s} & = & \begin{bmatrix}
 	C_{m,s}  & C_{m+1,s}\\
 	C_{m,s+1}& C_{m+1,s+1}
 \end{bmatrix},
\end{IEEEeqnarray}
with the matrix $C_{m,s}$ in~\eqref{EqDefCellCm}.
From~\eqref{EqDefImgAsCell} and~\eqref{EqDefBlockAsCell}, a block $B_{m,s}$ is a sub-matrix of size $8\times8$, \ie, $B_{m,s} \in \reals^{8\times8}$. The \emph{size} of a block is given by the ratio of the total number of pixels in a block to the number of pixels in a cell:
\begin{subequations}
\label{EqNumSizeBlock}
\begin{IEEEeqnarray}{rCl}
\frac{8\times 8}{4\times 4}
  & = & 4,
\end{IEEEeqnarray}
\end{subequations}
The normalized histogram of a cell $C_{w,h}$ in a block $B_{m,s}$ is denoted by the vector $\hat{\vect{H}}^{(m,s)}_{w,h} \in \reals^9$. This normalization of is typically done using the $\ell_2$-norm, such that
\begin{IEEEeqnarray}{rCl}
\hat{\vect{H}}^{(m,s)}_{w,h} & = & \frac{\vect{H}_{w,h}}{\sqrt{\displaystyle{\sum_{(i,j) \in \{m,m+1\}\times\{s,s+1\}}}\vect{H}_{i,j}^2 + \epsilon^2}},
\end{IEEEeqnarray}
where $\vect{H}_{w,h}$ in~\eqref{EqDefHistVect} is the unnormalized histogram, and $\epsilon = 10^{-4}$ to avoid division by zero.

$6.\big)$
For an image with 36 blocks (see~\eqref{EqNumBlocks}), 9 orientation bins, and a block size of 4 (see~\eqref{EqNumSizeBlock}), the dimension of the HOG feature vector $\hat{\mathbf{x}}$ is:
\begin{IEEEeqnarray}{rCl}
\label{EqIntHOGVectLength}
36 \times 4 \times 9 	& = & 1296.
\end{IEEEeqnarray}
The HOG feature vector $\hat{\mathbf{x}} \in \reals^{1296}$ is formed by concatenating all the normalized histograms $\hat{\vect{H}}^{(m,s)}_{w,h}$ such that
\begin{IEEEeqnarray}{rCl}
\label{EqDefHOGPattern}
\hat{\mathbf{x}} 
	& = & [ \hat{\vect{H}}^{(1,1)}_{1,1},\hat{\vect{H}}^{(1,1)}_{1,2},\hat{\vect{H}}^{(1,1)}_{2,1},\hat{\vect{H}}^{(1,1)}_{2,2},
	\right.\nonumber \\ &  &\left. \>
	\hat{\vect{H}}^{(2,1)}_{2,1},\hat{\vect{H}}^{(2,1)}_{2,2},\hat{\vect{H}}^{(2,1)}_{3,1},\hat{\vect{H}}^{(2,1)}_{3,2}, \cdots, 
	\right.\nonumber \\ &  &\left. \>
	\hat{\vect{H}}^{(6,6)}_{6,6},\hat{\vect{H}}^{(6,6)}_{6,7},\hat{\vect{H}}^{(6,6)}_{7,6},\hat{\vect{H}}^{(6,6)}_{7,7}]^\top.
\end{IEEEeqnarray}

\subsection{Principal Component Analysis}
The final step in the data processing is to reduce the dimensionality of the pattern $\hat{\mathbf{x}}$ in~\eqref{EqDefHOGPattern} from $\reals^{1296}$ to $\reals^{2}$, while ensuring that the important structure of the pattern is preserved. In this simulation, \textit{principal component analysis (PCA)} is used to project the high-dimensional data onto a lower-dimensional subspace. 
From $60{,}000$ images for training in the MNIST, the HOG of two handwritten numbers (in this simulation $6$ and $7$) are computed, as mentioned in Appendix~\ref{SubsectionHOG}. The resulting $12{,}183$ HOG vectors $\hat{\mathbf{x}}  \in \reals^{1296}$ are reduced to $\reals^{2}$ using PCA as follows:

$1.\big)$ To reduce the dimensionality, the first step in PCA is to compute the \textit{covariance matrix} of the data. This matrix captures the relationships between the different features (or dimensions) of the data. The covariance matrix is calculated as follows:

\begin{IEEEeqnarray}{rCl}
\label{EqCovarianceMatrix}
\mathbf{C} & = & \frac{1}{n-1} \sum_{i=1}^n (\hat{\mathbf{x}} _i - \muv)(\hat{\mathbf{x}} _i - \muv)^\top,
\end{IEEEeqnarray}

where $n = 12{,}183$, $\mathbf{C} \in \reals^{1296 \times 1296}$, and $\muv$ is the mean of all the training patterns given by
\begin{IEEEeqnarray}{rCl}
	\muv & = & \frac{1}{n} \sum_{i=1}^n \hat{\mathbf{x}} _i.
\end{IEEEeqnarray}

$2.\big)$ The next step in PCA is to perform an \textit{eigenvalue decomposition} of the covariance matrix $\mathbf{C}$ in~\eqref{EqCovarianceMatrix}. The decomposition can be written as:
\begin{IEEEeqnarray}{rCl}
\mathbf{C} & = & \mathbf{V} \mathbf{\Lambda} \mathbf{V}^\top,
\end{IEEEeqnarray}
where $\mathbf{V} \in \reals^{1296 \times 1296}$ is a matrix whose columns are the eigenvectors of $\mathbf{C}$ and $\mathbf{\Lambda} \in \reals^{1296 \times 1296}$ is a diagonal matrix containing the corresponding eigenvalues.

$3.\big)$ Following the computation of the eigenvectors, the dimensionality is reduced from $\reals^{1296}$ to $\reals^2$ by selecting the two eigenvectors associated with the largest eigenvalues. Denote these top two eigenvectors as $\mathbf{w}_1$ and $\mathbf{w}_2$. These eigenvectors constitute the columns of the projection matrix $\mathbf{W} \in \reals^{1296\times 2}$, defined as
\begin{IEEEeqnarray}{rCl}
\label{EqDefWPACsim}
\mathbf{W} & = & \left[ \mathbf{w}_1 \, \mathbf{w}_2 \right].
\end{IEEEeqnarray}

$4.\big)$ Once the projection matrix $\mathbf{W}$ is computed, each high-dimensional pattern $\hat{\mathbf{x}}  \in \reals^{1296}$ can be projected onto the new $\reals^2$ subspace. The projection is performed as follows:
\begin{IEEEeqnarray}{rCl}
\label{EqDefRlToR2}
\mathbf{x} & = & \mathbf{W}^\top \hat{\mathbf{x}} ,
\end{IEEEeqnarray}
with $\hat{\mathbf{x}}$ in~\eqref{EqDefHOGPattern}, $\vect{W}$ in~\eqref{EqDefWPACsim} and $\mathbf{x} \in \reals^2$ represents the 2-dimensional coordinates of the original pattern $\hat{\mathbf{x}}$ in the reduced-dimensional space.

\subsection{Simulation Dataset}

In this simulation, a datapoint is a tuple $(\hat{\mathbf{x}},y) \in \reals^{1296}\times\{6,7\}$, with $\hat{\mathbf{x}}$ in~\eqref{EqDefHOGPattern} and $y$ being the label assigned by MNIST to the image $I$ in~\eqref{EqDefImgAsCell}. The label $y$ corresponds to the digit in the image $I$. Such an image produces the vector $\hat{\mathbf{x}}$, when its HOG features are computed.

\end{document}